\documentclass{article}

\usepackage[main,final]{neurips_2026}

\usepackage[utf8]{inputenc} % allow utf-8 input
\usepackage[T1]{fontenc}    % use 8-bit T1 fonts
\usepackage{hyperref}       % hyperlinks
\usepackage{url}            % simple URL typesetting
\usepackage{booktabs}       % professional-quality tables
\usepackage{amsfonts}       % blackboard math symbols
\usepackage{nicefrac}       % compact symbols for 1/2, etc.
\usepackage{microtype}      % microtypography
\usepackage{xcolor}         % colors

\usepackage[pdftex]{graphicx}

\title{\mbox{\hspace{-1.5em} \raisebox{-0.2em}{\includegraphics[height=1em]{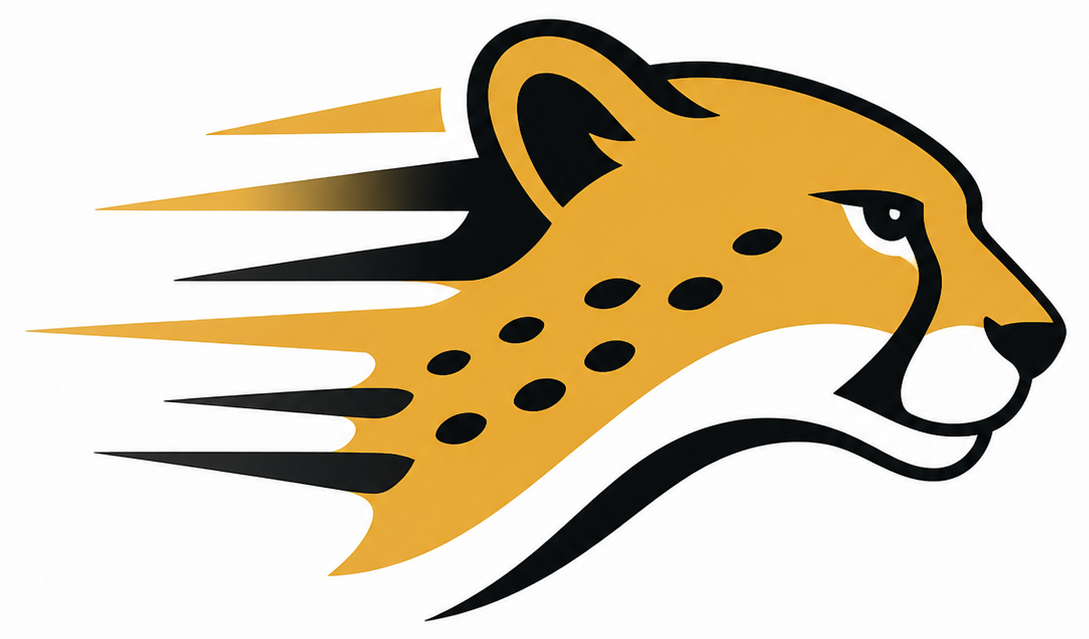}} Efficient LLM Adaptation with Forward-Only Passes}}

\author{%
  Baichuan Huang\thanks{baichuan.huang@eit.lth.se}\\
  Lund University, Sweden\\
  \And
  Ananth Balashankar\thanks{contributed in an advisory role.} \\
  Google DeepMind, USA \\
  \And
  Amir Aminifar \\
  Lund University, Sweden\\
}

\usepackage[acronym]{glossaries}
\usepackage{pifont}
\usepackage{xspace}
\usepackage{caption}
\usepackage{subcaption}
\usepackage[most]{tcolorbox}
\usepackage{booktabs}
\usepackage{multirow}
\usepackage{multicol}
\usepackage{tikz}
\usepackage{textcomp}

\newacronym{FF}{FF}{the Forward-Forward algorithm}
\newacronym{FO}{FO}{first-order}
\newacronym{ZO}{ZO}{zeroth-order}
\newacronym{SGD}{SGD}{stochastic gradient descent}
\newacronym{LLM}{LLM}{large language model}
\newacronym{LoRA}{LoRA}{low-rank adaptation}
\newacronym{BP}{BP}{Backpropagation}
\newacronym{IP}{IP}{in-place}
\newacronym{PEFT}{PEFT}{parameter-efficient fine-tuning}
\newacronym{FT}{FT}{fine-tuning}
\newacronym{MeZO}{MeZO}{memory-efficient zeroth-order optimizer}
\newacronym{CPU}{CPU}{central processing unit}
\newacronym{GPU}{GPU}{graphics processing unit}
\newacronym{RAM}{RAM}{random-access memory}
\newacronym{SPSA}{SPSA}{Simultaneous Perturbation Stochastic Approximation}
\newacronym{BEFT}{BEFT}{bias-efficient fine-tuning}
\newacronym{ICL}{ICL}{in-context learning}
\newacronym{seconds}{s}{seconds}
\newacronym{VeRA}{VeRA}{vector-based random matrix adaptation}
\newacronym{DoRA}{DoRA}{weight-decomposed low-rank adaptation}
\newacronym{SDPA}{SDPA}{scaled dot-product attention}
\newacronym{SVD}{SVD}{singular value decomposition}
\newacronym{FLOPs}{FLOPs}{floating point operations}
\newacronym{MB}{MB}{MegaByte}
\newacronym{GB}{GB}{Gigabyte}
\newacronym{Mul}{Mul}{Multiplication}
\newacronym{Add}{Add}{Addition}
\newacronym{CNN}{CNN}{convolutional neural network}
\newacronym{FC}{FC}{fully connected network}
\newacronym{GQA}{GQA}{grouped query attention}
\newacronym{RoPE}{RoPE}{rotary positional embedding}
\newacronym{MoE}{MoE}{mixture-of-experts} 
\newacronym{SFT}{SFT}{supervised fine-tuning} 
\newacronym{AI}{AI}{artificial intelligence}
\newacronym{RAG}{RAG}{retrieval-augmented generation}
\newcommand{\boldeq}{\ensuremath{\boldsymbol}} %for math
\usepackage{booktabs} 
\usepackage{amssymb}
\usepackage{amsmath}
\usepackage{algorithm}
\usepackage{algorithmic}

\glsdisablehyper

\usepackage{listings}
\usepackage{xcolor}

\usepackage[table]{xcolor}
\usepackage{colortbl}
\definecolor{mygreen}{HTML}{D5E8D4}
\definecolor{mydarkgreen}{HTML}{82B366}%{81b366}%{55A868}
\definecolor{myred}{HTML}{F8CECC}
\definecolor{mydarkred}{HTML}{B85450}%{C44F51}
\definecolor{myblue}{HTML}{DAE8FC}
\definecolor{mydarkblue}{HTML}{6C8EBF}%{6b8dbe}%{4C72B0}
\definecolor{mypurple}{HTML}{E1D5E7}
\definecolor{mydarkpurple}{HTML}{9673A6}
\definecolor{myyellow}{HTML}{FFF2CC}
\definecolor{mydarkyellow}{HTML}{D6B656}%{d5b556}
\definecolor{mygrey}{HTML}{F5F5F5}
\definecolor{mydarkgrey}{HTML}{666666}

\usepackage{threeparttable}

\usepackage{wrapfig}
\usepackage{fontawesome5}
\usepackage{pifont}

\usepackage{makecell}

\newcommand{\camera}{\textcolor{black}}
\begin{document}

\maketitle

\maketitle

\glsresetall
\begin{abstract}
\Gls{LLM} inference serving is undergoing rapid growth and large-scale deployment, motivating us to rethink how the inference process itself can be leveraged to enable efficient task-specific \gls{LLM} adaptation. In this paper, we propose a forward-only approach for efficient \gls{LLM} adaptation with forward-only passes in \gls{LLM} inference serving. We exploit the angle concentration of activations induced by each \gls{SVD} component to measure its contribution—\textit{dispersed} or \textit{concentrated}—to the angle concentration of the hidden states. Based on this, our forward-only search then efficiently identifies the weight matrix with the highest dispersion that merits rank reduction. We then selectively remove the higher-order components and retain the lower-order components in \gls{SVD}. Empirical results across diverse datasets demonstrate the competitive accuracy of our forward-only approach, while theoretical analysis shows lower peak memory and greater speedup than the gradient-based approach and greater speedup than the exhaustive search approach. The experiments further present the robustness and generalization of our forward-only approach to various \glspl{LLM} with up to 57B parameters. The code is included in the supplementary~material.
\end{abstract}

\glsresetall

\section{Introduction}
\Gls{AI} serving is emerging as a key driver of growth in energy consumption as \gls{AI} systems are deployed at scale \citep{Bashir2024Climate,maslej2025artificial,thundersaidenergy2026_internet_energy}, particularly for Transformer-based \glspl{LLM} inference serving \citep{elsworth2025measuringenvironmentalimpactdelivering,EpochAIModels2025}. Large-scale \gls{LLM} inference generated from real user interactions during \gls{LLM} serving motivates us to rethink how the inference process itself can be leveraged to improve the serving system, particularly enabling efficient task-specific \gls{LLM} adaptation.

The mainstream \gls{LLM} adaptation is based on tuning techniques (for instance, \gls{PEFT} \citep{hu2022lora,houlsby2019parameter}, prompt tuning \citep{lester2021power}, etc.) and knowledge augmentation techniques (for instance, \gls{RAG} \citep{lewis2020retrieval}, prompt engineering \citep{white2023prompt}, etc.). The knowledge augmentation techniques primarily operate by injecting knowledge and are orthogonal to modifying the internal weights during \gls{LLM} inference. The tuning techniques usually rely on backward passes, incurring extra resource overhead compared to inference itself. On the contrary, the emerging forward-only learning  \citep{hinton2022forward,ren2023scaling,zhang2024revisiting,huang2025efficient,Huang_FO_Survey_2026}, which relies solely on forward passes, is becoming a promising approach for adaptation in LLM inference.

\begin{figure}[!t]
 \centering
    \includegraphics[width=1\linewidth]{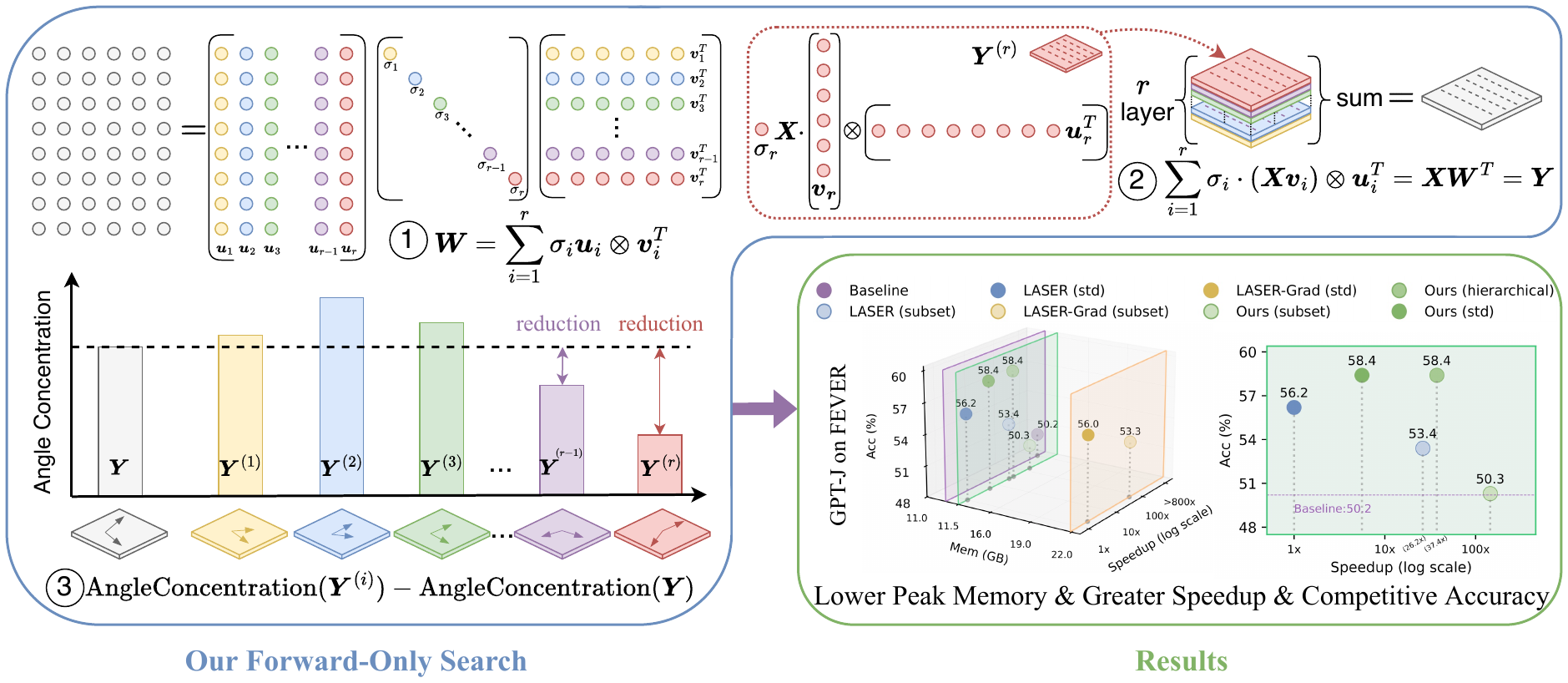}
    \vspace{-15pt}
  \caption{Efficient \gls{LLM} adaptation in \gls{LLM} inference serving. \textbf{Our forward-only search} contains \ding{172} \Gls{SVD} with multiple \gls{SVD} components; \ding{173} The sum of activations from all \gls{SVD} components recovers the standard forward pass; \ding{174} Forward-only passes measure the angle concentration to %identify which \gls{SVD} components should be reduced, and further
  identify which weight matrix has the highest dispersion that should be applied with rank reduction. \textbf{Results} show that our forward-only approach achieves lower peak memory and greater speedup than the gradient-based approach (LASER-Grad \citep{sreeram2025compress}), as well as greater speedup and competitive accuracy compared to the exhaustive search approach (LASER \citep{sharma2024the}).}
  \label{fig:main}
  \vspace{-15pt}
\end{figure}

In this paper, we aim to achieve efficient \gls{LLM} adaptation with forward-only passes during \gls{LLM} inference serving. We exploit the \gls{SVD} in \gls{LLM} compression techniques, based on the empirical observation that selectively removing the higher-order components of weight matrices in \gls{SVD} can improve downstream accuracy \citep{sharma2024the,sreeram2025compress}.\footnote{Unlike other \gls{LLM} compression techniques \citep{sanh2019distilbert,dettmers2023qlora,frantar2023sparsegpt,lin2025qserve} designed to maintain downstream accuracy, the standard \gls{SVD} is expected to enhance downstream accuracy by removing the higher-order components (smaller singular values) and retaining lower-order components (bigger singular values) as suggested by recent studies \citep{sharma2024the,sreeram2025compress}.} We propose the forward-only search to speed up the identification of the selected weight matrix, by the difference in angle concentration between activations induced by \gls{SVD} components and the hidden states. The angle concentration of activations induced by each \gls{SVD} component is to measure its contribution---\textit{dispersed} or \textit{concentrated}---to the angle concentration of the hidden states, as shown in Fig. \ref{fig:main}. If the angle concentration of activations induced by one \gls{SVD} component is more dispersed than the angle concentration of the hidden states, the corresponding \gls{SVD} component tends to be removed. Conversely, if more concentrated, the \gls{SVD} component tends to be retained. 
Our forward-only search relies solely on forward passes of \gls{LLM} inference, hence eliminating the need for backpropagated gradients in \cite{sreeram2025compress} and avoiding the computational burden of exhaustive search that sweeps all the weights in \cite{sharma2024the}.
Then, we investigate the rank-reduction strategy across different sizes of validation datasets to further improve the computational efficiency. 
The contributions of our paper are shown below:
\begin{itemize}
  \item We propose a forward-only approach for efficient \gls{LLM} adaptation in \gls{LLM} inference. By exploiting the difference in angle concentration between activations induced by \gls{SVD} components and the hidden states, our forward-only search identifies which \gls{SVD} components should be reduced, and further identifies which weight matrix merits rank reduction.
  \item We demonstrate that our forward-only approach achieves lower peak memory and greater speedup than the gradient-based approach, as well as greater speedup and competitive accuracy compared to the exhaustive search approach, across diverse datasets (including knowledge reasoning, knowledge intervention, multi-hop reasoning, fact verification, fairness and bias evaluation, hallucination detection, and information retrieval). The extended experiments further show the effectiveness and robustness of our forward-only approach, as well as its ability to enhance the fine-tuning process. Additionally, we conduct extensive experiments on various \glspl{LLM} with up to 57B parameters, including LLaMA2, DeepSeekV1, Mistral, Qwen2 and its \gls{MoE} version, and \camera{the multimodal} Gemma 4. The results show the strong generalization of our forward-only approach.
\end{itemize}
 \vspace{-15pt}

\section{Our Forward-Only Approach}
\vspace{-10pt}
\label{sec:method}
Our forward-only approach aims to identify which weight matrix in \glspl{LLM} should be applied with rank reduction and how to apply it using \gls{SVD}, for effective adaptation in \gls{LLM} inference serving.

\begin{minipage}[t]{0.4\textwidth}
\vspace{0pt}
\paragraph{Preliminary.}
Given a weight matrix $\boldeq{W} \in \mathbb{\mathbb{R}}^{h\times d}$, at a given layer of a pre-trained \gls{LLM}, we denote the hidden states of input as $\boldeq{X} \in \mathbb{\mathbb{R}}^{n\times d}$, the hidden states of output as $\boldeq{Y} \in \mathbb{\mathbb{R}}^{n\times h}$, where $n$ is the sequence length, $d$ is the size of input, and $h$ is the size of output. Therefore, the forward pass of this given layer is $\boldeq{X}\boldeq{W}^T=\boldeq{Y}$. Meanwhile, as shown in Fig. \ref{fig:main} \ding{172}, the \gls{SVD} of the weight matrix $\boldeq{W} \in \mathbb{\mathbb{R}}^{h\times d}$ is $\boldeq{W} = \boldeq{U}\boldeq{\Sigma}\boldeq{V}^T$, where $\boldeq{U}=[\boldeq{u}_1,\boldeq{u}_2,...,\boldeq{u}_r] \in \mathbb{\mathbb{R}}^{h\times r} $, $\boldeq{\Sigma}=\mathrm{diag}(\sigma_1,\sigma_2,...,\sigma_r) \in \mathbb{\mathbb{R}}^{r\times r} $, and $\boldeq{V}^T=[\boldeq{v}_1^T;\boldeq{v}_2^T;...;\boldeq{v}_r^T] \in \mathbb{\mathbb{R}}^{r\times d} $ as well as the rank $r \camera{\le}  \mathrm{min}(h,d)$. Equivalently, the forward pass of this given layer can be reformulated as the element-wise sum of activations induced by each \gls{SVD} component $\boldeq{Y}^{(i)} \in \mathbb{\mathbb{R}}^{n\times h}$, in Fig. \ref{fig:main} \ding{173}:
\end{minipage}
\hspace{4pt}
\scalebox{0.95}{
\begin{minipage}[t]{0.6\linewidth}
\vspace{-10pt}
\begin{algorithm}[H]
   \caption{Our Proposed Forward-Only Approach}
   \label{alg:main}
    \begin{algorithmic}[1]
   \STATE {\bfseries Given:} Weight matrices $\boldeq{W} $ with the \gls{SVD} ($\boldeq{U}, \boldeq{\Sigma}, \boldeq{V}^T$), and the corresponding input $\boldeq{X}$ with output $\boldeq{Y}$ at a given layer of \gls{LLM}, subset $\mathcal{D}_{\mathrm{subset}}$, Candidate Set $\mathbb{S}$%$=\emptyset$
     \FOR{\camera{multiple $\boldeq{W}$ matrices in all layers}}
     \STATE \textcolor{gray}{\# Step 1: Forward Passes Averaged Over $\mathcal{D}_{\mathrm{subset}}$}
     \STATE $\boldeq{Y} =\sum_{i=1}^r\sigma_i\cdot(\boldeq{X}\boldeq{v}_i)\otimes\boldeq{u}_i^T = \sum_{i=1}^r\boldeq{Y}^{(i)} $
    \STATE \textcolor{gray}{\# Step 2: Search for $\boldeq{W}$ with High Dispersion}
     \STATE \textcolor{gray}{\# Angle Concentration is simplified as $\mathrm{AC}(\cdot)$}
     \STATE $\Delta_i  = \mathrm{AC}(\boldeq{Y}^{(i)})-\mathrm{AC}(\boldeq{Y})$ \textcolor{gray}{}
    \STATE $\mathbb{S} \leftarrow \mathbb{S} \cup  \camera{\left\{  \left (\boldeq{W}, -\sum_{i=r-m+1}^{r}\mathrm{min}(\Delta_i,0) \right )\right\}} $
   \ENDFOR
    \STATE \textcolor{gray}{\# Step 3: Rank Reduction for Evaluation}
     \STATE {\bfseries within}  {top-$j\,\boldeq{W}$ in $\mathbb{S}$}
     \STATE \hspace{0.8em} $\widehat{\boldeq{\Sigma}} \leftarrow$  keep top-$k\%$ ranks in 
    $\boldeq{\Sigma}$
    \STATE \hspace{0.8em}$\widehat{\boldeq{W}}=\boldeq{U}\widehat{\boldeq{\Sigma}}\boldeq{V}^T$
    \STATE Best $\widehat{\boldeq{W}} \leftarrow \left \{\mathrm{evaluate\,over\,validation\,datasets}  \right \} $
    \STATE {\bfseries Return:} \gls{LLM} with the rank-reduced $\widehat{\boldeq{W}}$
\end{algorithmic}
\end{algorithm}
\end{minipage}}
\vspace{-5pt}
\begin{align}
\boldeq{X}\boldeq{W}^T = \sum_{i=1}^r\sigma_i\cdot(\boldeq{X}\boldeq{v}_i)\otimes\boldeq{u}_i^T=\sum_{i=1}^r\boldeq{Y}^{(i)}=\boldeq{Y}, \label{eq:rank_forward}
\end{align}
where the \gls{SVD} of $\boldeq{W}$ is reformulated as multiple rank-$1$ components, i.e., $\boldeq{W}=\sum_{i=1}^r\sigma_{i}\boldeq{u}_i\otimes\boldeq{v}_i^T$.
\vspace{-5pt}
\subsection{Forward-Only Search}
\label{sec:forward-only-search}
We introduce \gls{SVD} components and their induced activations, where the activations are naturally accessible in the forward pass. More specifically, we propose to exploit the angle concentration of activations induced by each \gls{SVD} component. The angle concentration of activations induced by the $i$-th \gls{SVD} component, i.e., $\boldeq{Y}^{(i)}$, is denoted as follows: 
\begin{align}
\mathrm{Angle\,Concentration}(\boldeq{Y}^{(i)})= \frac{1}{n\times(n-1)}\sum_{q=1}^{n}\sum_{p=1}^n \frac{\boldeq{Y}_{q, :}^{(i)}{\boldeq{Y}_{p, :}^{(i)}}^T}{\left \| \boldeq{Y}_{q, :}^{(i)} \right \|\cdot \left \| \boldeq{Y}_{p, :}^{(i)} \right \|} \quad (q\neq p) \label{eq:angle_concentration}
\end{align}
where $\boldeq{Y}_{q, :}^{(i)} \in \mathbb{\mathbb{R}}^{1\times h}$ is the $q$-th row of $\boldeq{Y}^{(i)}$ and $\boldeq{Y}_{p, :}^{(i)} \in \mathbb{\mathbb{R}}^{1\times h}$ is the $p$-th row of $\boldeq{Y}^{(i)}$. Similarly, the angle concentration of the hidden states $\boldeq{Y}$ is given by $\mathrm{Angle\,Concentration}(\boldeq{Y})$, where $\boldeq{Y}$ is defined as the sum of activations induced by all \gls{SVD} components. 

We exploit the angle concentration of activations induced by each \gls{SVD} component $\boldeq{Y}^{(i)}$, \camera{as a heuristic indicator}, to measure its contribution---\textit{dispersed} or \textit{concentrated}---to the angle concentration of the hidden states $\boldeq{Y}$. We aim to identify the weight matrix with the highest dispersion that benefits most from rank reduction to promote greater concentration. Empirically, we show that \glspl{LLM} with angles becoming more concentrated, through selective rank reduction, achieve higher accuracy (Appendix \ref{sec:ac_acc}). Moreover, \cite{wang2025angles} also demonstrate that the angle between hidden states concentrates progressively when fine-tuning. Accordingly, making the angle between hidden states more concentrated may have the same effect as fine-tuning.\footnote{We also provide the intuition for the relationship between angle concentration of hidden states (available in the forward pass) and the gradients of singular values (computed in the backward pass) in Appendix \ref{sec:appendix_theory}. The gradients of singular values can be used to pinpoint which matrices merit reduction \citep{sreeram2025compress}, which avoids the exhaustive search that sweeps all the weights \citep{sharma2024the} but \cite{sreeram2025compress} requires backward passes.
} As shown in Fig. \ref{fig:main} \ding{174}, if the angle concentration of activations induced by one \gls{SVD} component $\boldeq{Y}^{(i)} $ is more dispersed than the angle concentration of $\boldeq{Y}$, namely, $\mathrm{Angle\,Concentration}(\boldeq{Y}^{(i)})-\mathrm{Angle\,Concentration}(\boldeq{Y})<0$, \textbf{the corresponding \gls{SVD} component tends to make $\boldeq{Y}$ more \textit{dispersed}, and can be removed}. In contrast, if the angle concentration of activations induced by one \gls{SVD} component $\boldeq{Y}^{(i)} $ is more concentrated than the angle concentration of $\boldeq{Y}$, namely, $\mathrm{Angle\,Concentration}(\boldeq{Y}^{(i)})-\mathrm{Angle\,Concentration}(\boldeq{Y})>0$, the corresponding \gls{SVD} component tends to make $\boldeq{Y}$ more \textit{concentrated}, and needs to be retained.

Next, we calculate the difference in angle concentration $\Delta_i$ between activations induced by the last-$m$-rank \gls{SVD} component and the hidden states $\boldeq{Y}$, as shown in Lines 7--8 of Algorithm~\ref{alg:main}. Then we sum these $\Delta_i$ only if $\Delta_i<0$ (i.e., the corresponding $i$-th \gls{SVD} component has the \textit{dispersed} effect). Different weight matrices have distinct levels of dispersion, in other words, distinct levels of necessity for rank reduction. We rank all weight matrices by the levels of dispersion and apply low-rank decomposition only for the top-$j$ weight matrices that have a high level of dispersion, avoiding exhaustive search. In the end, the rank-reduction strategy for retaining the top-$k\%$ ranks is determined by the accuracy evaluated on validation datasets, as presented in Appendix \ref{sec:appendix_efficient_hierarchical}.

\subsection{Efficient Forward-Only Approach}
We further analyze our proposed forward-only approach in terms of memory efficiency and computational efficiency. For memory efficiency, given peak memory consumption, we refer to the baseline (direct inference) as all layers' weight matrices $ \sum_1^L \left |  \boldeq{W}_l \right |$ plus the dynamic memory for maximum activations among $L$ layers, i.e., $ {\mathrm{max}}_{l\in \left \{  1,...,L\right \} } \left \{  \mathrm{max}(\left | \boldeq{X}_l  \right |,\left | \boldeq{Y}_l  \right |)\right \} $, based on the memory analysis in \cite{zhang2024revisiting}. Moreover, in dynamic memory, $\left |  \boldeq{W}_l\right |$ also captures the memory consumption of the corresponding weight gradient, and $\left |  \boldeq{\Sigma}_l\right |$ is for the gradient of singular values. Then we can have the peak memory consumption for the baseline (direct inference), LASER \citep{sharma2024the}, LASER-Grad \citep{sreeram2025compress}\footnote{Note that the backward pass typically consumes about twice the memory of the forward pass in \gls{CNN}-based networks, where activation memory is the primary bottleneck \citep{cai2020tinytl,tenison2025adabet}. For parameter-dominated networks, e.g., \Glspl{FC} \citep{huang2025tinyfoa} and \glspl{LLM} \citep{zhang2024revisiting} with Adam optimizer, the memory overhead of the backward pass is typically more than twice that of the forward pass.}, and our proposed forward-only search in Table~\ref{tab:memo_analysis}. Our proposed forward-only search is more memory-efficient than LASER-Grad \citep{sreeram2025compress} by avoiding the backward pass. The detailed analysis for \glspl{LLM} is derived in Appendix \ref{sec:appendix_peak_memory}. 

\begin{table}[!ht]
\vspace{-13pt}
\centering
  \caption{The peak memory consumption includes memory for the model (Weight Memory) and the dynamic allocation (Dynamic Memory) for the gradient (if applicable), the hidden states, the \gls{SVD} components and their induced activations. Here, we consider the baseline (direct inference), LASER \citep{sharma2024the}, LASER-Grad \citep{sreeram2025compress}, and our forward-only search.}
  \label{tab:memo_analysis}
  \centering
    \resizebox{\columnwidth}{!}{
  \begin{tabular}{ccl}
    \toprule
    Method     & Weight Memory     & Dynamic Memory \\
    \midrule
    Baseline & $ \sum_1^L \left |  \boldeq{W}_l \right |$ & $ 
    \underset{l\in \left \{  1,...,L\right \} }{\mathrm{max}}  \left(\mathrm{max}\left \{ \left | \boldeq{X}_l  \right |,\left | \boldeq{Y}_l  \right |\right \} \right)$\\
    LASER  & $ \sum_1^L \left |  \boldeq{W}_l \right |$  & $\underset{l\in \left \{  1,...,L\right \} }{\mathrm{max}} \left (\mathrm{max}(\left | \boldeq{X}_l  \right |,\left | \boldeq{Y}_l  \right |)+\left | \boldeq{U}_l \right | +\left | \boldeq{\Sigma}_l \right |+ \left | \boldeq{V}_l^T   \right |\right )$     \\
    LASER-Grad     & $ \sum_1^L \left |  \boldeq{W}_l \right |$  & $ \sum_{l=1}^{l=L} \mathrm{max}\left \{ \left | \boldeq{Y}_l  \right |+\left | \boldeq{U}_l \right | +\left | \boldeq{\Sigma}_l \right |+ \left | \boldeq{V}_l^T \right |, \left |  \boldeq{W}_l \right | + \left |  \boldeq{\Sigma}_l \right | \right \} $       \\
    Ours     & $ \sum_1^L \left |  \boldeq{W}_l \right |$     & $ \underset{l\in \left \{  1,...,L\right \} }{\mathrm{max}}  \left(\mathrm{max} \left \{\left | \boldeq{X}_l  \right |,\left | \boldeq{Y}_l  \right |+m\cdot\left | \boldeq{Y}_l  \right |\right \}+\left | \boldeq{U}_l \right | +\left | \boldeq{\Sigma}_l \right |+ \left | \boldeq{V}_l^T \right | \right)$ \\
    \bottomrule
  \end{tabular}}
\end{table}
\vspace{-7pt}
For computational efficiency, to avoid the extra computational complexity of Equation (\ref{eq:rank_forward}) and Equation (\ref{eq:angle_concentration}), we further propose an efficient and analytically equivalent forward-only search that includes an efficient forward pass and an efficient angle concentration, as documented in Appendix \ref{sec:appendix_efficient_search}. The computational complexity of our efficient forward pass for the last-$m$-rank \gls{SVD} component is negligible compared to the standard forward pass. Our efficient angle concentration also does not incur extra computational complexity. Overall, the computational complexity of our efficient forward-only search is on par with that of the standard forward pass (see Appendix \ref{sec:appendix_speedup_overall}). To further speed up our forward-only approach, we propose three variants of our rank-reduction strategy, namely, ours (subset), ours (std), and ours (hierarchical) by efficient hierarchical search (see Appendix \ref{sec:appendix_efficient_hierarchical}).
\vspace{-15pt}
\section{Experimental Results}
\vspace{-5pt}
\subsection{Experimental Setting}
\label{sec:setting}
Experiments are evaluated on CounterFact \citep{meng2022locating}, HotPotQA \citep{yang2018hotpotqa}, FEVER \citep{thorne2018fever}, Bios Gender and Bios Profession from Bias in Bios \citep{de2019bias}, TruthfulQA \citep{lin2022truthfulqa}, BigBench-Epistemic Reasoning \citep{bowman2015large}, and BigBench-WikidataQA \citep{wikidataqa,srivastava2023beyond}. We exploit GPT-J \citep{gpt-j} (GPT-J-6B) and Roberta
\citep{liu2019roberta} (RoBERTa$_\mathrm{BASE}$) with the weight matrices in the feed-forward network of Transformer architecture ($\boldeq{W}_{in/out}$) and the retained top-$k\%$ ranks are $0.5\%, 1\%, 5\%, 10\%, 20\%, 40\%, 60\%, 80\%$, and $90\%$, similar to the previous study \citep{sreeram2025compress}. For the variants of our rank-reduction strategy, we adopt $20\%$ of the whole dataset as $\mathcal{D}_{\mathrm{std}}$, and $100$ samples from $\mathcal{D}_{\mathrm{std}}$ as $\mathcal{D}_{\mathrm{subset}}$, to evaluate the weight matrices with rank reduction. For ours (hierarchical), we exploit the computational efficiency over $\mathcal{D}_{\mathrm{subset}}$ and then only evaluate the top-$j$ weight matrices over $\mathcal{D}_{\mathrm{std}}$ (see Appendix \ref{sec:appendix_efficient_hierarchical}). The accuracy is reported on $80\%$ of the whole dataset by the final candidate with rank reduction. There is no randomness for deterministic \gls{SVD} and we provide the means and standard deviations with multiple random seeds for the enhanced fine-tuning process. The main experiments utilize 1 NVIDIA Tesla A100 \gls{GPU} with 40GB \gls{RAM}. The details of datasets and \glspl{LLM} used in the main experiments are documented in Appendix \ref{sec:appendix_comparison}, Table \ref{tab:appendix_datasets} and Appendix \ref{sec:appendix_peak_memory}, Table \ref{tab:setting_llm}.

\subsection{Results}
\vspace{-5pt}
\label{sec:results}
We compare the baseline (direct inference), LASER \citep{sharma2024the}, LASER-Grad \citep{sreeram2025compress}, and our forward-only approach, in terms of accuracy (Acc), computational complexity (Speedup), and peak memory (Mem). We analyze the computational complexity for these methods in Appendix \ref{sec:appendix_comparison} and peak memory for these methods in Appendix \ref{sec:appendix_peak_memory}. The final candidates found by our forward-only approach on validation dataset, along with the justification for considering top-$j$ and last $m$ for GPT-J are presented in Appendix \ref{sec:appendix_para_gptj}, while those for Roberta are presented in Appendix \ref{sec:appendix_para_roberta}. 

\vspace{-5pt}
\begin{table}[!t]
  \caption{GPT-J evaluation with reported accuracy (Acc, \%), Speedup, and peak memory (Mem) in \gls{GB}. Our approach achieves the best average improvement in Acc and the best average Speedup. Moreover, our approach shows comparable Mem to Baseline and LASER \citep{sharma2024the}, and $1.9\times$ lower Mem to LASER-Grad \citep{sreeram2025compress} due to our forward-only passes. }
  \label{tab:gptj_main}
  \centering
  \resizebox{\columnwidth}{!}{
  {\setlength{\tabcolsep}{2pt}
  \begin{tabular}{lc>{\columncolor{mygreen!60}}cc>{\columncolor{mygreen!60}}c@{\hspace{5pt}}c>{\columncolor{myblue!60}}c>{\columncolor{mygreen!60}}c@{\hspace{5pt}}c>{\columncolor{myblue!60}}c>{\columncolor{mygreen!60}}c}
    \toprule
    \multirow{2}{*}{Dataset}     & \multicolumn{2}{c}{Baseline}     & \multicolumn{2}{c}{LASER (std)} & \multicolumn{3}{c}{LASER-Grad (std)} & \multicolumn{3}{c}{Ours (hierarchical)}\\
    \cmidrule(lr){2-3}
    \cmidrule(lr){4-5}
    \cmidrule(lr){6-8}
    \cmidrule(lr){9-11}
    &Acc&Mem$\downarrow$&Acc&Mem$\downarrow$&Acc&Speedup$\uparrow$&Mem$\downarrow$&Acc&Speedup$\uparrow$&Mem$\downarrow$\\
    \midrule
    CounterFact &  13.1 & 11.3 \gls{GB}& 24.0&11.5 \gls{GB}&\textbf{24.4}&1.99$\times$ &21.8 \gls{GB} &24.0&47.1$\times$& 11.5 \gls{GB}   \\
    HotPotQA     & 19.6  &11.3 \gls{GB}& 19.5&11.5 \gls{GB}&19.9&1.99$\times$ &21.9 \gls{GB} &\textbf{20.6}&38.4$\times$&11.5 \gls{GB}  \\
    FEVER     &  50.2 & 11.3 \gls{GB}& 56.2&11.5 \gls{GB}&56.0& 1.99$\times$&21.9 \gls{GB} &\textbf{58.4}&37.4$\times$&11.5 \gls{GB}  \\
    Bios Gender     & 70.9  & 11.3 \gls{GB}& \textbf{97.5}&11.5 \gls{GB}&88.4& 1.99$\times$&22.2 \gls{GB} &\textbf{97.5}&45.2$\times$&11.5 \gls{GB}   \\
    Bios Profession  & 75.6  &11.3 \gls{GB}& \textbf{82.1}&11.5 \gls{GB}&80.5&1.99$\times$ &22.2 \gls{GB} &79.1&40.7$\times$&11.5 \gls{GB}   \\
    TruthfulQA  & 54.9  & 11.3 \gls{GB}& 55.6&11.5 \gls{GB}&\textbf{56.1}& 1.99$\times$&21.9 \gls{GB} &55.7&28.4$\times$&11.5 \gls{GB}    \\
    Epistemic Reasoning  & 37.1  & 11.3 \gls{GB}& 38.3&11.5 \gls{GB}&62.3& 1.98$\times$&22.1 \gls{GB} &\textbf{62.8}&15.4$\times$&11.6 \gls{GB}   \\
    WikidataQA    & 51.8  & 11.3 \gls{GB}&65.9 & 11.5 \gls{GB}&\textbf{66.5}&1.99$\times$ &21.8 \gls{GB} &65.3&41.2$\times$&11.5 \gls{GB}  \\
    \midrule
    Average $\Delta$Acc$\uparrow$  & \multicolumn{2}{c}{0.00} &\multicolumn{2}{c}{8.24}&\multicolumn{3}{c}{10.1}&\multicolumn{3}{c}{\textbf{11.3}} \\
    \rowcolor{myblue!60} Average Speedup$\uparrow$ &\multicolumn{2}{c}{-}&\multicolumn{2}{c}{1$\times$}&\multicolumn{3}{c}{1.99$\times$}&\multicolumn{3}{c}{\textbf{36.7$\times$}}\\
    \rowcolor{mygreen!60} Average Mem$\downarrow$ &\multicolumn{2}{c}{11.3 \gls{GB}}&\multicolumn{2}{c}{ 11.5 \gls{GB}}&\multicolumn{3}{c}{22.0 \gls{GB}}&\multicolumn{3}{c}{11.5 \gls{GB}}\\
    \bottomrule
  \end{tabular}}}
\end{table}
\subsubsection{Lower Peak Memory \& Greater Speedup \& Competitive Accuracy}
We present the empirical results for GPT-J in Table \ref{tab:gptj_main} across various datasets. For GPT-J, our forward-only approach with efficient hierarchical search achieves higher average improvement in Acc, and greater average Speedup compared to LASER (std) and LASER-Grad (std). In terms of average Speedup, our approach (hierarchical) is 36.7$\times$ faster than LASER (std), and 18.4$\times$ faster than LASER-Grad (std). Meanwhile, our forward-only approach with efficient hierarchical search consumes comparable average Mem to Baseline (direct inference) and LASER, as well as 1.9$\times$ less average Mem to LASER-Grad. Moreover, our approach (hierarchical) achieves the best Acc on HotPotQA, FEVER, Bios Gender, and Epistemic Reasoning, i.e., four out of eight datasets. Especially for Epistemic Reasoning, our approach (hierarchical) achieves 62.8\% in Acc with 15.4$\times$ Speedup, which is a significant improvement than LASER. Taking Bios Gender as an example, our approach (hierarchical) has higher Acc, 22.7$\times$ Speedup, and 1.9$\times$ lower Mem, than LASER-Grad (std).

For Roberta, our proposed forward-only approach achieves greater average Speedup and comparable Acc compared to LASER (std) and LASER-Grad (std) (see Appendix \ref{sec:performence_on_roberta}, Table \ref{tab:roberta_main}). In terms of average Speedup, our approach is 26.6$\times$ faster than LASER (std), while LASER-Grad (std) is actually slower than LASER (std). For the Acc, our approach achieves comparable Acc to LASER (std) and LASER-Grad (std) for the majority of datasets, namely, six out of eight datasets. We notice a substantial difference for Epistemic Reasoning, a pattern that is similarly reflected in LASER-Grad (std) \citep{sreeram2025compress}. Therefore, we reproduce the Acc$^{\dagger}$ for LASER (std). Regarding Bios Profession, our approach achieves 40.5$\times$ Speedup and 1.7$\times$ lower Mem with a slight decrease in Acc than LASER-Grad (std). Despite this, our approach still demonstrates the effective adaptation on Bios Profession due to the improved Acc from the Baseline. Overall, our proposed forward-only approach achieves lower Mem and greater Speedup than LASER-Grad (std) \citep{sreeram2025compress}, as well as greater Speedup and competitive Acc compared to LASER (std) \citep{sharma2024the}.

\begin{figure}[!t]
\vspace{-10pt}
 \centering
  \begin{subfigure}{1\textwidth}\includegraphics[width=1.0\linewidth]{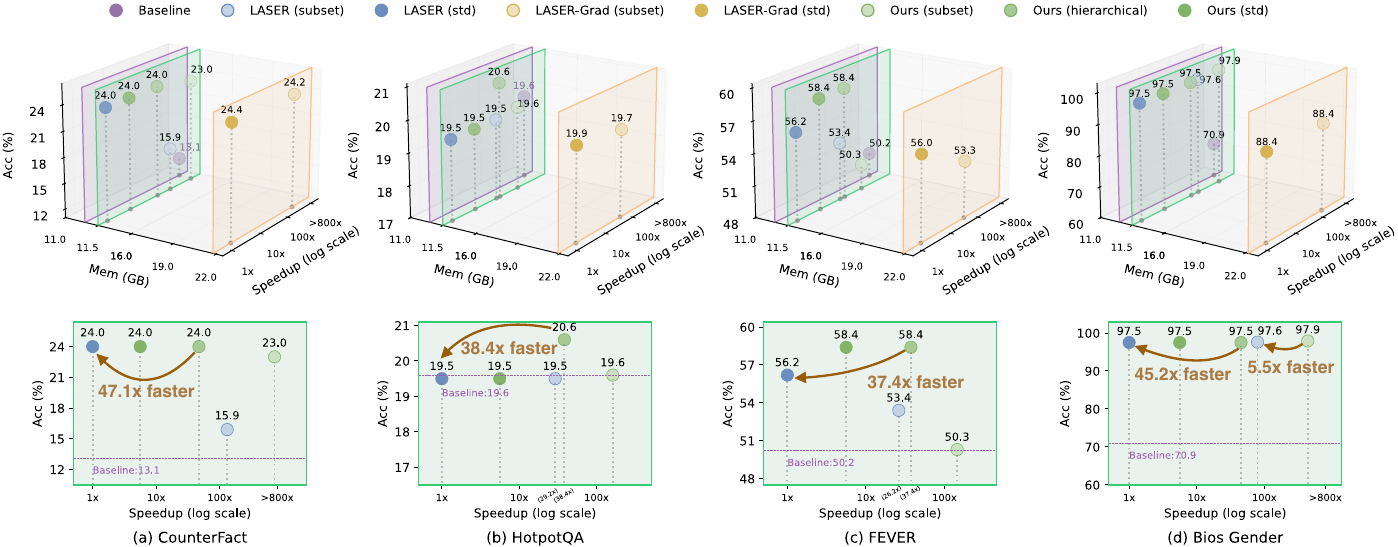}
    \end{subfigure}
  \begin{subfigure}{1\textwidth}\includegraphics[width=1.0\linewidth]{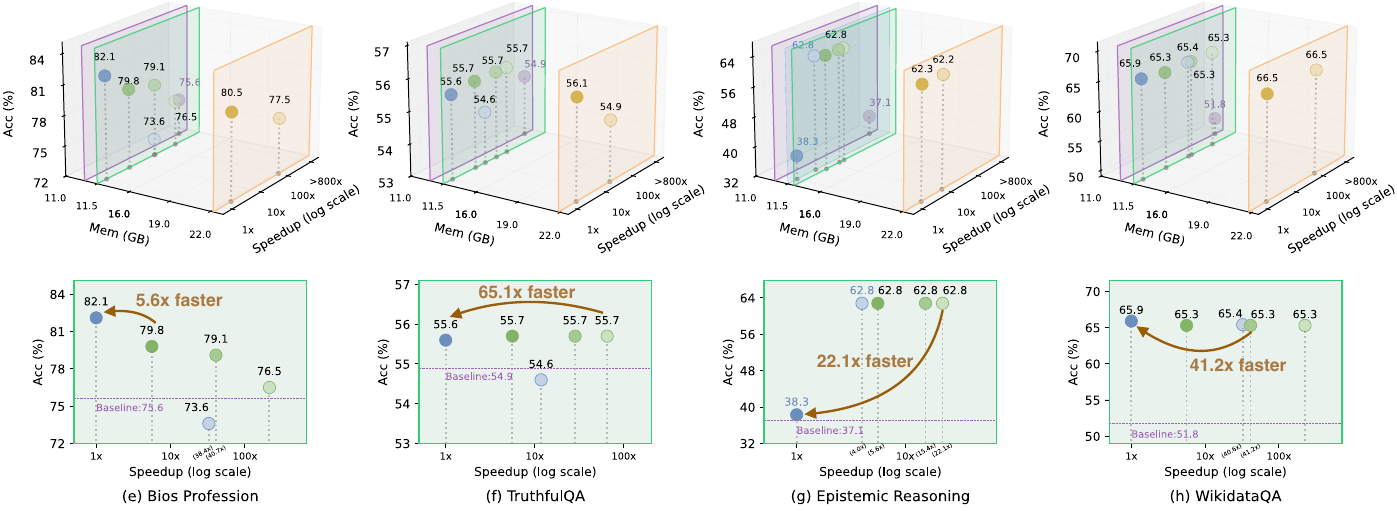}
    \end{subfigure}
  \caption{Ablation study on GPT-J: our forward-only approach consumes lower peak memory and greater speedup than LASER-Grad \citep{sreeram2025compress}, and achieves faster computation than LASER \citep{sharma2024the}, while maintaining comparable or even superior downstream accuracy.}
  \label{fig:compare_gptj}
  \vspace{-10pt}
\end{figure}

\subsubsection{Ablation Study}
We conduct empirical experiments for LASER (subset), LASER-Grad (subset), ours (subset), and ours (std) on GPT-J, to investigate the rank-reduction strategy for LASER \citep{sharma2024the}, LASER-Grad \citep{sreeram2025compress}, and our forward-only approach across different sizes of validation datasets \camera{and under the same compute}. As shown in Fig. \ref{fig:compare_gptj}, LASER-Grad (subset) and LASER-Grad (std) are located on the orange plane, which consumes more Mem than LASER and our approach. Moreover, our approach (subset) achieves faster average Speedup and greater average improvement in Acc than LASER-Grad (subset) (see Appendix \ref{sec:appendix_para_gptj}, Table \ref{tab:gptj_side}). Therefore, we unfold the green plane into two dimensions to compare LASER and our approach. Our approach (subset) always has faster Speedup than LASER (subset), and our approach (std) always has faster Speedup than LASER (std). Besides, our approach (subset) achieves better Acc than LASER (subset) on CounterFact, Bios Profession, TruthfulQA, and \camera{Bios Gender}, namely, four out of eight datasets. Our approach (std) also achieves comparable and even better Acc than LASER (std) on CounterFact, HotPotQA, FEVER, Bios Gender, TruthfulQA, Epistemic Reasoning and WikidataQA, namely, seven out of eight datasets. Overall, our forward-only approach achieves lower peak memory and greater speedup compared to LASER-Grad \citep{sreeram2025compress}, and achieves faster computation compared to LASER \citep{sharma2024the}, with comparable and sometimes even superior downstream performance. In addition, we also present the ablation study on Roberta in Table \ref{tab:roberta_side} (Mem in \gls{MB}) and the visualization in Appendix \ref{sec:appendix_para_roberta}, Fig. \ref{fig:compare_roberta}. Similarly, 
our forward-only approach achieves lower peak memory and greater speedup compared to LASER-Grad \citep{sreeram2025compress}, and achieves faster computation compared to LASER \citep{sharma2024the}, with comparable downstream performance.
\begin{table}[!t]
\vspace{-10pt}
  \caption{Roberta evaluation with reported accuracy (Acc, \%), Speedup, and peak memory (Mem).
  }
  \label{tab:roberta_side}
  \centering
  \resizebox{\columnwidth}{!}{
  {\setlength{\tabcolsep}{1pt}
  \begin{tabular}{lc>{\columncolor{myblue!60}}c>{\columncolor{mygreen!60}}c@{\hspace{5pt}}c>{\columncolor{myblue!60}}cc>{\columncolor{myblue!60}}cc>{\columncolor{myblue!60}}c>{\columncolor{mygreen!60}}c}
    \toprule
    & \multicolumn{3}{c}{\makecell[c]{\vspace{-16pt}LASER (subset)}}  &  \multicolumn{7}{c}{Ours} \\
    \cmidrule(lr){5-11}
    Dataset &&\cellcolor{white}&\cellcolor{white}&\multicolumn{2}{c}{(std)}&\multicolumn{2}{c}{(hierarchical)}&\multicolumn{2}{c}{(subset)} &\cellcolor{white}\\
    \cmidrule(lr){2-4}
    \cmidrule(lr){5-6}
    \cmidrule(lr){7-8}
    \cmidrule(lr){9-10}
    &Acc&Speedup$\uparrow$&Mem$\downarrow$&Acc&Speedup$\uparrow$&Acc&Speedup$\uparrow$&Acc&Speedup$\uparrow$&Mem$\downarrow$\\
    \midrule
    CounterFact &17.7&131$\times$&489 \gls{MB}&\textbf{19.2}&2.4$\times$&17.\camera{7} &40.4$\times$&17.7&617$\times$&493 \gls{MB}\\
    HotPotQA     &5.9&29.2$\times$&489 \gls{MB} &\textbf{6.3}&4.8$\times$&\textbf{6.3}&32.9$\times$&5.5&137$\times$&496 \gls{MB}\\
    FEVER      &52.4&26.2$\times$&489 \gls{MB}  &\textbf{52.5}&4.8$\times$&\textbf{52.5}&31.9$\times$&\textbf{52.5}&123$\times$&496 \gls{MB}\\
    Bios Gender  &92.2&79.3$\times$&495 \gls{MB} &\textbf{93.4}&4.8$\times$&\textbf{93.4}&38.7$\times$&93.2&372$\times$&514 \gls{MB}\\
    Bios Profession &70.9&38.4$\times$&495 \gls{MB} &\textbf{71.9}&4.8$\times$&\textbf{71.9}&34.8$\times$&70.9&180$\times$&514 \gls{MB}\\
    TruthfulQA  &\textbf{56.2}&11.8$\times$&490 \gls{MB} &55.6&4.8$\times$&56.0&24.2$\times$&56.0&55.2$\times$&498 \gls{MB}\\
    Epistemic Reasoning  &37.1&4.0$\times$&492 \gls{MB}&37.1&4.8$\times$&37.1&13.1$\times$&37.1&18.8$\times$&510 \gls{MB}\\
    WikidataQA  &30.7&40.6$\times$&489 \gls{MB} &30.8&4.8$\times$&30.8&35.2$\times$&\textbf{30.9}&191$\times$&492 \gls{MB}\\
    \midrule
    Average $\Delta$Acc$\uparrow$  & \multicolumn{3}{c}{2.05}&\multicolumn{2}{c}{\textbf{2.51}}&\multicolumn{2}{c}{2.\camera{38}}&\multicolumn{2}{c}{2.14}&\cellcolor{white} \\
    \rowcolor{myblue!60} Average Speedup$\uparrow$ & \multicolumn{3}{c}{45.1$\times$}&\multicolumn{2}{c}{4.5$\times$}&\multicolumn{2}{c}{31.4$\times$}&\multicolumn{2}{c}{\textbf{211.\camera{8}$\times$}}&\\
    \rowcolor{mygreen!60} Average Mem$\downarrow$ &\multicolumn{3}{c}{491 \gls{MB}}&\multicolumn{7}{c}{502 \gls{MB}} \\
    \bottomrule
  \end{tabular}}}
\end{table}

\subsubsection{Extended Experiments}
\label{sec:extend_exp}
\paragraph{The effectiveness and robustness of efficient hierarchical search.} 
The final candidates for GPT-J in Appendix \ref{sec:appendix_para_gptj}, Table \ref{tab:para_gptj} found by variants of our approach demonstrate that, there are five datasets having the \textit{Hit}, i.e., the final candidates found in ours (hierarchical) are the same as in ours (std). For the remaining three datasets without \textit{Hit}, our approach (hierarchical) achieves comparable accuracy to ours (std), as shown in Appendix \ref{sec:appendix_effect_hierarchical}, Table \ref{tab:gptj_no_hit}. The similar pattern is also observed for Roberta in Appendix \ref{sec:appendix_effect_hierarchical_roberta}.
These results show that the top-$j$ weight matrices with rank reduction achieving the top accuracies on $\mathcal{D}_{\mathrm{subset}}$ also tend to rank among the top performers on $\mathcal{D}_{\mathrm{std}}$. Furthermore, we try different sizes of $\mathcal{D}_{\mathrm{subset}}$ (from default $100$ to $10$, $5$, and $1$) and the results show the robustness of our efficient hierarchical search and that our proposed efficient hierarchical search can speed up the forward-only search at the extremely small size of $\mathcal{D}_{\mathrm{subset}}$, yielding the same final candidates as the default $\mathcal{D}_{\mathrm{subset}}$ (see Appendix \ref{sec:appendix_hierarchical_differentD_roberta}, Table \ref{tab:subset_roberta}). Therefore, our proposed efficient hierarchical search is effective and robust in speeding up the computation while keeping the downstream accuracy.
\begin{figure*}[!t]
  \centering
  \begin{minipage}{0.66\textwidth}
  \begin{subfigure}{0.5\textwidth}
    \includegraphics[width=\linewidth]{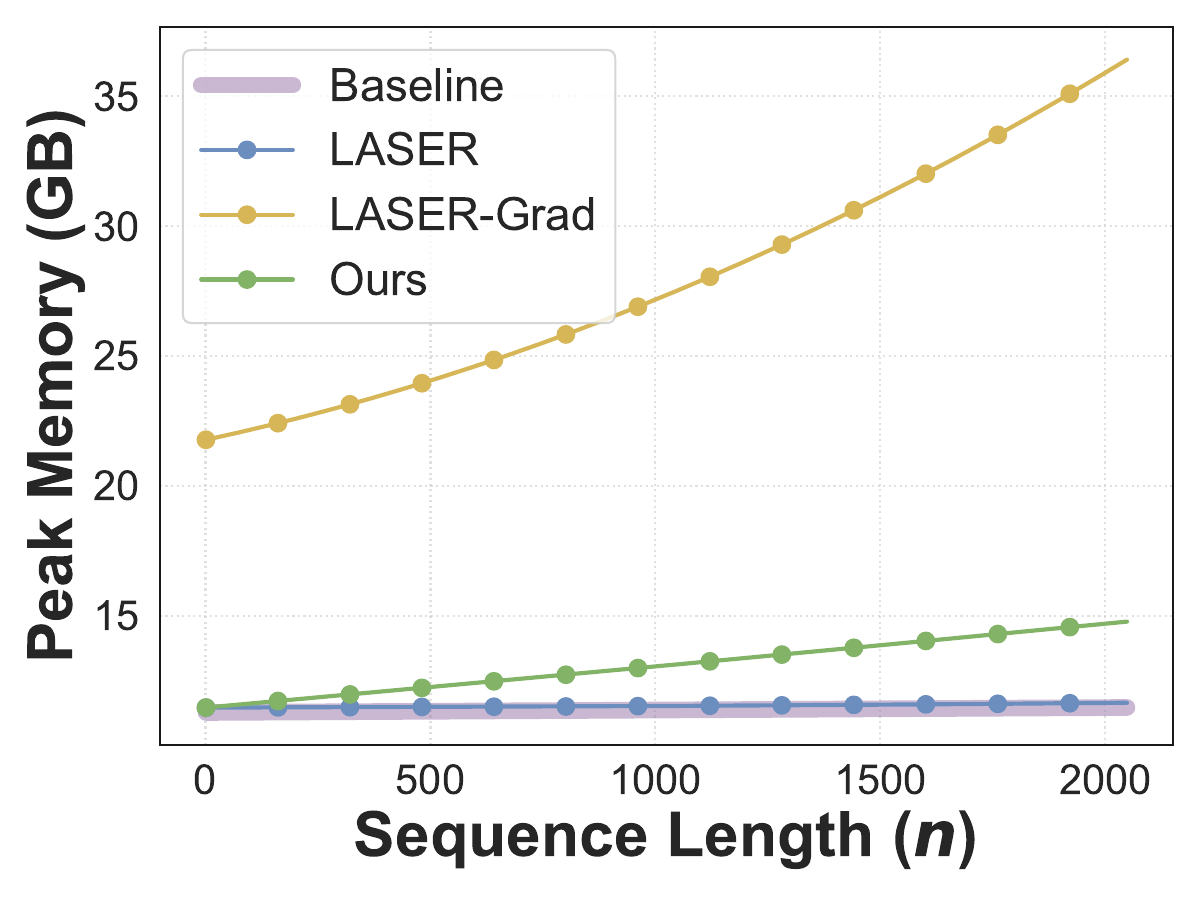}
  \end{subfigure}
  \hspace{-0.2cm}
  \begin{subfigure}{0.5\textwidth}
    \includegraphics[width=\linewidth]{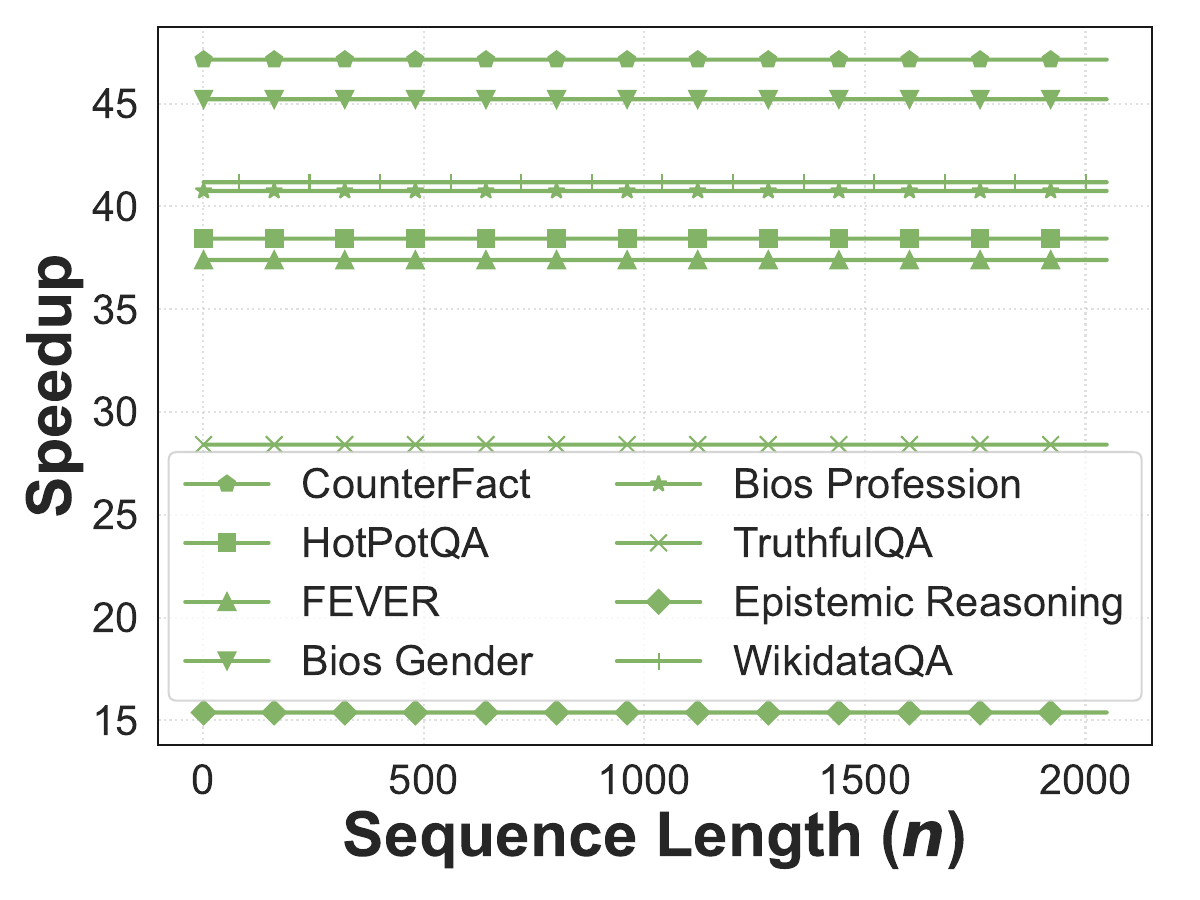}
  \end{subfigure}
  \captionsetup{skip=1pt}
    \caption*{(a) GPT-J}
  \end{minipage}
  \begin{minipage}{0.33\textwidth}
  \begin{subfigure}{\textwidth}
    \includegraphics[width=\linewidth]{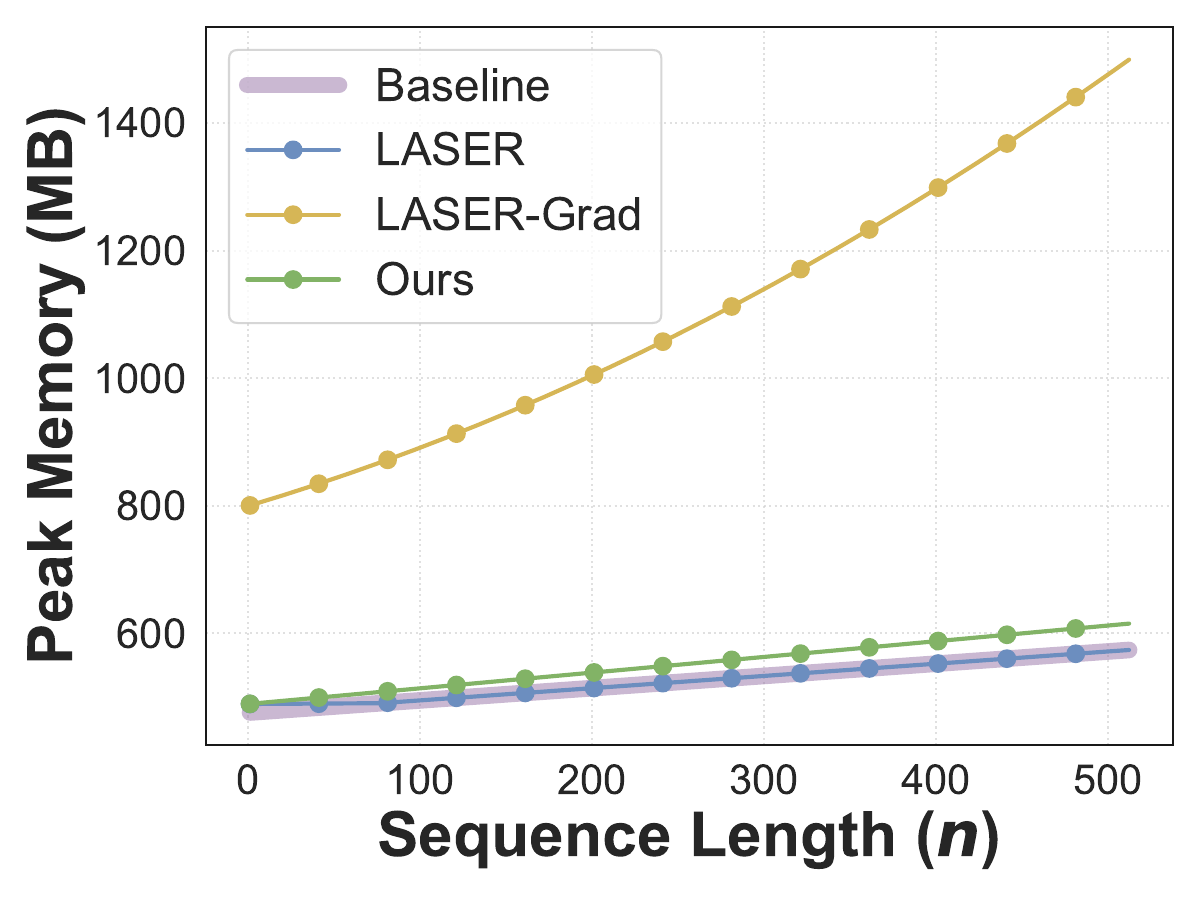}
  \end{subfigure}
  \captionsetup{skip=1pt}
  \caption*{(b) Roberta}
  \end{minipage}
  \vspace{-5pt}
  \caption{The peak memory with different sequence lengths $n$ for Baseline, LASER \citep{sharma2024the}, LASER-Grad \citep{sreeram2025compress}, and our approach, as well as the Speedup with different sequence lengths $n$ for our approach (hierarchical). The Speedup with different sequence lengths $n$ for Roberta is in Appendix \ref{sec:appendix_efficient_hierarchical}, Fig. \ref{fig:speedup_different}. Our forward-only approach requires substantially lower peak memory with greater Speedup than LASER-Grad across different sequence lengths $n$.}
  \label{fig:mem_different}
\end{figure*}
\paragraph{Resource efficiency across different sequence lengths.} To further investigate computational efficiency and memory efficiency of our forward-only approach, we evaluate the peak memory consumption with different sequence lengths $n$ for various methods, and show the Speedup with different sequence lengths $n$ for ours (hierarchical) in Fig. \ref{fig:mem_different}. For GPT-J in Fig. \ref{fig:mem_different} (a), the peak memory consumption for our approach increases with a small slope, whereas LASER-Grad exhibits substantial growth. 
The similar pattern is also observed for Roberta in Fig. \ref{fig:mem_different} (b) and Appendix \ref{sec:appendix_efficient_hierarchical}, Fig. \ref{fig:speedup_different}.
Furthermore, our forward-only approach achieves a greater Speedup as the dataset size of $\mathcal{D}$ increases (see Appendix \ref{sec:appendix_efficient_hierarchical}, Fig. \ref{fig:gptj_roberta_datasize}). Overall, our forward-only approach has significantly lower peak memory and greater Speedup than LASER-Grad across different sequence lengths.
  \vspace{-5pt}
\paragraph{Considering more types of weight matrices.} We add the weight matrices of key, query, value, and out projection to default $\boldeq{W}_{in/out}$. %to supplement previous studies \citep{sharma2024the,sreeram2025compress}
The corresponding results in Appendix \ref{sec:appendix_more_weight} show that our forward-only search can obtain the same top-$j$ weight matrices as when only considering default $\boldeq{W}_{in/out}$, for Roberta. In other words, the weight matrices in attention modules can be excluded from consideration to speed up the computation. Furthermore, the same top-$j$ weight matrices incur the same accuracy as \camera{ours (hierarchical) in Table \ref{tab:roberta_side}}, which further justifies the effectiveness of our forward-only search. Moreover, \cite{sharma2024the} present that the weight matrices of key, query, value, and out projection in the attention layers contribute to very little performance increase (GPT-J). Moreover, \cite{sreeram2025compress} demonstrate that the weight matrices of key and query can only reach the accuracy of Baseline (GPT-J). Therefore, considering that more weight matrices will increase the computation, it is typical to adopt only the weight matrices in the feed-forward network. 

\begin{wrapfigure}{r}{0.4\textwidth}
\vspace{-15pt}
  \centering
  \includegraphics[width=\linewidth]{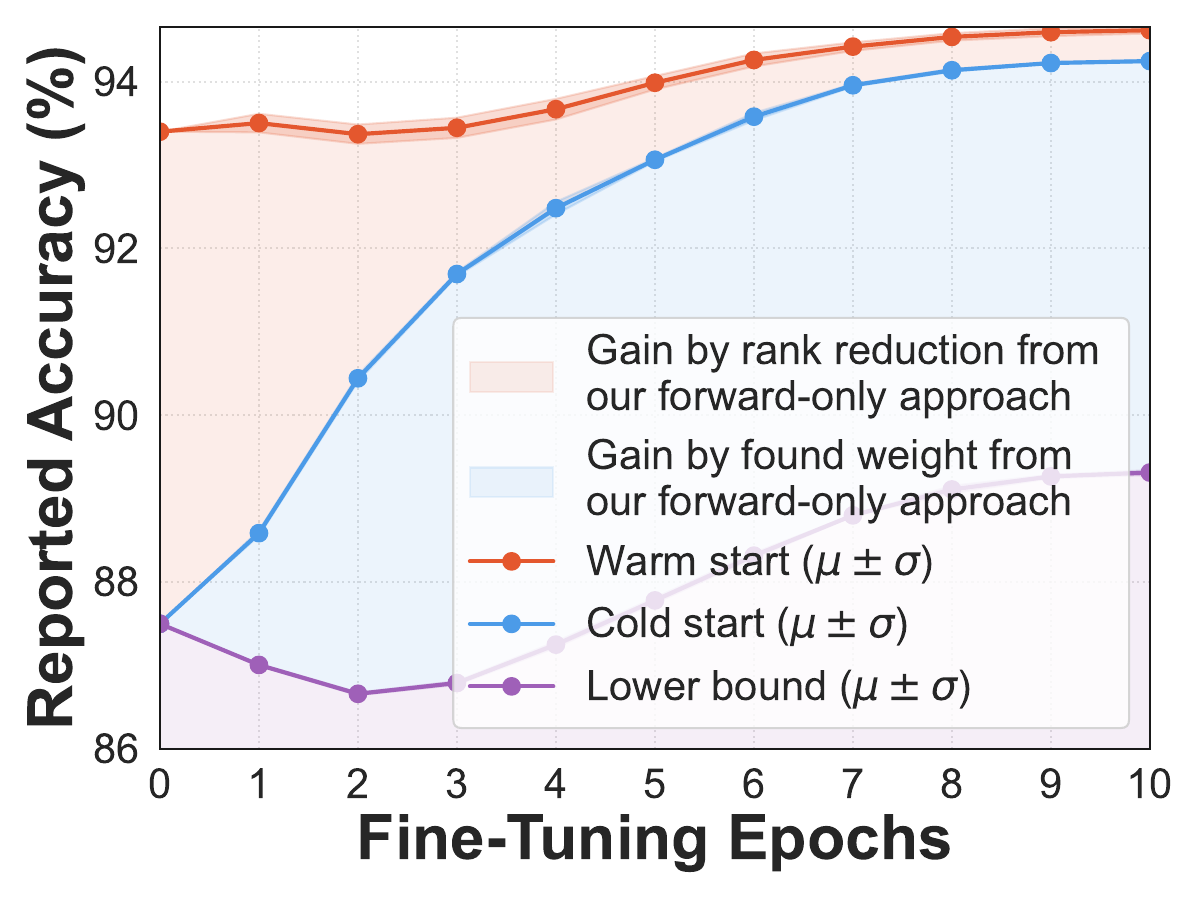}
  \caption{Final candidate with rank reduction from our forward-only approach enhances the fine-tuning process.}
  \label{fig:fine-tuning}
  \vspace{-10pt}
\end{wrapfigure}
\paragraph{Enhancing Fine-Tuning.} Our forward-only approach enhances fine-tuning rather than directly competing with it. Taking Roberta on Bios Gender as an example (Appendix \ref{sec:appendix_finetuning}), fine-tuning the final candidate found by our approach (only 1.89\% of the total parameters are trained) achieves the accuracy of 95.1\text{\scriptsize$\pm$0.01}\%, while fine-tuning all the $\boldeq{W}_{in}$ across twelve-layer Transformer modules (22.7\% of the total parameters are trained) achieves the accuracy of 95.6\text{\scriptsize$\pm$0.02}\%. Similarly, fine-tuning all the $\boldeq{W}_{out}$ achieves the accuracy of 95.6\text{\scriptsize$\pm$0.01}\%. Meanwhile, fine-tuning the worst candidate found by our approach (\textit{lower bound}) only gets the accuracy of 89.3\text{\scriptsize$\pm$0.03}\% in Appendix \ref{sec:appendix_finetuning}, Fig. \ref{fig:candidate_fine-tuning}. These results show that the final candidate is effective for fine-tuning and can achieve comparable accuracy with $12\times$ fewer trainable parameters than fine-tuning all $\boldeq{W}_{in}$ or all $\boldeq{W}_{out}$. Notably, obtaining the final candidate \camera{without rank reduction} from our approach does not require labels. Furthermore, in Fig. \ref{fig:fine-tuning}, the final candidate gives a warm start for effective fine-tuning, where the \textit{warm start} is fine-tuning the final candidate with rank reduction. Especially at the early fine-tuning epochs, the warm start performs better than the cold start (\textit{cold start} is directly fine-tuning the final candidate without rank reduction). More empirical results across datasets in Appendix \ref{sec:appendix_finetuning} consistently show that our forward-only approach enhances the fine-tuning process. In Appendix \ref{sec:appendix_weights_to_prompt}, we \textbf{transmute weight updates to steering prompts}, analytically and empirically showing that this transmutation achieves nearly identical accuracy without modifying \gls{LLM} parameters.

\begin{wraptable}{r}{0.5\textwidth}
\vspace{-12pt}
\resizebox{\linewidth}{!}{
\begin{threeparttable}
  \caption{The strong generalization of our forward-only approach across various \glspl{LLM} up to 57B.}
  \label{tab:otherllms}
  \centering
  \setlength{\tabcolsep}{2pt}
  \begin{tabular}{lccc>{\columncolor{myblue!60}}c}
    \toprule
    \multirow{2}{*}{\glspl{LLM}} & Baseline  & LASER & \multicolumn{2}{c}{Ours}  \\
    \cmidrule(lr){2-2}
    \cmidrule(lr){3-3}
    \cmidrule(lr){4-5}
    &Acc &Acc&Acc&Speedup \\
    \midrule
    LLaMA2-7B & 44.8 & 63.4&\textbf{63.5} &6.4$\times$\\
    %DeepseekV1-7B & 63.1 &\textbf{62.2}&61.4&58.4$\times$\\
    DeepseekV1-7B\tnote{${\ast}$} & 55.1 &61.4&\textbf{61.8}&29.6$\times$\\
    Mistral-7B &54.8 &\textbf{61.4}&\textbf{61.4}&9.1$\times$ \\
    Qwen2-7B\tnote{${\ast}$} &51.4 & \textbf{58.3}   & \textbf{58.3} &27.6$\times$ \\
    Gemma-7B &38.3 & 60.2 &\textbf{61.4}&11.1$\times$\\
    Gemma4-E2B-it &48.5 & \textbf{60.5} & 60.2 & 11.6$\times$ \\
    Qwen2-57B-A14B\tnote{${\ast}$} & 39.9&\textbf{62.1} &\textbf{62.1}& 13.9$\times$\\
    \midrule
    Average $\Delta$Acc$\uparrow$ & 0.00& 13.5 & \multicolumn{2}{c}{\textbf{13.7}}\\
    \rowcolor{myblue!60} Average Speedup$\uparrow$ &-&1$\times$&\multicolumn{2}{c}{\textbf{15.6$\times$}}\\
    \bottomrule
  \end{tabular}
  \begin{tablenotes}
\footnotesize
\item[]The ${\ast}$ indicates the chat version, i.e., DeepSeekV1-7B-Chat, Qwen2-7B-Instruct and Qwen2-57B-A14B-Instruct.
\end{tablenotes}
\end{threeparttable}}
\vspace{-10pt}
\end{wraptable}

\paragraph{Extension to various \glspl{LLM}.} In addition to Roberta (an encoder-only Transformer with bidirectional self-attention) and GPT-J (parallel scheme and \gls{RoPE})% and pre-norm Transformer)
, we also test LLaMA2 \citep{touvron2023llama} with gated feed-forward networks (Llama2-7B), Mistral \citep{jiang2023mistral7b} with \gls{GQA} and sliding-window attention (Mistral-7B), Qwen2 \citep{yang2024qwen2technicalreport} with instruction-tuning and multilingual capabilities (Qwen2-7B-Instruct), Qwen2 with \gls{MoE} (Qwen2-57B-A14B-Instruct), Gemma \citep{team2024gemma} (Gemma-7B), DeepSeek \citep{bi2024deepseek} (DeepSeekV1-7B-Chat) with data engineering, as well as the latest Gemma 4 \citep{google_gemma4_2026} (Gemma4-E2B-it) with the hybrid attention and per-layer embeddings. These extended \glspl{LLM}' performance on Epistemic Reasoning (knowledge and belief reasoning) is presented in Table \ref{tab:otherllms} with the found final candidates in Appendix \ref{sec:appendix_other_llm}, Table \ref{tab:para_otherllm}. Our approach (std) achieves greater speedup and competitive accuracy compared to LASER. Additionally, the results across datasets, including CounterFact (knowledge intervention), HotPotQA (multi-hop reasoning),   FEVER (fact verification), Bios Gender and Bios Profession (fairness and bias evaluation), TruthfulQA (hallucination detection), and BigBench-WikidataQA (information retrieval) with LLaMA2 (Appendix \ref{sec:appendix_other_llm}, Table \ref{tab:llama_main}) further verify the generalization of our forward-only approach.

\vspace{-10pt}
\camera{\paragraph{Additional Experiments and Analyses.}
The measurement of realistic memory consumption, computational complexity, and execution time of our approach and LASER-Grad \cite{sreeram2025compress} is presented in Appendix \ref{sec:emirical_laser_grad}. In addition, other simpler forward-only baselines are evaluated in Appendix \ref{sec:compare_simple_FO}. Moreover, more baselines in \gls{LLM} adaptation are presented, including tuning-based techniques (e.g., \gls{PEFT} in Appendix \ref{sec:appendix_finetuning}, Tables \ref{tab:finetuning_fullepoch} and \ref{tab:finetuning} but such methods require the backward pass, i.e., more memory), and the tuning-free method (e.g., \gls{ICL} in Appendix \ref{sec:compare_ICL}). Together with the enhanced fine-tuning in Fig. \ref{fig:fine-tuning} and steering prompts in Appendix \ref{sec:appendix_weights_to_prompt}, the corresponding results show the effectiveness of our approach compared with tuning-based and tuning-free adaptation methods. %Overall, all the results demonstrate the effectiveness and competitiveness of our forward-only approach.
}

\section{Related Work}

\paragraph{LLM Compression.} Model compression techniques are typically designed to reduce model storage without affecting the downstream accuracy \citep{han2016deepcompressioncompressingdeep}. To date, \gls{LLM} compression, including quantization \citep{dettmers2023qlora,xiao2023smoothquant,lin2025qserve}, distillation \citep{hinton2015distilling,sanh2019distilbert}, tensor decomposition \citep{maalouf2021deep}, pruning \citep{frantar2023sparsegpt,ashkboos2024slicegpt}, and KV cache compression \citep{zhang2023h2o,sheng2023flexgen,hooper2024kvquant}, also aim to maintain the downstream accuracy, but to some extent require extra steps such as fine-tuning \citep{dettmers2023qlora,ma2023llm}, alignment \citep{ma2023llm}, or enhancing the downstream accuracy by reasoning \citep{hsieh-etal-2023-distilling} after compression. Notably, the recent studies \citep{sharma2024the,sreeram2025compress} demonstrate that selectively removing higher-order components of weight matrices in \gls{SVD} can improve downstream accuracy.

\paragraph{LLM Adaptation.} Adapting \glspl{LLM} to downstream tasks typically involves tuning techniques and knowledge augmentation techniques. Tuning techniques include \gls{SFT} \citep{radford2019language,brown2020language}, \gls{PEFT} (such as \gls{LoRA} \citep{hu2022lora}, adapter tuning \citep{houlsby2019parameter}, etc.), prompt tuning \citep{lester2021power}, prefix-tuning \citep{li2021prefix}, \gls{BEFT} \citep{huang2026beft}, instruction tuning \citep{wang2023self} and the following alignment techniques \citep{ouyang2022training,rafailov2023direct,guo2025deepseek}, as well as test-time training \citep{sun2020test,gandelsman2022test,tandon2025end}. However, these tuning techniques usually rely on backward passes. Even the training-free alignment of \glspl{LLM} \citep{pan-etal-2025-survey} still incurs extra resource overhead compared to inference itself. Besides, knowledge augmentation techniques, including \gls{ICL} \citep{brown2020language}, prompt instruction \citep{ye2025neural}, retrieval-augmented generation \citep{lewis2020retrieval}, prompt engineering \citep{white2023prompt}, activation steering \citep{li2023inference}, and the tool-based approach \citep{schick2023toolformer} primarily operate by augmenting the prompt or injecting information and are orthogonal to modifying the internal weights during \gls{LLM} inference.

\paragraph{Forward-Only Learning.} The emerging forward-only algorithms \citep{frenkel2021learning,hinton2022forward,dellaferrera2022error,journe2023hebbian,jiang2024one,aminifar2024lightff,binaryFO,huang2025efficient,huang2025tinyfoa,sun2025deeperforward,Huang_FO_Survey_2026} reveal a promising trend in model adaptation from relying on backward passes to more directly leaning on forward passes, which is especially relevant in this era of unprecedented scale of \glspl{LLM}. The scalability of forward-only learning to deeper models (e.g., \glspl{LLM}) is still a challenging problem \citep{jiang2024one,mei2024survey}.  
Among forward-only techniques, the forward gradient \citep{baydin2022gradients} can be scaled to deeper models by either averaging over multiple perturbations or guided from local losses for a better gradient estimation \citep{ren2023scaling,xu2024fwdllm}, while the zeroth-order optimization can be scaled to deeper models \citep{chen2024deepzero} and \glspl{LLM} \citep{malladi2023fine,zhang2024revisiting,guo2025zerothorder}. 
Besides, the evolutionary strategies scaled to \glspl{LLM} either need to tune the input prompt/embedding \citep{sun2022black,prasad2023grips} or need several iterative (tuning) forward passes \citep{yang2024iterative}, similarly for \cite{niu2024testtime}.

\section{Conclusion and Future Work}
\label{sec:conclusion}

Amid the rapid growth and large-scale deployment of \gls{LLM} inference serving, we are motivated to rethink how the inference process itself can be leveraged to improve the serving system, particularly enabling efficient task-specific \gls{LLM} adaptation. In this paper, we propose a forward-only approach that efficiently identifies the weight matrix with the highest dispersion based on the angle concentration, and subsequently applies rank reduction for efficient \gls{LLM} adaptation with forward-only passes in \gls{LLM} inference serving. The results across diverse datasets demonstrate that our forward-only approach achieves lower peak memory and greater speedup than the gradient-based approach, as well as greater speedup and competitive accuracy compared to the exhaustive search approach. The experiments further present the robustness and generalization of our forward-only approach to various \glspl{LLM} with up to 57B parameters. This paper presents a new perspective on leveraging the inference process itself to improve \gls{LLM} serving by enabling efficient task-specific \gls{LLM} adaptation.

Future work includes investigating activation truncation rather than weight truncation \citep{qinsi2025dobisvd}, exploring dynamic adaptation \citep{mi2025layer,li2025adasvd} within the routed experts in \gls{MoE} \citep{yang2024qwen2technicalreport,dai2024deepseekmoe}, decomposing stacked weight matrices \citep{luo2025trawl}, combining forward-only adaptation with the post fine-tuning \citep{li2024basis,wang2025svdllm,wang2025milora,balazy2025lora,sun2026unleashing}, and studying the potential joint subspaces \citep{kaushik2026do}.

\section*{Acknowledgments and Disclosures}
\addcontentsline{toc}{section}{Acknowledgments and Disclosures}
\camera{This research has been partially supported by the Swedish Wallenberg AI, Autonomous Systems and Software Program (WASP), Swedish Research Council, Swedish Foundation for Strategic Research, ELLIIT Strategic Research Environment, and an unrestricted gift from Google. Computational resources were provided by the National Academic Infrastructure for Supercomputing in Sweden (NAISS), funded by the Swedish Research Council.}

\camera{Ananth Balashankar is an employee of Google DeepMind. The remaining authors declare no competing interests.}

\bibliographystyle{plainnat}
\bibliography{ref}

%%%%%%%%%%%%%%%%%%%%%%%%%%%%%%%%%%%%%%%%%%%%%%%%%%%%%%%%%%%%
\newpage
\appendix
\section*{Appendix}
\addcontentsline{toc}{section}{Appendix}
\tableofcontents

\glsresetall

\section{Final Candidates for Our Forward-Only Approach on GPT-J}
\label{sec:appendix_para_gptj}
\begin{table}[!ht]
  \caption{Final candidates for our forward-only approach on GPT-J in Table \ref{tab:gptj_main} and \ref{tab:gptj_side}, including $\boldeq{W}_{in/out}$ in the feed-forward network of Transformer architecture, the layer index $l$ ($0 \le  l \le 27$) of stacked Transformer modules, and the retained top-$k\%$ ranks ($0.5\% \le k\% \le 90\%$).}
  \label{tab:para_gptj}
  \centering
    \resizebox{\columnwidth}{!}{
  \begin{tabular}{lcccc}
    \toprule
    Dataset   & Ours (subset) & Ours (std)  &  Ours (hierarchical)    & Hit?   \\
    \midrule
    CounterFact & $[\boldeq{W}_{in},27,10\%]$ & $[\boldeq{W}_{in},27, 1\%]$ & $[\boldeq{W}_{in},27, 1\%]$ & Yes \\
    HotPotQA  &$[\boldeq{W}_{in}, 10, 80\%]$ & $[\boldeq{W}_{in}, 22, 80\%]$ & $[\boldeq{W}_{out}, 27, 1\%]$  & No \\
    FEVER  & $[\boldeq{W}_{in}, 10, 0.5\%]$ & $[\boldeq{W}_{in}, 10, 10\%]$ & $[\boldeq{W}_{in}, 10, 10\%]$   &  Yes \\
    Bios Gender&$[\boldeq{W}_{in},14,0.5\%]$ & $[\boldeq{W}_{in}, 14,1\%]$ &  $[\boldeq{W}_{in}, 14,1\%]$ & Yes \\
    Bios Profession& $[\boldeq{W}_{in}, 25,40\%]$& $[\boldeq{W}_{in}, 14, 1\%]$ & $[\boldeq{W}_{in}, 14, 10\%]$  & No \\
    TruthfulQA & $[\boldeq{W}_{out}, 27, 5\%]$ & $[\boldeq{W}_{out}, 27, 1\%]$ & $[\boldeq{W}_{out}, 27, 5\%]$  & No \\
    Epistemic Reasoning & $[\boldeq{W}_{in}, 4,0.5\%]$ & $[\boldeq{W}_{in}, 4, 1\%]$ & $[\boldeq{W}_{in}, 4, 1\%]$  & Yes \\
    WikidataQA & $[\boldeq{W}_{in}, 27, 0.5\%]$ & $[\boldeq{W}_{in}, 27, 0.5\%]$ & $[\boldeq{W}_{in}, 27, 0.5\%]$  & Yes \\
    \bottomrule
  \end{tabular}}
\end{table}

\subsection{Justification for considering \texorpdfstring{top-$j$}{top-j} in Algorithm \ref{alg:main}}
\label{sec:appendix_topj_gptj}
The final candidates for ours (hierarchical) with $j=10$ are presented in Table \ref{tab:para_gptj}. Here, we also test other top-$j$ values. When $j=5$, for all eight datasets except Epistemic Reasoning, the Acc for ours (hierarchical) with $j=5$ is lower than the Acc for ours (hierarchical) in Table \ref{tab:gptj_main}. Only for Epistemic Reasoning, the final candidate for ours (hierarchical) is the same as in Table \ref{tab:para_gptj}. Therefore, we do not adopt $j=5$. Moreover, when $j=7$, the final candidate for ours (hierarchical) on CounterFact is $[\boldeq{W}_{in},26,1\%]$ with the Acc of 17.3$\%$, which is worse than 24.0$\%$ for ours (hierarchical) in Table \ref{tab:gptj_main}. Similarly, for WikidataQA, the final candidate for ours (hierarchical) is $[\boldeq{W}_{out},27,0.5\%]$ with the Acc of 63.2$\%$, which is worse than 65.3$\%$ for ours (hierarchical) in Table \ref{tab:gptj_main}. In addition, the final candidate for ours (hierarchical) on HotPotQA is $[\boldeq{W}_{in},25,80\%]$ with the Acc of 19.5$\%$, and on TruthfulQA is $[\boldeq{W}_{out},27,5\%]$ with the Acc of 55.7$\%$, which are comparable to ours (hierarchical) in Table \ref{tab:gptj_main}. For the other four datasets (FEVER, Bios Gender, Bios Profession, Epistemic Reasoning), the final candidates for ours (hierarchical) are the same as in Table \ref{tab:para_gptj}. Overall, we adopt $j=10$ for GPT-J, while noting that smaller values of $j$ can yield higher Speedup (Equation (\ref{eq:our_hierarchical})) with comparable Acc on certain datasets.

\subsection{Justification for considering last \texorpdfstring{$m$}{m} \texorpdfstring{\gls{SVD}}{SVD} components in Algorithm \ref{alg:main}}
\label{sec:appendix_lastm_gptj}
\begin{figure*}[!ht]
  \centering
  \begin{subfigure}{0.328\textwidth}
    \includegraphics[width=\linewidth]{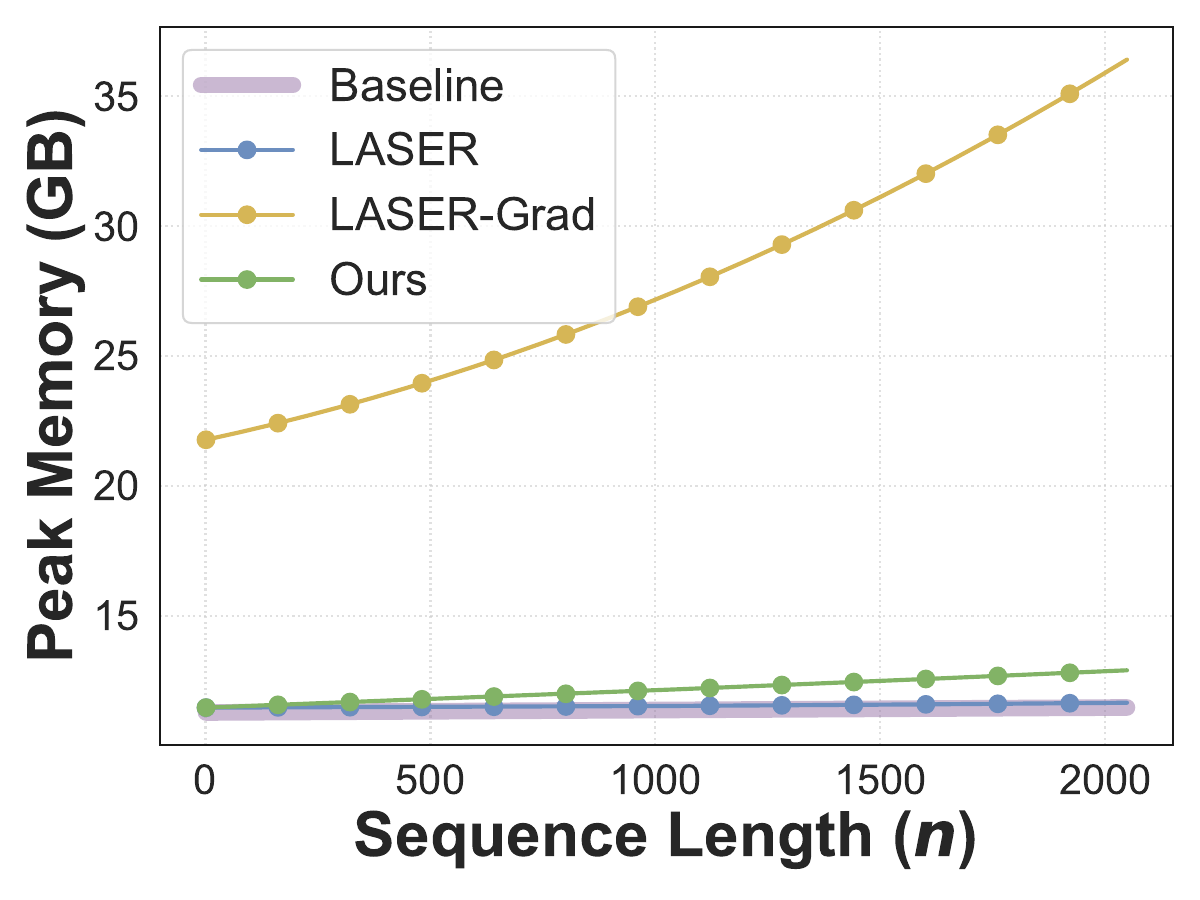}
    \caption{GPT-J ($m=20$)}
  \end{subfigure}
  \begin{subfigure}{0.328\textwidth}
    \includegraphics[width=\linewidth]{./img/gptj_mem_50_new.pdf}
    \caption{GPT-J ($m=50$)}
  \end{subfigure}
  \begin{subfigure}{0.328\textwidth}
    \includegraphics[width=\linewidth]{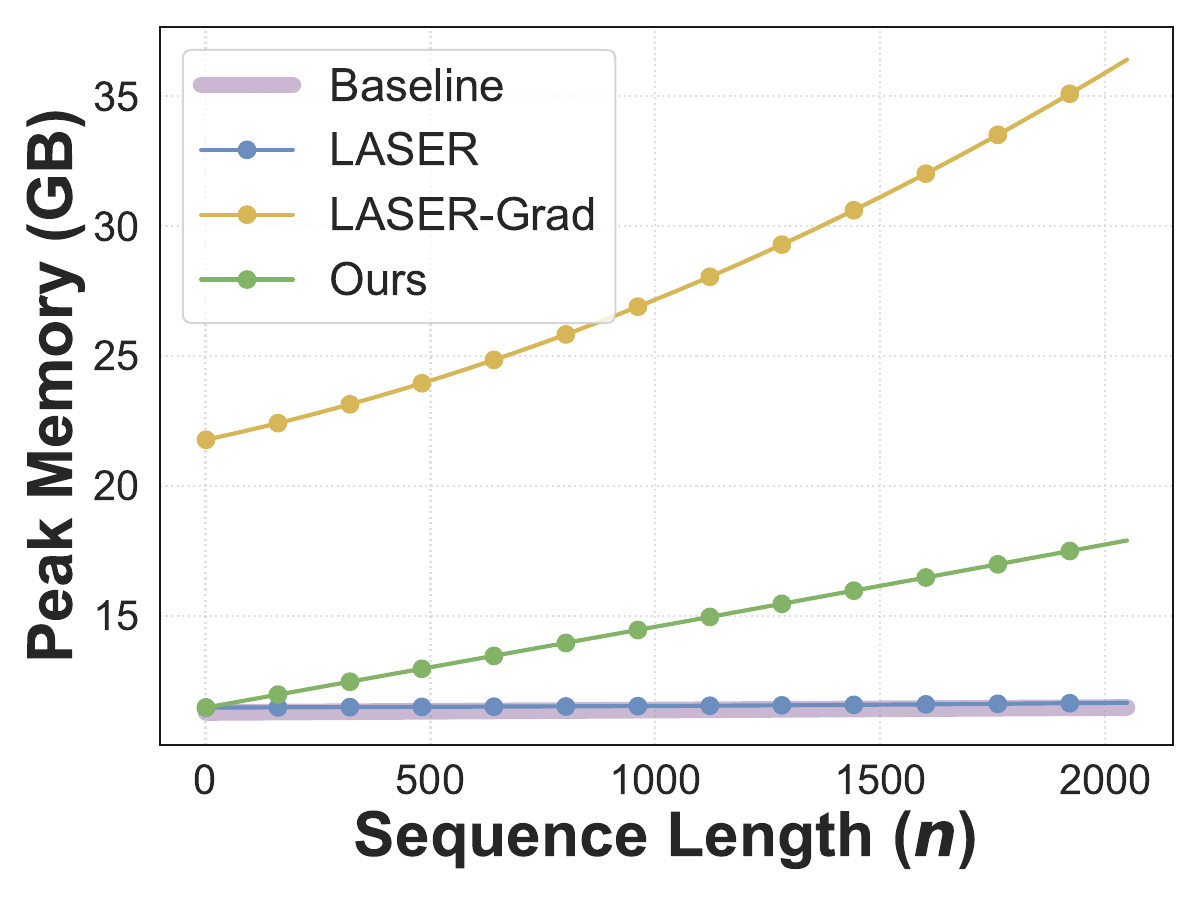}
    \caption{GPT-J ($m=100$)}
  \end{subfigure}
  \caption{The peak memory consumption for Baseline, LASER \cite{sharma2024the}, LASER-Grad \cite{sreeram2025compress}, and our forward-only approach with different sequence lengths $n$ and different last-$m$-rank \gls{SVD} components for GPT-J. Our forward-only approach requires significantly lower peak memory than LASER-Grad. The smaller $m$, the lower peak memory consumption.}
  \label{fig:mem_different_appendix}
\end{figure*}

First, we evaluate the peak memory consumption with different last-$m$-rank ($m=20,50,100$) \gls{SVD} components. As presented in Fig. \ref{fig:mem_different_appendix}, higher $m$ will increase the peak memory consumption of our approach, especially with the longer sequence length. Therefore, the smaller $m$, the better choice as long as the Acc is competitive. Note that the $m=20,50,100$ will not cause great differences in the ratio of our forward-only approach compared to the standard forward pass based on Equation (\ref{eq:ratio}); meanwhile, the $j$ value of top-$j$ weight matrices in Algorithm \ref{alg:main} will not influence the peak memory consumption of our approach.

Next, we evaluate different $m=20,50,100$ with $j=10$. When $m=20$, the Acc for ours (hierarchical) on five datasets (CounterFact, HotPotQA, FEVER, Epistemic Reasoning, and WikidataQA) are worse than the Acc for ours (hierarchical) in Table \ref{tab:gptj_main}. Meanwhile, the Acc for ours (hierarchical) on Bios Gender is 97.5$\%$, which is same as the Acc of LASER (std) \cite{sharma2024the} in Table \ref{tab:gptj_main}. Therefore, $m=20$ is enough for Bios Gender. For Bios Profession, the ACC for ours (hierarchical) from $m=20$ to $m=50$ is comparable; therefore, $m=20$ is adopted for Bios Profession. For the other \camera{six} datasets, we evaluate different $m=50,100$ and compare their final candidates for ours (hierarchical). For CounterFact, FEVER, Epistemic Reasoning, and WikidataQA, the final candidates for ours (hierarchical) with $m=50,100$ are the same. While for HotPotQA and TruthfulQA, although the final candidates for ours (hierarchical) are not the same among $m=50,100$, the Acc are comparable. Therefore, $m=50$ is adopted for these six datasets.

\subsection{The effectiveness of efficient hierarchical search}
\label{sec:appendix_effect_hierarchical}
As presented in Table \ref{tab:para_gptj}, we present the final candidates $[\boldeq{W},l,k\%]$ for variants of our approach. We select the top-$j\,[\boldeq{W},l]$ from our forward-only search, and then evaluate top-$j\,[\boldeq{W},l]$ with 9 different retained top-$k\%$ ranks in $\mathcal{D}_{\mathrm{subset}}$. Next, only the top-$j\,[\boldeq{W},l,k\%]$ from $\mathcal{D}_{\mathrm{subset}}$ are directly evaluated on $\mathcal{D}_{\mathrm{std}}$. The best $[\boldeq{W},l,k\%]$ from $\mathcal{D}_{\mathrm{std}}$ is the final candidate of ours (hierarchical). Therefore, if the top-$j\,[\boldeq{W},l,k\%]$ from $\mathcal{D}_{\mathrm{subset}}$ include the best $[\boldeq{W},l,k\%]$ from $\mathcal{D}_{\mathrm{std}}$, we call it \textit{Hit}. In GPT-J, there are 5 datasets (CounterFact, FEVER, Bios Gender, Epistemic Reasoning, and WikidataQA) that have \textit{Hit}, which means that ours (hierarchical) can get exactly the same final candidates with improving the evaluation efficiency (i.e., increases the Speedup as shown in Table \ref{tab:gptj_main} and Table \ref{tab:gptj_side}) compared to ours (std). On the contrary, for the datasets (HotPotQA, Bios Profession, TruthfulQA) without \textit{Hit}, the final candidates for ours (std) and ours (hierarchical) are different. However, as presented in Table \ref{tab:gptj_no_hit}, ours (hierarchical) can get comparable accuracy to ours (std) with greater Speedup.

\begin{table}[!ht]
  \caption{Acc (\%) and Speedup of ours (std) and ours (hierarchical) for datasets without \textit{Hit}.}
  \label{tab:gptj_no_hit}
  \centering
  \begin{tabular}{lc>{\columncolor{myblue!60}}cc>{\columncolor{myblue!60}}c}
    \toprule
    \multirow{2}{*}{Dataset}   & \multicolumn{2}{c}{Ours (std)} & \multicolumn{2}{c}{Ours (hierarchical)}     \\
    \cmidrule(lr){2-3}
    \cmidrule(lr){4-5}
    &Acc&Speedup$\uparrow$&Acc&Speedup$\uparrow$\\
    \midrule
    HotPotQA  & 19.5 & 5.6$\times$ &  20.6  & 38.4$\times$ \\
    Bios Profession& 79.8&5.6$\times$&  79.1&40.7$\times$   \\
    TruthfulQA & 55.7&5.6$\times$& 55.7  & 28.4$\times$  \\
    \bottomrule
  \end{tabular}
\end{table}

\newpage
\subsection{Various Rank-Reduction Strategies}
\vspace{-10pt}
\begin{table}[!ht]
  \caption{GPT-J evaluation with reported accuracy (Acc, \%), Speedup, and peak memory (Mem) in \gls{GB}. Ours (std) achieves the best average improvement in Acc; ours (subset) achieves the best average Speedup. Ours shows $1.9\times$ lower Mem to LASER-Grad due to our forward-only passes.}
  \label{tab:gptj_side}
  \centering
  \resizebox{\columnwidth}{!}{
  {\setlength{\tabcolsep}{1pt}
  \begin{tabular}{lc>{\columncolor{myblue!60}}c>{\columncolor{mygreen!60}}c@{\hspace{5pt}}c>{\columncolor{myblue!60}}c>{\columncolor{mygreen!60}}c@{\hspace{5pt}}c>{\columncolor{myblue!60}}c>{\columncolor{mygreen!60}}cc>{\columncolor{myblue!60}}c}
    \toprule
    \multirow{2}{*}{Dataset}  & \multicolumn{3}{c}{LASER (subset)} & \multicolumn{3}{c}{LASER-Grad (subset)} & \multicolumn{2}{c}{Ours (std)} &\cellcolor{white}& \multicolumn{2}{c}{Ours (subset)} \\
    \cmidrule(lr){2-4}
    \cmidrule(lr){5-7}
    \cmidrule(lr){8-12}
    &Acc&Speedup$\uparrow$&Mem$\downarrow$&Acc&Speedup$\uparrow$&Mem$\downarrow$&Acc&Speedup$\uparrow$&Mem$\downarrow$&Acc&Speedup$\uparrow$\\
    \midrule
    CounterFact &15.9&131$\times$&11.5 \gls{GB}& 24.2 &349$\times$&21.8 \gls{GB}&24.0&5.6$\times$&11.5 \gls{GB}&23.0&728$\times$\\
    HotPotQA &19.5&29.2$\times$&11.5  \gls{GB}&19.7 &77.5$\times$&21.9 \gls{GB}&19.5&5.6$\times$&11.5 \gls{GB}&19.6&162$\times$\\
    FEVER &53.4&26.2$\times$&11.5  \gls{GB}& 53.3&69.4$\times$&21.9 \gls{GB}&58.4&5.6$\times$&11.5 \gls{GB}&50.3&145$\times$\\
    Bios Gender &97.6&79.3$\times$&11.5  \gls{GB}&88.4 &210$\times$&22.2 \gls{GB}&97.5&5.6$\times$&11.5 \gls{GB}&97.9&439$\times$\\
    Bios Profession &73.6&38.4$\times$&11.5  \gls{GB}&77.5 &102$\times$&22.2 \gls{GB}&79.8&5.6$\times$& 11.5 \gls{GB}&76.5&213$\times$\\
    TruthfulQA &54.6&11.8$\times$&11.5 \gls{GB}&54.9 &31.2$\times$&21.9 \gls{GB}&55.7&5.6$\times$&11.5 \gls{GB}&55.7&65.1$\times$\\
    Epistemic Reasoning &62.8&4.0$\times$&11.5  \gls{GB}&62.2 &10.6$\times$&22.1 \gls{GB}&62.8&5.6$\times$&11.6 \gls{GB}&62.8&22.1$\times$\\
    WikidataQA &65.4&40.6$\times$&11.5  \gls{GB}& 66.5&108$\times$&21.8 \gls{GB}&65.3&5.6$\times$&11.5 \gls{GB}&65.3&225$\times$\\
    \midrule
    Average $\Delta$Acc$\uparrow$  & \multicolumn{3}{c}{8.7}&\multicolumn{3}{c}{9.2}&\multicolumn{2}{c}{\textbf{11.2}}&\cellcolor{white}&\multicolumn{2}{c}{9.7} \\
    \rowcolor{myblue!60} Average Speedup$\uparrow$ & \multicolumn{3}{c}{45.1$\times$}&\multicolumn{3}{c}{120$\times$}&\multicolumn{2}{c}{5.6$\times$}&&\multicolumn{2}{c}{\textbf{250$\times$}}\\
    \rowcolor{mygreen!60} Average Mem$\downarrow$ &\multicolumn{3}{c}{11.5  \gls{GB}}&\multicolumn{3}{c}{22.0 \gls{GB}}&\multicolumn{5}{c}{11.5 \gls{GB}} \\
    \bottomrule
  \end{tabular}}}
\end{table}

\subsection{Angle Concentration and LLM Adaptation}
\label{sec:ac_acc}
\begin{figure}[H]
 \centering
    \includegraphics[width=1.0\linewidth]{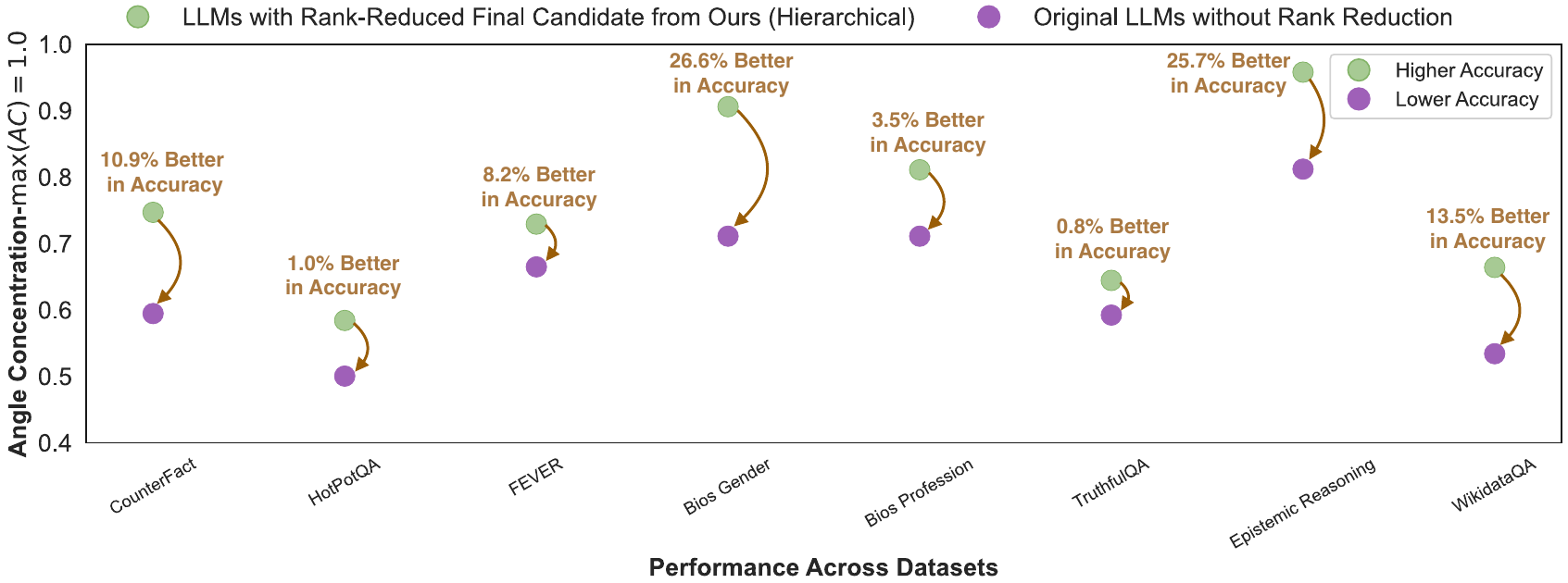}
  \caption{\glspl{LLM} with rank-reduced final candidate from our forward-only approach achieve higher downstream accuracy, with hidden-state angles becoming more concentrated.}
  \label{fig:ac_acc}
\end{figure}

We empirically measure the angle concentration of the hidden states in the corresponding layer, before rank reduction (purple dots) and after rank reduction (green dots) for the final candidate found by ours (hierarchical), where the maximum value of angle concentration is $1.0$ based on Equation \ref{eq:angle_concentration}. The results for GPT-J on eight datasets are presented in Fig. \ref{fig:ac_acc}, showing that the angle concentration of the hidden states after rank reduction is higher than that of the hidden states before rank reduction. Meanwhile, \glspl{LLM} with a higher angle concentration perform better with accuracy across datasets. Moreover, as mentioned in Section \ref{sec:forward-only-search}, the study \citep{wang2025angles} demonstrates that the angle between hidden states concentrates progressively when fine-tuning. Similarly to the fine-tuning–like effect, \glspl{LLM} with rank-reduced final candidate from our approach achieve higher downstream accuracy, with hidden-state angles becoming more concentrated.

\section{Final Candidates for Our Forward-Only Approach on Roberta}
\label{sec:appendix_para_roberta}

\begin{table}[!ht]
  \caption{Final candidates for our forward-only approach on Roberta in Table \ref{tab:roberta_side}, including $\boldeq{W}_{in/out}$ in the feed-forward network of Transformer architecture, the layer index $l$ ($0 \le  l \le 11$) of stacked Transformer modules, and the retained top-$k\%$ ranks ($0.5\% \le k\% \le 90\%$).}
  \label{tab:para_roberta}
  \centering
    \resizebox{\columnwidth}{!}{
  \begin{tabular}{lcccc}
    \toprule
    Dataset   & Ours (subset) & Ours (std)  &  Ours (hierarchical)    & Hit?   \\
    \midrule
    CounterFact & $[\boldeq{W}_{in},1,60\%]$ & $[\boldeq{W}_{in},8,80\%]$ &  $[\boldeq{W}_{in},1,60\%]$ & No \\
    HotPotQA  & $[\boldeq{W}_{out},11,60\%]$ & $[\boldeq{W}_{in},1,90\%]$  &  $[\boldeq{W}_{in},1,90\%]$ & Yes \\
    FEVER  & $[\boldeq{W}_{in},3,40\%]$ & $[\boldeq{W}_{in},3,40\%]$  & $[\boldeq{W}_{in},3,40\%]$  &  Yes\\
    Bios Gender& $[\boldeq{W}_{in},9,80\%]$ & $[\boldeq{W}_{in},9,90\%]$ & $[\boldeq{W}_{in},9,90\%]$  &  Yes \\
    Bios Profession& $[\boldeq{W}_{in},1,80\%]$ & $[\boldeq{W}_{in},3,90\%]$ & $[\boldeq{W}_{in},3,90\%]$  & Yes \\
    TruthfulQA & $[\boldeq{W}_{in},2,80\%]$ & $[\boldeq{W}_{in},3,80\%]$ & $[\boldeq{W}_{in},2,80\%]$  & No \\
    Epistemic Reasoning & $[\boldeq{W}_{in},5,0.5\%]$ & $[\boldeq{W}_{in},9,1\%]$  & $[\boldeq{W}_{in},9,0.5\%]$  &  No \\
    WikidataQA & $[\boldeq{W}_{out},11,1\%]$ & $[\boldeq{W}_{out},11,20\%]$ & $[\boldeq{W}_{out},11,20\%]$  &  Yes \\
    \bottomrule
  \end{tabular}}
\end{table}

\subsection{Justification for considering \texorpdfstring{top-$j$}{top-j} and last \texorpdfstring{$m$}{m} SVD components in Algorithm \ref{alg:main}} 
\label{sec:appendix_topj_lastm_roberta}

The final candidates for ours (hierarchical) with $j=5$ are presented in Table \ref{tab:para_roberta}. Here, we also test other top-$j$ values. When $j=7$, for all eight datasets except FEVER and TruthfulQA, the final candidates are the same as $j=5$. While for FEVER ($[\boldeq{W}_{in},2,40\%]$) and TruthfulQA ($[\boldeq{W}_{in},5,10\%]$), $j=7$ for ours (hierarchical) does not gain any improvement in Acc compared to $j=5$, but decreases the Speedup due to Equation (\ref{eq:our_hierarchical}). When $j=10$, for all eight datasets except HotPotQA and Epistemic Reasoning, the final candidates are the same as $j=7$. Similarly, while for HotPotQA ($[\boldeq{W}_{in},1,80\%]$) and Epistemic Reasoning ($[\boldeq{W}_{in},3,5\%]$), $j=10$ for ours (hierarchical) does not gain any improvement in Acc compared to $j=7$. Overall, we do not adopt $j=7,10$. Instead, $j=5$ is enough for ours (hierarchical).

We evaluate different $m=20,50,100$ with $j=5$. The results of $m=20$ are in Table \ref{tab:para_roberta}. When $m=50$, the final candidates for ours (hierarchical) on six datasets (CounterFact, HotPotQA, Bios Profession, Epistemic Reasoning, TruthfulQA, and WikidataQA) are the same as $m=20$ in Table \ref{tab:para_roberta}. While for FEVER and Bios Gender, $m=50$ for ours (hierarchical) does not gain any improvement in Acc. Similarly, there is no gain in Acc when $m=100$. Therefore, $m=20$ is adopted for Roberta.

\subsection{Performance on Roberta}
\label{sec:performence_on_roberta}
\begin{table}[H]
  \caption{Roberta evaluation with reported accuracy (Acc, \%), Speedup, and peak memory (Mem) in \gls{MB}. Ours achieves an average speedup of $26.6\times$, with comparable Acc and Mem to reproduced LASER${\dagger}$. Moreover, ours shows around $1.7\times$ lower Mem to LASER-Grad due to our forward-only passes. 
  }
  \label{tab:roberta_main}
  \centering
  \resizebox{\columnwidth}{!}{
  {\setlength{\tabcolsep}{2pt}
  \begin{threeparttable}
  \begin{tabular}{lc>{\columncolor{mygreen!60}}cc>{\columncolor{myblue!60}}c>{\columncolor{mygreen!60}}c@{\hspace{5pt}}c>{\columncolor{mygreen!60}}cc@{\hspace{5pt}}c>{\columncolor{myblue!60}}c>{\columncolor{mygreen!60}}c}
    \toprule
    \multirow{2}{*}{Dataset}     & \multicolumn{2}{c}{Baseline}  & \multicolumn{3}{c}{LASER-Grad (std)} & \multicolumn{3}{c}{LASER (std)} & \multicolumn{3}{c}{Ours}\\
    \cmidrule(lr){2-3}
    \cmidrule(lr){4-6}
    \cmidrule(lr){7-9}
    \cmidrule(lr){10-12}
    &Acc&Mem$\downarrow$&Acc&Speedup$\uparrow$&Mem$\downarrow$&Acc&Mem$\downarrow$&Acc\tnote{${\dagger}$}&Acc&Speedup$\uparrow$&Mem$\downarrow$\\
    \midrule
    CounterFact &  17.3 &479 \gls{MB}&19.3& 0.86$\times$&814 \gls{MB} & 19.3&489 \gls{MB}&19.2 &19.2&2.4$\times$&493 \gls{MB}   \\
    HotPotQA     & 6.1&481 \gls{MB}&6.5 & 0.86$\times$&824 \gls{MB}& 6.7&489 \gls{MB}&6.8 &6.3&32.9$\times$& 496 \gls{MB} \\
    FEVER     &  50.0 &481 \gls{MB}&52.7& 0.86$\times$&822 \gls{MB}& 52.3&489 \gls{MB}&51.9 &52.5&31.9$\times$& 496 \gls{MB}\\
    Bios Gender     & 87.5&495 \gls{MB} &93.1& 0.86$\times$&893 \gls{MB}&93.7&495 \gls{MB}&93.5 &93.4&38.7$\times$& 514 \gls{MB} \\
    Bios Profession     & 64.5& 495 \gls{MB}&75.1& 0.86$\times$& 891 \gls{MB}&72.5&495 \gls{MB}&71.6&71.9&34.8$\times$& 514 \gls{MB} \\
    TruthfulQA     &  56.2&483 \gls{MB}&56.3 & 0.85$\times$&830 \gls{MB}&56.2&490 \gls{MB}&55.7 &56.0&24.2$\times$& 498 \gls{MB}  \\
    Epistemic Reasoning     & 37.1&492 \gls{MB} &37.2& 0.85$\times$&877 \gls{MB}&41.8&492 \gls{MB}&37.3&37.1&13.1$\times$&510 \gls{MB} \\
    WikidataQA     & 28.0 &478 \gls{MB}&32.7& 0.86$\times$&809 \gls{MB}& 30.7&489 \gls{MB}&29.8 &30.8&35.2$\times$&492 \gls{MB} \\
    \midrule
    \rowcolor{myblue!60} Average Speedup$\uparrow$ &\multicolumn{2}{c}{-}&\multicolumn{3}{c}{0.86$\times$}&\multicolumn{3}{c}{1$\times$}&\multicolumn{3}{c}{\textbf{26.6$\times$}}\\
    \rowcolor{mygreen!60} Average Mem$\downarrow$ &\multicolumn{2}{c}{486 \gls{MB}}&\multicolumn{3}{c}{845 \gls{MB}}&\multicolumn{3}{c}{491 \gls{MB}}&\multicolumn{3}{c}{502 \gls{MB}}\\
    \bottomrule
  \end{tabular}
  \begin{tablenotes}
\footnotesize
\item[]The reproduced accuracy${\dagger}$ indicates a substantial difference for Epistemic Reasoning, a pattern similarly reflected in LASER-Grad (std) \cite{sreeram2025compress}.
\end{tablenotes}
\end{threeparttable}}}
\end{table}

\subsection{The effectiveness of efficient hierarchical search}
\label{sec:appendix_effect_hierarchical_roberta}

As presented in Table \ref{tab:para_roberta}, we present the final candidates $[\boldeq{W},l,k\%]$ for variants of our approach. We select the top-$j\,[\boldeq{W},l]$ from our forward-only search, and then evaluate top-$j\,[\boldeq{W},l]$ with 9 different retained top-$k\%$ ranks in $\mathcal{D}_{\mathrm{subset}}$. Next, only the top-$j\,[\boldeq{W},l,k\%]$ from $\mathcal{D}_{\mathrm{subset}}$ are directly evaluated on $\mathcal{D}_{\mathrm{std}}$. The best $[\boldeq{W},l,k\%]$ from $\mathcal{D}_{\mathrm{std}}$ is the final candidate of ours (hierarchical). Therefore, if the top-$j\,[\boldeq{W},l,k\%]$ from $\mathcal{D}_{\mathrm{subset}}$ include the best $[\boldeq{W},l,k\%]$ from $\mathcal{D}_{\mathrm{std}}$, we call it \textit{Hit}. In Roberta, there are 5 datasets (HotPotQA, FEVER, Bios Gender, Bios Profession, and WikidataQA) that have \textit{Hit}, which means that ours (hierarchical) can get exactly the same final candidates with improving the evaluation efficiency (i.e., increases the Speedup as shown in Table \ref{tab:roberta_side}) compared to ours (std). On the contrary, for the datasets (CounterFact, TruthfulQA, and Epistemic Reasoning) without \textit{Hit}, the final candidates for ours (std) and ours (hierarchical) are different. However, as presented in Table \ref{tab:roberta_side}, ours (hierarchical) can overall get comparable accuracy to ours (std) with greater Speedup.

\subsection{The robustness of efficient hierarchical search across different \texorpdfstring{$\mathcal{D}_{\mathrm{subset}}$}{subsets}}
\label{sec:appendix_hierarchical_differentD_roberta}
Consider the forward-only search for weight matrices, namely, $\underset{\mathrm{forward\text{-}only}}{\mathrm{ratio} \times  \mathcal{D}_{\mathrm{subset}}}$ in Equation (\ref{eq:our_hierarchical}), smaller size of $\mathcal{D}_{\mathrm{subset}}$ will lead to greater Speedup. Here, we try $\mathcal{D}_{\mathrm{subset}} = 10,5,1$ in addition to the default 100. As presented in Table \ref{tab:subset_roberta}, \ding{52} means the top-$j$ weight matrices are identical and preserve the same ordering as the weight matrices by default. \ding{52}$_{\mathrm{reordered}}$ means the top-$j$ weight matrices are identical but with a different order from the weight matrices by default. Taking CounterFact with $\mathcal{D}_{\mathrm{subset}} = 1$ as an example, the weight matrices are $ \left \{[\boldeq{W}_{out},11],[\boldeq{W}_{in},0],[\boldeq{W}_{in},1],[\boldeq{W}_{in},2],[\boldeq{W}_{in},3]\right \} $. Although the order is different from the default $ \left \{[\boldeq{W}_{out},11],[\boldeq{W}_{in},0],[\boldeq{W}_{in},1],[\boldeq{W}_{in},3],[\boldeq{W}_{in},2]\right \} $, the final candidate will not be influenced. Moreover, \ding{56}$\rightarrow$ \ding{52}$_{\mathrm{ours}}$ means that the top-$j$ weight matrices are not identical to the weight matrices by default, but get the same final candidates as in Table \ref{tab:para_roberta}, by ours (hierarchical). Taking Bios Gender with $\mathcal{D}_{\mathrm{subset}} = 1$ as an example, the weight matrices are $ \left \{[\boldeq{W}_{out},11],[\boldeq{W}_{in},0],[\boldeq{W}_{in},1],[\boldeq{W}_{in},9],[\boldeq{W}_{in},2]\right \} $. Although these weight matrices are different from the default $ \left \{[\boldeq{W}_{out},11],[\boldeq{W}_{in},0],[\boldeq{W}_{in},1],[\boldeq{W}_{in},9],[\boldeq{W}_{in},3]\right \} $, ours (hierarchical) still selects the $[\boldeq{W}_{in},9,90\%]$ as the final candidate (same as in Table \ref{tab:para_roberta}). Overall, ours (hierarchical) can speed up the forward-only search at the extreme case at $\mathcal{D}_{\mathrm{subset}}=1$, yielding the same final candidates at $\mathcal{D}_{\mathrm{subset}}=100$. Our forward-only search with efficient hierarchical search is robust to the extremely small size of $\mathcal{D}_{\mathrm{subset}}$.

Another interesting observation is that from Table \ref{tab:subset_roberta}, the first two weight matrices, i.e., $ \left \{[\boldeq{W}_{out},11],[\boldeq{W}_{in},0]\right \} $, are the same for different datasets with Roberta. However, as presented in Table \ref{tab:para_roberta}, only WikidataQA selects one of these two weight matrices as the final candidate for ours (hierarchical). Instead, for other seven datasets, different final candidates are selected for the best Acc. Inspired by the recent study \citep{kaushik2026do}, the potential joint-candidates (such as $ [\boldeq{W}_{out},11] $ or $[\boldeq{W}_{in},0] $) for Roberta on various datasets are left for future investigation.

\begin{table}[!ht]
  \caption{Weight matrices for our forward-only search on Roberta with $j=5$ and $m=20$, with different size of $\mathcal{D}_{\mathrm{subset}} = 100\,(\mathrm{default}),10,5,1$. Our forward-only search with efficient hierarchical search is robust to the extremely small size of $\mathcal{D}_{\mathrm{subset}}$.}
  \label{tab:subset_roberta}
  \centering
    \resizebox{\columnwidth}{!}{
  \begin{tabular}{lcccc}
    \toprule
    Size of $\mathcal{D}_{\mathrm{subset}}$   & 100 & 10  &  5    & 1  \\
    \midrule
    CounterFact & $ \left \{[\boldeq{W}_{out},11],[\boldeq{W}_{in},0],[\boldeq{W}_{in},1],[\boldeq{W}_{in},3],[\boldeq{W}_{in},2]\right \} $ & \ding{52} & \ding{52} & \ding{52}$_{\mathrm{reordered}}$ \\
    HotPotQA   & $ \left \{[\boldeq{W}_{out},11],[\boldeq{W}_{in},0],[\boldeq{W}_{in},1],[\boldeq{W}_{in},3],[\boldeq{W}_{in},2]\right \} $ & \ding{52} & \ding{52} & \ding{52}\\
    FEVER   & $ \left \{[\boldeq{W}_{out},11],[\boldeq{W}_{in},0],[\boldeq{W}_{in},1],[\boldeq{W}_{in},3],[\boldeq{W}_{in},6]\right \} $ & \ding{52}$_{\mathrm{reordered}}$ & \ding{52} &  \ding{52}\\
    Bios Gender & $ \left \{[\boldeq{W}_{out},11],[\boldeq{W}_{in},0],[\boldeq{W}_{in},1],[\boldeq{W}_{in},9],[\boldeq{W}_{in},3]\right \} $ & \ding{56}$\rightarrow$ \ding{52}$_{\mathrm{ours}}$ &  \ding{56}$\rightarrow$ \ding{52}$_{\mathrm{ours}}$& \ding{56}$\rightarrow$ \ding{52}$_{\mathrm{ours}}$\\
    Bios Profession & $ \left \{[\boldeq{W}_{out},11],[\boldeq{W}_{in},0],[\boldeq{W}_{in},1],[\boldeq{W}_{in},9],[\boldeq{W}_{in},3]\right \} $ & \ding{52}$_{\mathrm{reordered}}$  & \ding{52}  &  \ding{56}$\rightarrow$ \ding{52}$_{\mathrm{ours}}$ \\
    TruthfulQA  & $ \left \{[\boldeq{W}_{out},11],[\boldeq{W}_{in},0],[\boldeq{W}_{in},1],[\boldeq{W}_{in},2],[\boldeq{W}_{in},3]\right \} $ & \ding{52}$_{\mathrm{reordered}}$ & \ding{52}$_{\mathrm{reordered}}$ & \ding{52}$_{\mathrm{reordered}}$\\
    Epistemic Reasoning  & $ \left \{[\boldeq{W}_{out},11],[\boldeq{W}_{in},0],[\boldeq{W}_{in},1],[\boldeq{W}_{in},9],[\boldeq{W}_{in},5]\right \} $ & \ding{52} & \ding{52} &  \ding{56}$\rightarrow$ \ding{52}$_{\mathrm{ours}}$ \\
    WikidataQA  & $ \left \{[\boldeq{W}_{out},11],[\boldeq{W}_{in},0],[\boldeq{W}_{in},1],[\boldeq{W}_{in},3],[\boldeq{W}_{in},2]\right \} $ & \ding{52} &  \ding{56}$\rightarrow$ \ding{52}$_{\mathrm{ours}}$ &  \ding{56}$\rightarrow$ \ding{52}$_{\mathrm{ours}}$ \\
    \bottomrule
  \end{tabular}}
\end{table}
\newpage
\subsection{Various Rank-Reduction Strategies}

\begin{figure}[!ht]
 \centering
  \begin{subfigure}{1\textwidth}\includegraphics[width=1.0\linewidth]{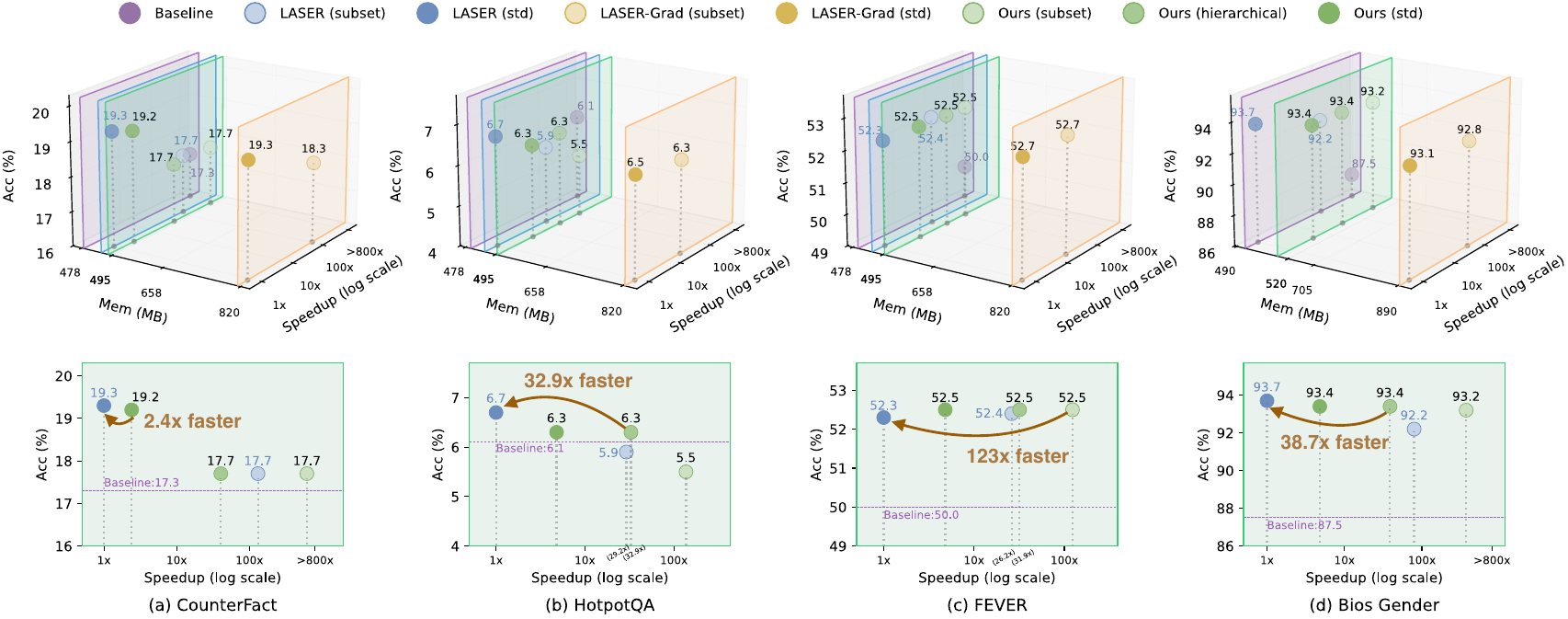}
    \end{subfigure}
  \begin{subfigure}{1\textwidth}\includegraphics[width=1.0\linewidth]{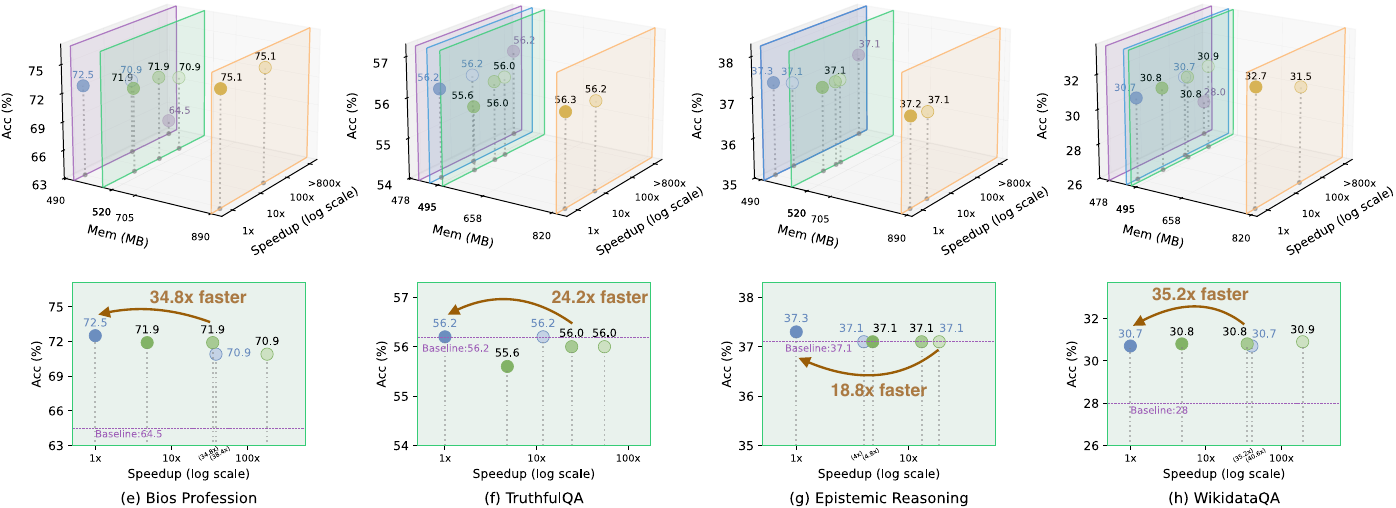}
    \end{subfigure}
  \caption{
  Ablation study on Roberta: our forward-only approach consumes lower peak memory compared to LASER-Grad \citep{sreeram2025compress}, and achieves faster computation compared to LASER \citep{sharma2024the}\protect\footnotemark, with comparable downstream accuracy.}
  \label{fig:compare_roberta}
\end{figure}
\footnotetext{For Epistemic Reasoning, the reproduced accuracy${\dagger}$ in Table \ref{tab:roberta_main}, is exploited.}

\section{Speedup in Computational Complexity}
\label{sec:appendix_speedup}

In this section, we will calculate the ratio of our efficient and analytically equivalent forward-only search compared to the standard forward pass in terms of computational complexity. We refer to \gls{FLOPs} as a measure of theoretical computational complexity. Since the inference with large batch sizes in \glspl{LLM}, especially during the prefill stage (similarly with next-token prediction for GPT-J and masked language modeling for Roberta), is dominated by compute-bound operations (e.g., large matrix multiplications), \gls{FLOPs} are expected to correlate well with computational time in practice. Accordingly, the \gls{FLOPs} of standard forward pass ($\boldeq{X}\boldeq{W}^T=\boldeq{Y}$) are $2nhd$.

\subsection{Efficient Forward-Only Search}
\label{sec:appendix_efficient_search}
\paragraph{Efficient forward pass.}For Equation (\ref{eq:rank_forward}) in Algorithm \ref{alg:main}, instead of considering all \gls{SVD} components in one forward pass, we can only consider activations induced by the last-$m$-rank \gls{SVD} components ($\boldeq{Y}^{(r-m+1)}$ to $\boldeq{Y}^{(r)}$) in one efficient forward pass, plus the hidden states $\boldeq{Y}$ in one standard forward pass. In the efficient forward pass, the last-$m$-rank \gls{SVD} components can be operated with batched outer product, as shown below: 
\begin{equation}
\scalebox{0.9}{$(\boldeq{X}\boldeq{V}_{:,r-m+1:r}\boldeq{\Sigma}_{r-m+1:,r-m+1:})[:,\mathrm{newaxis},:]\odot\boldeq{U}_{:,r-m+1:r}[\mathrm{newaxis},:,:]= [\boldeq{Y}^{(r-m+1)},...,\boldeq{Y}^{(r)}]$}, \label{eq:one_forward_with_m_batch}
\end{equation}
where  $\boldeq{\Sigma}_{r-m+1:,r-m+1:}=\mathrm{diag}(\sigma_{r-m+1},\sigma_{r-m+2},...,\sigma_r) \in \mathbb{\mathbb{R}}^{m\times m}  $, $\boldeq{V}_{:,r-m+1:r} = [\boldeq{v}_{r-m+1},\boldeq{v}_{r-m+2},...,\boldeq{v}_r] \in \mathbb{\mathbb{R}}^{d\times m} $, and $\boldeq{U}_{:,r-m+1:r} = [\boldeq{u}_{r-m+1},\boldeq{u}_{r-m+2},...,\boldeq{u}_r] \in \mathbb{\mathbb{R}}^{h\times m} $. $\odot$ denotes the Hadamard (element-wise) product, where the elements are broadcasted before/after $\odot$, and therefore $[\boldeq{Y}^{(r-m+1)},...,\boldeq{Y}^{(r)}] \in \mathbb{\mathbb{R}}^{n\times h\times m} $. For each \gls{SVD} component, this element-wise product with broadcasting is equivalent to the Kronecker product in Equation (\ref{eq:rank_forward}). %As presented in Appendix-\ref{sec:appendix_speedup_in_efficient_fo} and Fig. \ref{fig:efficient}, the computational complexity of our efficient forward pass for the last $m$ \gls{SVD} component is negligible compared to the standard forward pass. 

\paragraph{Efficient angle concentration.}For Equation (\ref{eq:angle_concentration}), given $\widetilde{\boldeq{Y}}_{q, :}^{(i)}=\frac{\boldeq{Y}_{q, :}^{(i)}}{\left \|\boldeq{Y}_{q, :}^{(i)}\right \|}$ and $\widetilde{\boldeq{Y}}_{p, :}^{(i)}=\frac{\boldeq{Y}_{p, :}^{(i)}}{\left \|\boldeq{Y}_{p, :}^{(i)}\right \|}$, by exploiting the identity that the sum of pairwise cosine similarities \camera{(excluding self-pairs)} equals the squared norm of the summed normalized directions \camera{minus $n$}, as well as considering $\left \|\widetilde{\boldeq{Y}}_{q, :}^{(i)}\right \|^2 =1$, the angle concentration for the hidden states of each \gls{SVD} component can be reformulated as follows ($p \neq q$ \camera{and $n \ge 2$}): 
\begin{equation}
\scalebox{0.89}{$
 \mathrm{Angle\,Concentration}(\boldeq{Y}^{(i)}) = \frac{\left \| \sum_{q=1}^n \widetilde{\boldeq{Y}}_{q, :}^{(i)} \right \|^2-n}{n\times(n-1)}\,;\sum_{q=1}^{n}\sum_{p=1}^n \widetilde{\boldeq{Y}}_{q, :}^{(i)}\widetilde{\boldeq{Y}}_{p, :}^{{(i)}^T} = \left \| \sum_{q=1}^n \widetilde{\boldeq{Y}}_{q, :}^{(i)} \right \|^2-\sum_{q=1}^n  \left \|\widetilde{\boldeq{Y}}_{q, :}^{(i)}\right \|^2$},
 \label{eq:efficient_angle_concentration}
\end{equation}
which reduces the computational complexity from $\mathcal{O}(hn^2)$ in Equation (\ref{eq:angle_concentration}) to $\mathcal{O}(hn)$. 

\subsubsection{Speedup in Efficient Forward Pass}
\label{sec:appendix_speedup_in_efficient_fo}
We recall Equation (\ref{eq:one_forward_with_m_batch}) for the last-$m$-rank \gls{SVD} components: 
\begin{equation}
\scalebox{0.98}{$(\boldeq{X}\boldeq{V}_{:,r-m+1:r}\boldeq{\Sigma}_{r-m+1:,r-m+1:})[:,\mathrm{newaxis},:]\odot\boldeq{U}_{:,r-m+1:r}[\mathrm{newaxis},:,:]= [\boldeq{Y}^{(r-m+1)},...,\boldeq{Y}^{(r)}]$}. \nonumber
\end{equation}

The \gls{FLOPs} for $\boldeq{X}\boldeq{V}_{:,r-m+1:r}\boldeq{\Sigma}_{r-m+1:,r-m+1:}$ is $2 n d m + nm$ and the \gls{FLOPs} for element-wise product by $\boldeq{U}_{:,r-m+1:r}[\mathrm{newaxis},:,:]$ is $n h  m$. Considering only the input matrix $ \boldeq{W}_{in}$ and the output matrix $ \boldeq{W}_{out} $ in the feed-forward network with $r=\mathrm{min}(h,d)$, and $\mathrm{max}(h,d) = 4\times \mathrm{min}(h,d)$,\footnote{The ratio of $4$ is applied to GPT-J and Roberta. Although for other \glspl{LLM} such as LLaMA \cite{touvron2023llama} this ratio is different, the trend is still kept, i.e., the efficient forward pass does not incur extra computational complexity.} the ratio of the efficient forward pass compared to the standard forward pass is as follows:
\begin{align}
\frac{2 n d  m + nm + nh  m }{2 nh  d} \approx 
\begin{cases}
\frac{6m}{2\times \mathrm{max}(h,d) } \approx 0 & \text{ for } \boldeq{W}_{in} \\
 \frac{9 m}{2\times \mathrm{max}(h,d) }  \approx 0& \text{ for } \boldeq{W}_{out},
\end{cases}
 \nonumber
\end{align}
where $ m \ll \mathrm{max}(h,d)$ as presented in Appendix \ref{sec:appendix_lastm_gptj}, \ref{sec:appendix_topj_lastm_roberta}, and \ref{sec:appendix_comparison}. Therefore, the calculation of the last-$m$-rank \gls{SVD} components does not introduce any significant extra computational complexity. The \gls{FLOPs} of the efficient forward passes are almost zero to the standard forward pass. 

Note that alternatively, we can use only-one-forward pass to get the hidden states $\boldeq{Y}$ and the activations induced by the last-$m$-rank \gls{SVD} components, i.e., $\boldeq{Y}^{(r-m+1)}$ to $\boldeq{Y}^{(r)}$ at the same time, as shown below:
\begin{align}
(\boldeq{X}\boldeq{V}\boldeq{\Sigma})[:,\mathrm{newaxis},:]\odot\boldeq{U}[\mathrm{newaxis},:,:]= [\boldeq{Y}^{(1)},\boldeq{Y}^{(2)},...,\boldeq{Y}^{(r)}], \nonumber
\end{align}
where the \gls{FLOPs} for $\boldeq{X}\boldeq{V}\boldeq{\Sigma}$ is $2n  d  r + n r$ and the \gls{FLOPs} for element-wise product by $\boldeq{U}[\mathrm{newaxis},:,:]$ is $nh  r$. In addition, for the hidden states $\boldeq{Y}=\mathrm{sum}([\boldeq{Y}^{(1)},\boldeq{Y}^{(2)},...,\boldeq{Y}^{(r)}])$, the \gls{FLOPs} is $n h  r$. Considering only the input matrix $ \boldeq{W}_{in}$ and the output matrix $ \boldeq{W}_{out} $ in the feed-forward network with $r=\mathrm{min}(h,d)$, and $\mathrm{max}(h,d) = 4\times \mathrm{min}(h,d)$, the ratio of only-one-forward pass compared to the standard forward pass is as follows:
\begin{align}
\frac{2n  d  r + n r + 2 n h  r }{2nh  d} \approx \frac{2 n r  (d+h)}{2 n  h  d} = 1.25.
 \nonumber
\end{align}
The ratio of $1.25$ means that only-one-forward pass introduces an extra $25\%$ of computational complexity compared to the standard forward pass. At the same time, this alternative, namely, only-one-forward pass, potentially introduces extra dynamic memory consumption for our approach as shown in Table \ref{tab:memo_analysis} because $m$ is increased to $r$, where $ m \ll r$.

\subsubsection{Speedup in Efficient Angle Concentration}
\label{sec:appendix_speedup_in_efficient_angle}
We recall Equation (\ref{eq:angle_concentration}) for the angle concentration:
\begin{align}
\mathrm{Angle\,Concentration}(\boldeq{Y}^{(i)})= \frac{1}{n\times(n-1)}\sum_{q=1}^{n}\sum_{p=1}^n \frac{\boldeq{Y}_{q, :}^{(i)}{\boldeq{Y}_{p, :}^{(i)}}^T}{\left \| \boldeq{Y}_{q, :}^{(i)} \right \| \cdot \left \| \boldeq{Y}_{p, :}^{(i)} \right \|} \quad (q\neq p) \nonumber
\end{align}
where $\boldeq{Y}_{q, :}^{(i)} \in \mathbb{\mathbb{R}}^{1\times h}$ is the $q$-th row of $\boldeq{Y}^{(i)}$ and $\boldeq{Y}_{p, :}^{(i)} \in \mathbb{\mathbb{R}}^{1\times h}$ is the $p$-th row of $\boldeq{Y}^{(i)}$. Similarly, the angle concentration of the hidden states $\boldeq{Y}$, defined as the sum of activations induced by all \gls{SVD} components, is given by $\mathrm{Angle\,Concentration}(\boldeq{Y})$.

For the \gls{FLOPs} of $\left \| \boldeq{Y}_{q, :}^{(i)} \right \|$, the square of $n$ rows of $\boldeq{Y}^{(i)}$ is $n h$ on \gls{Mul} and the sum is $n(h-1)$ on \gls{Add}. The square root only executes $1$ time for each row of $\boldeq{Y}^{(i)}$, which is much smaller than the square and the sum in terms of \gls{FLOPs}. Next, the $\boldeq{Y}_{q, :}^{(i)}{\boldeq{Y}_{p, :}^{(i)}}^T$ needs $h$ \gls{Mul} and $h-1$ \gls{Add}, as well as $1$ \gls{Mul} in $\left \| \boldeq{Y}_{q, :}^{(i)} \right \| \cdot \left \| \boldeq{Y}_{p, :}^{(i)} \right \|$. The division also only executes $1$ time for each $q,p$ pair, which is much smaller than $\boldeq{Y}_{q, :}^{(i)}{\boldeq{Y}_{p, :}^{(i)}}^T$ in terms of \gls{FLOPs}. Then, the $1$ \gls{Add} is consumed for being added to the sum for each $q,p$ pair. Moreover, the \gls{FLOPs} of square root and division can be neglected. Therefore, the \gls{FLOPs} of the standard calculation for one angle concentration is $2n^2(h+1)-2n$. For last-$m$-rank \gls{SVD} components and the hidden states $\boldeq{Y}$, the ratio of this standard angle concentration compared to the standard forward pass is as follows:
\begin{align}
\frac{(m+1)\times\left(2n^2(h+1)-2n\right) }{2 n h  d} =\frac{(m+1)\times (n h+n-1)}{ h  d} \approx \underset{\mathrm{with}\, n \ll h}{\frac{(m+1)\times n}{ d}}.
 \nonumber
\end{align}
In addition, using the symmetry property in Equation (\ref{eq:angle_concentration}), i.e., only calculating unique $q-p$ pairs, will reduce the above ratio to half, i.e., $\frac{(m+1)\times n}{2 d}$. But still, this introduced computational complexity scales linearly with the sequence lengths $n$, which can not be neglected, e.g., when $m=20$, $n=100$, and $d=768$ for Roberta with $ \boldeq{W}_{in}$ on Bios Profession, $\frac{(m+1)\times n}{2 d} = 1.37$.

Instead, as shown in Equation (\ref{eq:efficient_angle_concentration}) below, for $\widetilde{\boldeq{Y}}_{q, :}^{(i)}=\frac{\boldeq{Y}_{q, :}^{(i)}}{\left \|\boldeq{Y}_{q, :}^{(i)}\right \|}$ or $\widetilde{\boldeq{Y}}_{p,:}^{(i)}=\frac{\boldeq{Y}_{p, :}^{(i)}}{\left \|\boldeq{Y}_{p, :}^{(i)}\right \|}$, the \gls{FLOPs} is $n h + n(h-1)+nh$. In addition, the \gls{FLOPs} for $\left \| \sum_{q=1}^n \widetilde{\boldeq{Y}}_{q, :}^{(i)} \right \|^2$ is $h(n-1)+h+h$. 
\begin{equation}
\scalebox{0.95}{$
 \mathrm{Angle\,Concentration}(\boldeq{Y}^{(i)}) = \frac{\left \| \sum_{q=1}^n \widetilde{\boldeq{Y}}_{q, :}^{(i)} \right \|^2-n}{n\times(n-1)}\,;\sum_{q=1}^{n}\sum_{p=1}^n \widetilde{\boldeq{Y}}_{q, :}^{(i)}\widetilde{\boldeq{Y}}_{p,:}^{{(i)}^T} = \left \| \sum_{q=1}^n \widetilde{\boldeq{Y}}_{q, :}^{(i)} \right \|^2-\sum_{q=1}^n  \left \|\widetilde{\boldeq{Y}}_{q, :}^{(i)}\right \|^2$}. \nonumber
\end{equation}
Overall, the \gls{FLOPs} of the efficient calculation for one angle concentration is $4hn+h-n$, which reduces the computational complexity from $\mathcal{O}(hn^2)$ in Equation (\ref{eq:angle_concentration}) to $\mathcal{O}(hn)$. For last-$m$-rank \gls{SVD} components and the hidden states $\boldeq{Y}$, the ratio of efficient angle concentration compared to the standard forward pass is as follows:
\begin{align}
\frac{(m+1)\times(4hn+h-n) }{2 n h  d} =\frac{2(m+1)}{ d} + \frac{(m+1)}{ 2 n  d} - \frac{(m+1)}{ 2h  d}
\approx \underset{\mathrm{with}\, m \ll d}{0}.
 \nonumber
\end{align}

\subsubsection{Computational Complexity of Efficient Forward-Only Search}
\label{sec:appendix_speedup_overall}
Based on the analysis of speedup in efficient forward pass in Appendix \ref{sec:appendix_speedup_in_efficient_fo} and efficient angle concentration in Appendix \ref{sec:appendix_speedup_in_efficient_angle}, we have the ratio of the efficient forward pass for the last-$m$-rank \gls{SVD} components compared to the standard forward pass is $ \begin{cases}
\frac{6 m}{2\times \mathrm{max}(h,d) } \approx 0 & \text{ for } \boldeq{W}_{in} \\
 \frac{9 m}{2\times \mathrm{max}(h,d) }  \approx 0& \text{ for } \boldeq{W}_{out},
\end{cases} $; and the ratio of efficient angle concentration for the last-$m$-rank \gls{SVD} components and the hidden states $\boldeq{Y}$ compared to the standard forward pass is $ \frac{2(m+1)}{ d} + \frac{(m+1)}{ 2n  d} - \frac{(m+1)}{ 2 h  d}
\approx \underset{\mathrm{with}\, m \ll d}{0} $. Therefore, the ratio of our efficient forward-only search compared to the standard forward pass in terms of computational complexity is:
\begin{align}
\mathrm{ratio} = 
1+\begin{cases}
\frac{6 m}{2\times \mathrm{max}(h,d) } \approx 0 & \text{ for } \boldeq{W}_{in} \\
 \frac{9 m}{2\times \mathrm{max}(h,d) }  \approx 0& \text{ for } \boldeq{W}_{out},
\end{cases} + \underset{\mathrm{with}\, m \ll d}{0} \approx 1.
 \label{eq:final_ratio}
\end{align}

\begin{figure*}[!ht]
  \centering
  \begin{subfigure}{1\textwidth}
    \includegraphics[width=\linewidth]{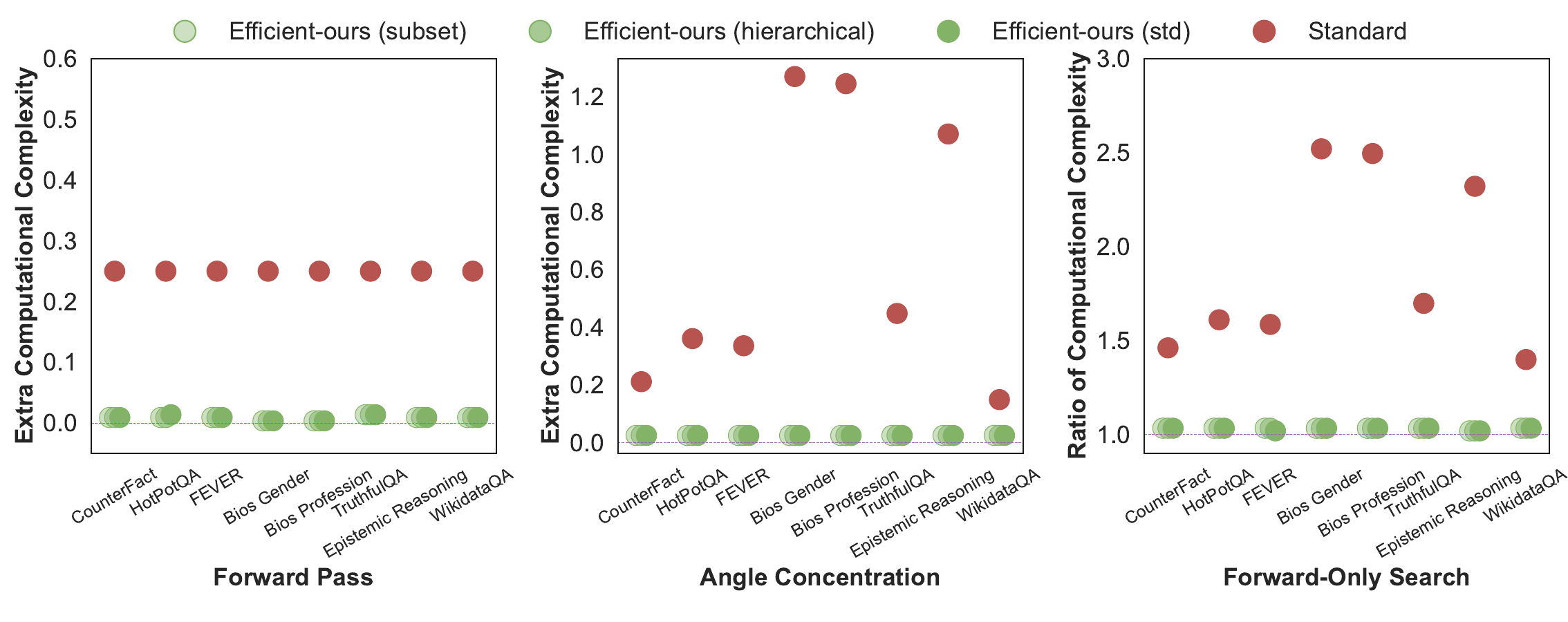}
    \caption{GPT-J}
  \end{subfigure}
  \begin{subfigure}{1\textwidth}
    \includegraphics[width=\linewidth]{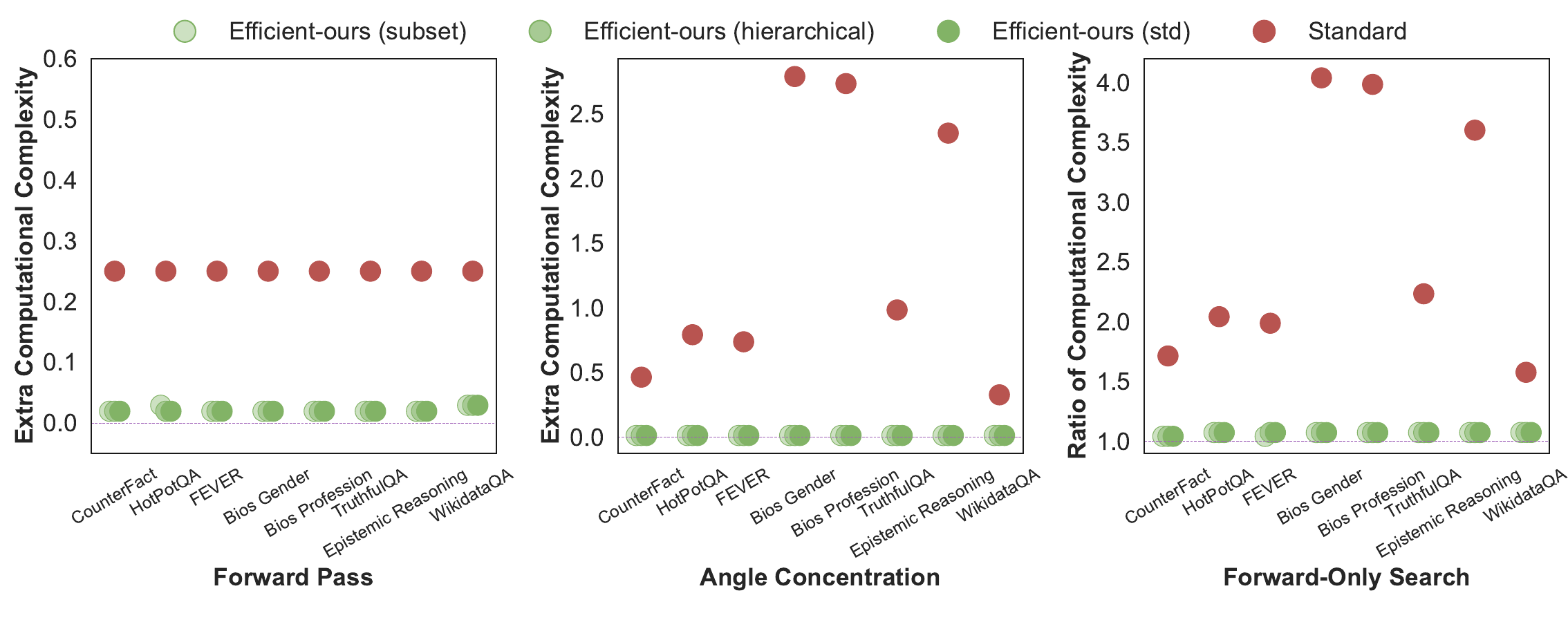}
    \caption{Roberta}
  \end{subfigure}
  \caption{Compared to the standard forward pass, our efficient forward pass introduces negligible extra computational complexity. Compared to the standard angle concentration, our efficient angle concentration does not incur extra computational complexity. Overall, the ratio of our efficient forward-only search compared to the standard forward pass in computational complexity is approximately $1$ for both (a) GPT-J and (b) Roberta on eight datasets.}
  \label{fig:efficient}
\end{figure*}

Moreover, Fig. \ref{fig:efficient} shows that the only-one-forward pass and the standard angle concentration both introduce non-negligible extra computational complexity. Furthermore, the ratio of our efficient forward-only search compared to the standard forward pass in computational complexity is approximately $1$ for both Roberta and GPT-J on eight datasets.\footnote{For {\color{mydarkred}\LARGE$\bullet$} Standard in angle concentration, we consider $ \boldeq{W}_{in}$ for presentation. While for $ \boldeq{W}_{out}$, the trend is still kept, i.e., the standard angle concentration introduces non-negligible extra computational complexity.}

Furthermore, our experimental setting in Section \ref{sec:setting} only considers the weight matrices $\boldeq{W}_{in/out}$ in the feed-forward network of the Transformer architecture, rather than all the weight matrices in the Transformer architecture. \textbf{Therefore, the actual computational complexity of our efficient forward pass plus our efficient angle concentration compared to the standard forward pass in \glspl{LLM} is significantly smaller than} $\begin{cases}
\frac{6 m}{2\times \mathrm{max}(h,d) } \approx 0 & \text{ for } \boldeq{W}_{in} \\
 \frac{9 m}{2\times \mathrm{max}(h,d) }  \approx 0& \text{ for } \boldeq{W}_{out},
\end{cases} + \underset{\mathrm{with}\, m \ll d}{0} $. The ratio of our efficient forward-only search compared to the standard forward pass in terms of computational complexity for \glspl{LLM} in our paper's experimental setting is almost $1$.
\newpage
\subsection{Computational Complexity Comparison Across Methods}
\label{sec:appendix_comparison}

\begin{table}[!ht]
  \caption{The size of $\mathcal{D}_{\mathrm{subset}}$, $\mathcal{D}_{\mathrm{std}}$, $\mathcal{D}$, and the average length of sequence $n$ for evaluated datasets.}
  \label{tab:appendix_datasets}
  \centering
    \resizebox{\columnwidth}{!}{
  \begin{tabular}{lcccc}
    \toprule
    Dataset   & $\mathcal{D}_{\mathrm{subset}}$ (100 from $\mathcal{D}_{\mathrm{std}}$) & Sequence Length $n$  &  $\mathcal{D}_{\mathrm{std}}$ (20\% of $\mathcal{D}$)   & $\mathcal{D}$   \\
    \midrule
    CounterFact & 100 & 17 & 13151  & 65757 \\
    HotPotQA  & 100 &29 &  2923   &14618   \\
    FEVER    & 100 & 27& 2617  &13086   \\
    Bios Gender& 100 & 102& 7928  &39642  \\
    Bios Profession & 100 &100 & 3844  &19223 \\
    TruthfulQA & 100 & 36& 1176  &5882 \\
    Epistemic Reasoning & 100 & 86& 400  &2000 \\
    WikidataQA & 100 & 12 &4064  &20321  \\
    \bottomrule
  \end{tabular}}
\end{table}

When referring to the comparison of computational complexity, we consider the speedup in search steps, excluding the computation of the test stage, i.e., over $80\%$ of the whole dataset $\mathcal{D}$. The sequence length $n$ is the averaged value across $\mathcal{D}_{\mathrm{subset}}$.

For \textbf{LASER} \cite{sharma2024the}, 
\begin{align}
\mathrm{LASER\,(Std)}:&\,\underset{\mathrm{layers}}{L}\times \underset{\mathrm{\boldeq{W}}_{\mathrm{in/out}}}{2}\times\underset{\mathrm{rates}}{9}\times \mathcal{D}_{\mathrm{std}}  
 \nonumber \\
\mathrm{LASER\,(subset)}:&\,\underset{\mathrm{layers}}{L}\times \underset{\mathrm{\boldeq{W}}_{\mathrm{in/out}}}{2}\times\underset{\mathrm{rates}}{9}\times \mathcal{D}_{\mathrm{subset}}, \nonumber
\end{align}
where $L=12$ for Roberta and $L=28$ for GPT-J.

For \textbf{LASER-Grad} \cite{sreeram2025compress}, LASER-Grad (std) has $j=7$ and LASER-Grad (subset) $j=7$ for Roberta and $j=5$ for GPT-J.
\begin{align}
\mathrm{LASER\text{-}Grad\,(std)}:&\,\underset{\mathrm{forward\,and\, backward}}{2.5\times4\times  \mathcal{D}_{\mathrm{subset}}}+\underset{\mathrm{top\text{-}j}}{j}\times \underset{\mathrm{clustering}}{4}\times\underset{\mathrm{rates}}{9}\times \mathcal{D}_{\mathrm{std}}  
 \nonumber \\
\mathrm{LASER\text{-}Grad\,(subset)}:&\,\underset{\mathrm{forward\,and\, backward}}{2.5\times4\times  \mathcal{D}_{\mathrm{subset}}}+\underset{\mathrm{top\text{-}j}}{j}\times \underset{\mathrm{clustering}}{4}\times\underset{\mathrm{rates}}{9}\times \mathcal{D}_{\mathrm{subset}}.   \nonumber
\end{align}

For our proposed forward-only approach (\textbf{ours}), the computational complexity of our efficient forward-only search compared to the standard forward pass is denoted as in Equation (\ref{eq:final_ratio}) and Fig. \ref{fig:efficient}, where $m$ is the last-$m$-rank \gls{SVD} components and $d$ is the dimension in a weight matrix $\boldeq{W} \in \mathbb{\mathbb{R}}^{h\times d}$,
\begin{align}
\mathrm{ratio} =  1+ \begin{cases}
\frac{6 m}{2\times \mathrm{max}(h,d) } +\frac{2(m+1)}{ d} + \frac{(m+1)}{ 2 n   d} - \frac{(m+1)}{ 2 h  d} & \text{ for } \boldeq{W}_{in} \\
 \frac{9 m}{2\times \mathrm{max}(h,d) } +\frac{2(m+1)}{ d} + \frac{(m+1)}{ 2 n   d} - \frac{(m+1)}{ 2 h  d} & \text{ for } \boldeq{W}_{out}, 
\end{cases} \approx 1  \label{eq:ratio}
\end{align}
where $m=20$ for Roberta and $m=50$ for GPT-J.\footnote{$m=20$ is enough for Bios Gender and Bios Profession.} Next, we will analyze the computational complexity for variants of rank-reduction strategy in our forward-only approach.

\subsubsection{Efficient Hierarchical Search}
\label{sec:appendix_efficient_hierarchical}
As we mentioned in Section \ref{sec:forward-only-search}, the rank-reduction strategy is determined based on the accuracy evaluated across the validation datasets. In other words, the rank-reduction strategy varies across different sizes of datasets. As shown in Algorithm \ref{alg:ours_subset}, Algorithm \ref{alg:ours_std}, and Algorithm \ref{alg:ours_hierarchical}, we present the variants of rank-reduction strategy in our forward-only approach, i.e., ours (subset), ours (std), ours (hierarchical) correspondingly. For ours (subset) in Algorithm \ref{alg:ours_subset}, the validation dataset is $\mathcal{D}_{\mathrm{subset}}$, which is only $100$ samples selected from $\mathcal{D}_{\mathrm{std}}$, as shown in Table \ref{tab:appendix_datasets}. For ours (std), the validation dataset is $\mathcal{D}_{\mathrm{std}}$ in Algorithm \ref{alg:ours_std}, which is $20\%$ samples from the whole dataset $\mathcal{D}$, as shown in Table \ref{tab:appendix_datasets}. For ours (hierarchical) in Algorithm \ref{alg:ours_hierarchical}, we exploit the computational efficiency over $\mathcal{D}_{\mathrm{subset}}$ and then only evaluate the top-$j$ weight matrices $\widehat{\boldeq{W}}$ over $\mathcal{D}_{\mathrm{std}}$, avoiding the costly evaluation with all weight matrices $\widehat{\boldeq{W}}$ over $\mathcal{D}_{\mathrm{std}}$. 

\begin{minipage}[t]{0.45\textwidth}
\vspace{0pt} 
\begin{algorithm}[H]
   \caption{Ours (subset)}
   \label{alg:ours_subset}
    \begin{algorithmic}[1]
    \STATE {\bfseries within}  {candidate $\widehat{\boldeq{W}}$ from Algorithm \ref{alg:main}}
    \STATE \hspace{0.8em} \textcolor{gray}{\# efficient evaluation}
    \STATE \hspace{0.8em} Best $\widehat{\boldeq{W}} \leftarrow \left \{\mathrm{evaluate\,over\,\mathcal{D}_{\mathrm{subset}}}  \right \} $
    \STATE {\bfseries Return:} \gls{LLM} with the best $\widehat{\boldeq{W}}$
\end{algorithmic}
\end{algorithm}
\end{minipage}
\hspace{4pt}
\begin{minipage}[t]{0.45\linewidth}
\vspace{-1pt} 
\begin{algorithm}[H]
   \caption{Ours (std)}
   \label{alg:ours_std}
    \begin{algorithmic}[1]
    \STATE {\bfseries within}  {candidate $\widehat{\boldeq{W}}$ from Algorithm \ref{alg:main}}
    \STATE \hspace{0.8em} \textcolor{gray}{\# costly evaluation}
    \STATE \hspace{0.8em} Best $\widehat{\boldeq{W}} \leftarrow \left \{\mathrm{evaluate\,over\,\mathcal{D}_{\mathrm{std}}}  \right \} $
    \STATE {\bfseries Return:} \gls{LLM} with the best $\widehat{\boldeq{W}}$
\end{algorithmic}
\end{algorithm}
\end{minipage}
\begin{center}
\begin{minipage}[t]{0.5\linewidth}
\vspace{0pt} 
\begin{algorithm}[H]
   \caption{Ours (hierarchical)}
   \label{alg:ours_hierarchical}
    \begin{algorithmic}[1]
    \FOR{each $\widehat{\boldeq{W}}$ from Algorithm \ref{alg:main}}
    \STATE \textcolor{gray}{\# efficient evaluation with Set $\mathbb{S}'$}%=\emptyset$}
    \STATE $\mathbb{S}' \leftarrow \mathbb{S}' \cup \camera{\left\{\left (\widehat{\boldeq{W}},\, \mathrm{evaluate\,over\,\mathcal{D}_{\mathrm{subset}}}  \right )\right\} }$
    \ENDFOR
    \STATE {\bfseries within}  {top-$j\,\widehat{\boldeq{W}}$ in $\mathbb{S}'$}
    \STATE \hspace{0.8em} \textcolor{gray}{\# efficient evaluation with only top-$j$}
    \STATE \hspace{0.8em} Best $\widehat{\boldeq{W}} \leftarrow \left \{\mathrm{evaluate\,over\,\mathcal{D}_{\mathrm{std}}}  \right \} $
    \STATE {\bfseries Return:}  \gls{LLM} with the best $\widehat{\boldeq{W}}$
\end{algorithmic}
\end{algorithm}
\end{minipage}
\end{center}

Consequently, we have the computational complexity for variants of the rank-reduction strategy in our forward-only approach, where $j=5$ for Roberta\footnote{Only $j=10$ for $\mathrm{ours\,(std)}$ with \camera{CounterFact} on Roberta.} and $j=10$ for GPT-J:
\begin{align}
\mathrm{Ours\,(hierarchical)}:&\,\underset{\mathrm{forward\text{-}only}}{\mathrm{ratio} \times  \mathcal{D}_{\mathrm{subset}}}+\underset{\mathrm{top\text{-}j}}{j}\times\underset{\mathrm{rates}}{9}\times \mathcal{D}_{\mathrm{subset}}+\underset{\mathrm{top\text{-}j}}{j}\times \mathcal{D}_{\mathrm{std}}  
 \label{eq:our_hierarchical} \\
\mathrm{Ours\,(std)}:&\,\underset{\mathrm{forward\text{-}only}}{\mathrm{ratio} \times  \mathcal{D}_{\mathrm{subset}}}+\underset{\mathrm{top\text{-}j}}{j}\times\underset{\mathrm{rates}}{9}\times \mathcal{D}_{\mathrm{std}}  
 \nonumber \\
\mathrm{Ours\,(subset)}:&\,\underset{\mathrm{forward\text{-}only}}{\mathrm{ratio} \times  \mathcal{D}_{\mathrm{subset}}}+\underset{\mathrm{top\text{-}j}}{j}\times \underset{\mathrm{rates}}{9}\times \mathcal{D}_{\mathrm{subset}}.   \nonumber
\end{align}

\begin{figure}[!hb]
  \centering
  \begin{subfigure}{0.47\textwidth}
    \includegraphics[width=\linewidth]{./img/gptj_speedup_new.pdf}
        \caption*{(a) GPT-J}
  \end{subfigure}
  \begin{subfigure}{0.47\textwidth}
    \includegraphics[width=\linewidth]{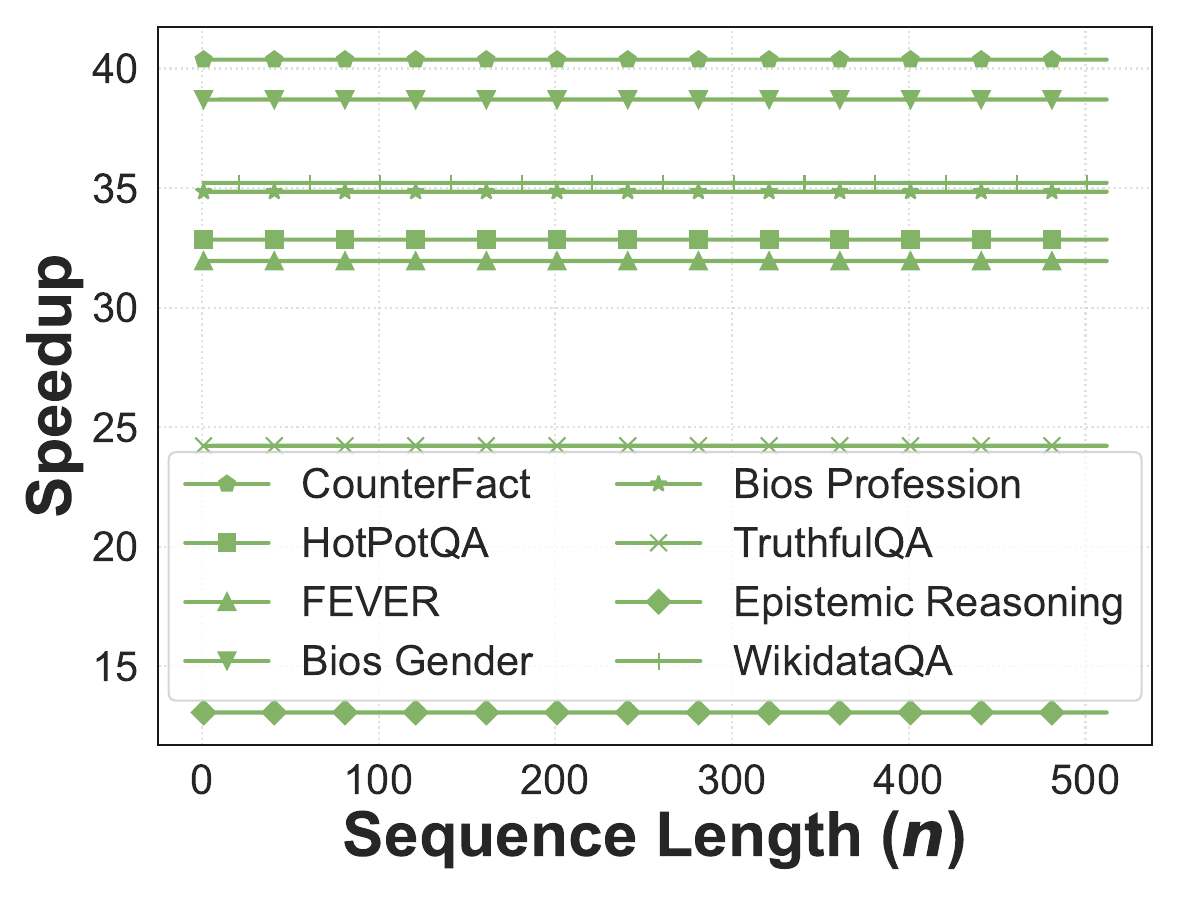}
  \caption*{(b) Roberta}
  \end{subfigure}
  \caption{The Speedup with different sequence lengths $n$ for ours (hierarchical). Our forward-only approach achieves greater Speedup across different $n$.}
  \label{fig:speedup_different}
\end{figure}

\begin{figure}[!ht]
 \centering
  \begin{subfigure}
  {1\textwidth}
  \centering \includegraphics[width=1\linewidth]{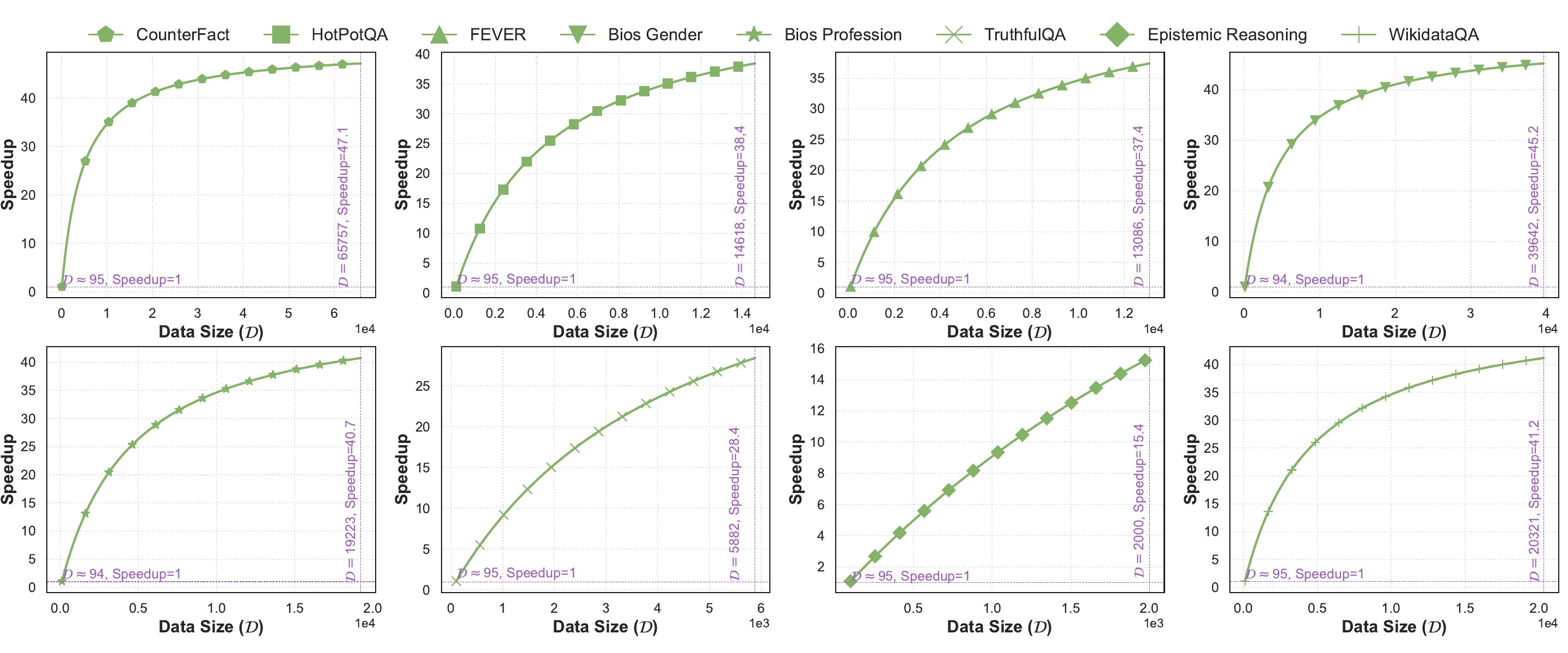}
  \caption{GPT-J}
    \end{subfigure}
  \begin{subfigure}{1\textwidth}
  \centering \includegraphics[width=1\linewidth]{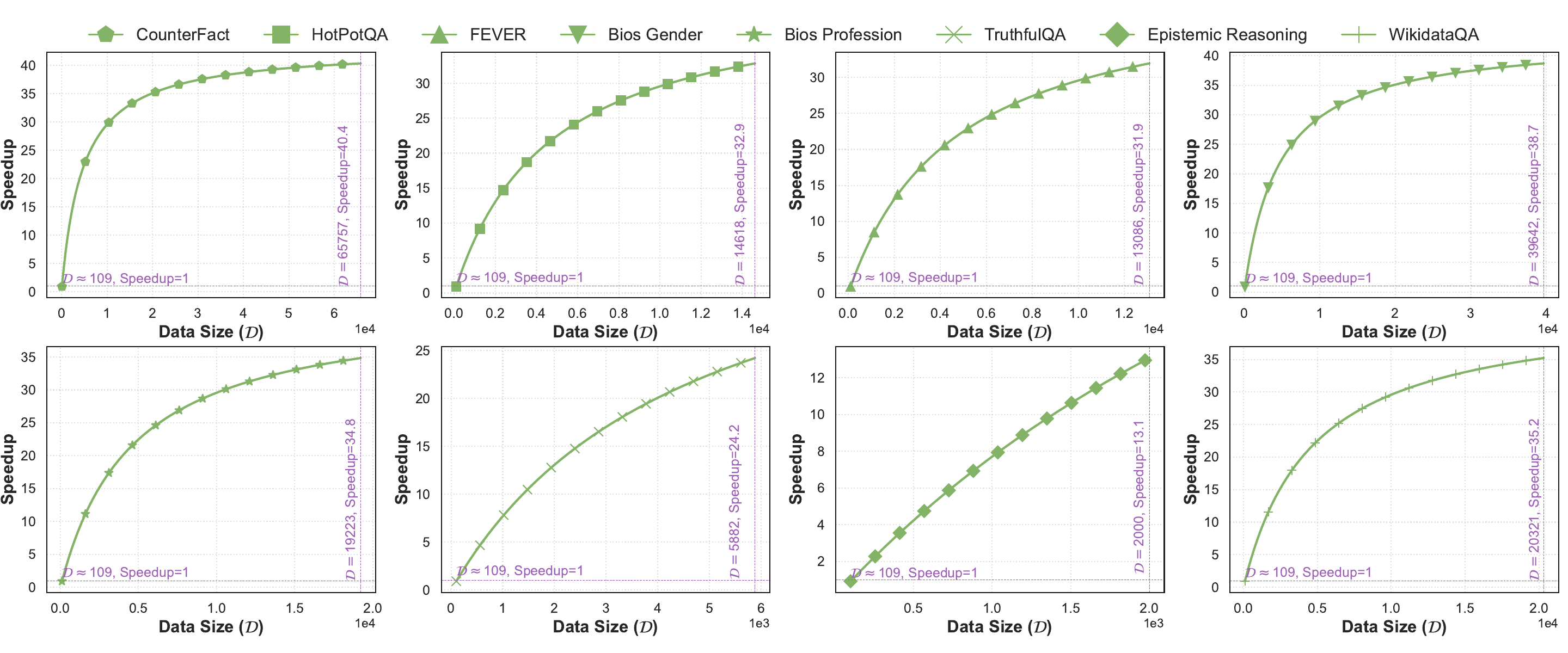}
  \caption{Roberta}
    \end{subfigure}
  \caption{The Speedup with assumed different data size $\mathcal{D}$ for ours (hierarchical). Our forward-only approach achieves greater Speedup as the data size $\mathcal{D}$ increases for both (a) GPT-J and (b) Roberta.}
  \label{fig:gptj_roberta_datasize}
\end{figure}

\newpage
Besides, we also show the Speedup with assumed different data size $\mathcal{D}$ for ours (hierarchical). Our forward-only approach achieves greater Speedup as the data size $\mathcal{D}$ increases for both (a) GPT-J and (b) Roberta on eight datasets, as demonstrated in Fig. \ref{fig:gptj_roberta_datasize}

\section{The Declaration of AI Usage}
\label{sec:ai_use}
The icon \includegraphics[height=1em]{./img/icon_new.png} (cheetah) is generated with ChatGPT \citep{betker2023improving,achiam2023gpt}. %Gemini \cite{team2023gemini} Nano Banana. 
In addition, we use \glspl{LLM} only for polishing the writing, and it does not impact the core methodology, scientific rigorousness, or originality of the research.

\clearpage
\section{Peak Memory Consumption}
\label{sec:appendix_peak_memory}
From Table \ref{tab:memo_analysis}, we analyze the peak memory consumption for conceptualized fully connected layers across baseline (direct inference), LASER \citep{sharma2024the}, LASER-Grad \citep{sreeram2025compress}, and our forward-only approach. In this section, we examine in detail the peak memory consumption specifically for Transformer-based \cite{vaswani2017attention} \glspl{LLM}, with the default precision in Float32 (4 bytes per data point) for Roberta and in Float16 (2 bytes per data point) for GPT-J. 

\begin{table}[!ht]
  \caption{The settings for Roberta \cite{liu2019roberta} and GPT-J \cite{gpt-j}, where $\mathrm{vocab\_size}$ (the size of vocabulary), $\mathrm{hidden\_size}$ (the dimension of the hidden layer), $L$ (the number of stacked Transformer modules), $\mathrm{attention\_head}$ (the number of attention heads).}
  \label{tab:setting_llm}
  \centering
  \begin{tabular}{cccccc}
    \toprule
    Setting     & $\mathrm{vocab\_size}$     & $\mathrm{hidden\_size}$ & $\mathrm{attention\_head}$ & $L$ & Max Sequence Length \\
    \midrule
    Roberta & 50265 &768&12&12&512 \\
    GPT-J &50400&4096&16&28&2048\\ 
    \bottomrule
  \end{tabular}
\end{table}

\subsection{Roberta}
The input sequence is firstly processed with an input embedding and a positional encoding, and then passed into $L$ stacked Transformer modules, finally with a fully-connected-layer block called the \gls{LLM} head. Moreover, each Transformer module has a multi-head self-attention and a feed-forward network. We assume that the input sequence is $ \mathbb{\mathbb{R}}^{n\times \mathrm{hidden\_size}}$ and the batch size is $1$. In the multi-head self-attention, 
\begin{align}
\mathrm{Attention}(\boldeq{Q},\boldeq{K},\boldeq{V}) = \mathrm{softmax}(\frac{\boldeq{Q}\boldeq{K}^T}{\sqrt{\mathrm{head\_dimension}} })\boldeq{V},
 \nonumber 
\end{align}
where $\mathrm{head\_dimension} = \frac{\mathrm{hidden\_size}}{\mathrm{attention\_head}}$ and the generated matrices $\boldeq{Q},\boldeq{K},\boldeq{V}$ are all in $ \mathbb{\mathbb{R}}^{n\times \mathrm{hidden\_size}}$. In this multi-head self-attention, the attention matrix $ \mathrm{softmax}(\frac{\boldeq{Q}\boldeq{K}^T}{\sqrt{\mathrm{head\_dimension}} }) $ consumes memory quadratically, i.e., in $ \mathbb{\mathbb{R}}^{n\times n\times\mathrm{attention\_{head}}}$. After multiplication with $\boldeq{V}$, the attention output ($ \mathbb{\mathbb{R}}^{n\times \mathrm{hidden\_size}}$) with a linear projection, is passed into the feed-forward network. In the feed-forward network, we have the input matrix $ \boldeq{W}_{in} \in \mathbb{\mathbb{R}}^{(4\times\mathrm{hidden\_size})\times \mathrm{hidden\_size}}$ and the output matrix $ \boldeq{W}_{out} \in\mathbb{\mathbb{R}}^{\mathrm{hidden\_size}\times (4\times\mathrm{hidden\_size})}$. In this paper, we also only search for $ \boldeq{W}_{in/out} $ as did in LASER \citep{sharma2024the} and LASER-Grad \citep{sreeram2025compress}. After the stacked Transformer modules, the \gls{LLM} head generates the output, in $\mathbb{\mathbb{R}}^{n\times \mathrm{vocab\_size}}$, over the vocabulary. 

For $\mathrm{\underset{\mathrm{Roberta}}{\textbf{Baseline}}}$ (direct inference), the \gls{LLM} is loaded with parameters (mainly weight memory). For dynamic memory, mainly the generated activations in Transformer modules, we consider the maximum peak memory in one Transformer module because this part of the maximum peak memory is the same for $L$ stacked Transformer modules. Then, we need to compare the maximum peak memory for Transformer modules and the peak memory in \gls{LLM} head. Usually, the attention matrix occupies the peak memory in Transformer modules, and the output in \gls{LLM} head also occupies lots of memory. For simplicity, we exclude the effect of the input embedding and the positional encoding, skip connection, and LayerNorm, because they can be ignored when compared to the attention matrix and the output in \gls{LLM} head. Furthermore, for the layers with $\boldeq{W}_{in}$ or $\boldeq{W}_{out}$, and activation function (GELU Activation for Roberta \cite{liu2019roberta}), in the feed-forward network, the maximum activations are the same, i.e., in $\mathbb{\mathbb{R}}^{n\times (4\times\mathrm{hidden\_size})}$.
\begin{align}
\underset{\mathrm{one\_Transformer\_module}}{\mathrm{max\_baseline}}\leftarrow& \mathrm{max}\Big\{\underset{\mathrm{attention\,matrix}}{n\times n\times\mathrm{attention\_head}},\underset{\mathrm{feed\text{-}forward\,network}}{n\times (4\times\mathrm{hidden\_size})}\Big\},
\nonumber \\
\mathrm{\underset{\mathrm{Roberta}}{\textbf{Baseline}}}:&\,\underset{\mathrm{LLM}}{\mathrm{Parameters}}+\mathrm{max}\Big\{\underset{\mathrm{one\_Transformer\_module}}{\mathrm{max\_baseline}},\underset{\mathrm{LLM\,head}}{n\times\mathrm{vocab\_size}}\Big\},
 \nonumber 
\end{align}
where the $\underset{\mathrm{one\_Transformer\_module}}{\mathrm{max\_baseline}}$ is simplified because the input to each Transformer module, the output of multi-head self-attention, and the output of the feed-forward network are all in $\mathbb{\mathbb{R}}^{n\times\mathrm{hidden\_size}}$. Moreover, the memory for $\boldeq{Q},\boldeq{K},\boldeq{V}$ are in $\mathbb{\mathbb{R}}^{n\times(3\times\mathrm{hidden\_size})}$, which is smaller than $\underset{\mathrm{feed\text{-}forward\,network}}{n\times (4\times\mathrm{hidden\_size})}$.

For $\mathrm{\underset{\mathrm{Roberta}}{\textbf{LASER}}}$ \cite{sharma2024the}, the \gls{LLM} is also loaded with parameters (mainly weight memory). For dynamic memory, we also only need to consider the maximum peak memory of one Transformer module. In addition to the generated activations as in $\mathrm{\underset{\mathrm{Roberta}}{\textbf{Baseline}}}$, the \gls{SVD} decomposition $\boldeq{U}\boldeq{\Sigma} \boldeq{V}^T$ is introduced, which changes the dynamic memory. Because of only considering $ \boldeq{W}_{in/out} $ \camera{and taking $ \boldeq{W}_{in} $ for example}, the introduced extra dynamic memory for $\boldeq{U}$ is in $\mathbb{\mathbb{R}}^{ (4\times\mathrm{hidden\_size})\times r}$, for $\boldeq{\Sigma}$ is in $\mathbb{\mathbb{R}}^{ r\times r}$, and for $\boldeq{V}^T$ is in $\mathbb{\mathbb{R}}^{  r \times \mathrm{hidden\_size}}$, where $r=\mathrm{hidden\_size}$. Totally, the introduced extra dynamic memory is in $\mathbb{\mathbb{R}}^{(4\times\mathrm{hidden\_size})\times r+r\times r+r\times \mathrm{hidden\_size}}$ for both $ \boldeq{W}_{in} $ and $ \boldeq{W}_{out} $.
\begin{align}
\underset{\mathrm{one\_Transformer\_module}}{\mathrm{max\_LASER}}\leftarrow& \mathrm{max}\Big\{\underset{\mathrm{attention\,matrix}}{n\times n\times\mathrm{attention\_head}},\nonumber \\ &\underset{\mathrm{feed\text{-}forward\,network}}{n\times (4\times\mathrm{hidden\_size})}+\underset{\mathrm{SVD}}{(5\times\mathrm{hidden\_size}+r)\times r}\Big\},
\nonumber \\
\mathrm{\underset{\mathrm{Roberta}}{\textbf{LASER}}}:&\,\underset{\mathrm{LLM}}{\mathrm{Parameters}}+\mathrm{max}\Big\{\underset{\mathrm{one\_Transformer\_module}}{\mathrm{max\_LASER}},\underset{\mathrm{LLM\,head}}{n\times\mathrm{vocab\_size}}\Big\}.
 \nonumber 
\end{align}
Similarly, we exclude the effect of the input embedding and the positional encoding, skip connection, LayerNorm, and activation function, as in $\mathrm{\underset{\mathrm{Roberta}}{\textbf{Baseline}}}$. The $\underset{\mathrm{one\_Transformer\_module}}{\mathrm{max\_LASER}}$ is also simplified by excluding the input to each Transformer module, the output of multi-head self-attention, and the output of the feed-forward network, as well as $\boldeq{Q},\boldeq{K},\boldeq{V}$.

For $\mathrm{\underset{\mathrm{Roberta}}{\textbf{Ours}}}$, the parameters (mainly weight memory) are the same as $\mathrm{\underset{\mathrm{Roberta}}{\textbf{Baseline}}}$ and $\mathrm{\underset{\mathrm{Roberta}}{\textbf{LASER}}}$. For dynamic memory, in addition to the generated activations as in $\mathrm{\underset{\mathrm{Roberta}}{\textbf{Baseline}}}$ and the \gls{SVD} decomposition $\boldeq{U}\boldeq{\Sigma} \boldeq{V}^T$ as in $\mathrm{\underset{\mathrm{Roberta}}{\textbf{LASER}}}$, the extra $m$ \gls{SVD} components change the dynamic memory. When only considering the maximum generated activations within $ \boldeq{W}_{in/out} $ in $\mathbb{\mathbb{R}}^{n\times (4\times\mathrm{hidden\_size})}$, the introduced extra dynamic memory of $m$ \gls{SVD} components is in $\mathbb{\mathbb{R}}^{m\times n\times (4\times\mathrm{hidden\_size})}$. Therefore, the introduced maximum dynamic memory here is totally in $\mathbb{\mathbb{R}}^{(m+1)\times n\times (4\times\mathrm{hidden\_size})}$.
\begin{align}
\underset{\mathrm{one\_Transformer\_module}}{\mathrm{max\_ours}}\leftarrow& \mathrm{max}\Big\{\underset{\mathrm{attention\,matrix}}{n\times n\times\mathrm{attention\_head}},\nonumber \\ &\underset{\substack{\mathrm{feed\text{-}forward\,network\,with}\\m\,\mathrm{singular\,components}}}{(m+1)\times n\times (4\times\mathrm{hidden\_size})} +\underset{\mathrm{SVD}}{(5\times\mathrm{hidden\_size}+r)\times r}\Big\},
\nonumber \\
\mathrm{\underset{\mathrm{Roberta}}{\textbf{Ours}}}:&\,\underset{\mathrm{LLM}}{\mathrm{Parameters}}+\mathrm{max}\Big\{\underset{\mathrm{one\_Transformer\_module}}{\mathrm{max\_ours}},\underset{\mathrm{LLM\,head}}{n\times\mathrm{vocab\_size}}\Big\}.
 \nonumber 
\end{align}
Again, we not only exclude the effect of the input embedding and the positional encoding, skip connection, LayerNorm, activation function, as in $\mathrm{\underset{\mathrm{Roberta}}{\textbf{Baseline}}}$, but also simplify the $\underset{\mathrm{one\_Transformer\_module}}{\mathrm{max\_ours}}$ by excluding the input to each Transformer module, the output of multi-head self-attention, and the output of the feed-forward network, as well as $\boldeq{Q},\boldeq{K},\boldeq{V}$, as in $\mathrm{\underset{\mathrm{Roberta}}{\textbf{LASER}}}$.

For $\mathrm{\underset{\mathrm{Roberta}}{\textbf{LASER-Grad}}}$ \cite{sreeram2025compress}, the generated activations in the forward pass and the SVD decomposition $\boldeq{U}\boldeq{\Sigma} \boldeq{V}^T$, as well as \gls{LLM} parameters are required to be stored in memory for getting the gradient of weights $\boldeq{W}$ and the corresponding singular values $\boldeq{\Sigma}$ in the backward pass. In other words, we need to store the generated activations that are prepared to backpropagate the gradients. In this case, we also need to include the intermediate generated activations in the analysis of dynamic memory. More specifically, when the input sequence is in $ \mathbb{\mathbb{R}}^{n\times \mathrm{hidden\_size}}$, for each Transformer module, there are $12\times$ (considering $\boldeq{Q},\boldeq{K},\boldeq{V}$, LayerNorm, activation function, the input to each Transformer module, and the output of multi-head self-attention) the memory of the input sequence, adding with $2\times$ the memory of the attention weights (due to Softmax). For the feed-forward network with $\boldeq{W}_{in/out}$, based on the memory analysis in \cite{zhang2024revisiting}, we assume that the generated activations in the forward pass and the SVD decomposition could be released once the gradient of weights $\boldeq{W}$ and the corresponding singular values $\boldeq{\Sigma}$ in the backward pass are available. Therefore, we get the maximum memory for the layers with $\boldeq{W}_{in/out} $ in the feed-forward network, which are denoted as follows,
\begin{align}
\underset{\boldeq{W}_{in}}{\mathrm{max\_LASER\text{-}Grad}}\leftarrow & \mathrm{max}\Big\{ \underset{\mathrm{feed\text{-}forward\,network}}{n\times (4\times\mathrm{hidden\_size})}+\underset{\mathrm{SVD}}{(5\times\mathrm{hidden\_size}+r)\times r},\nonumber \\ &\underset{\mathrm{gradient\,of\,}\boldeq{W}_{in}}{(4\times\mathrm{hidden\_size})\times \mathrm{hidden\_size}}+\underset{\mathrm{gradient\,of\,}\boldeq{\Sigma}}{r\times r}\Big\}, \nonumber \\
\underset{\boldeq{W}_{out}}{\mathrm{max\_LASER\text{-}Grad}}\leftarrow & \mathrm{max}\Big\{ \underset{\mathrm{feed\text{-}forward\,network}}{n\times\mathrm{hidden\_size}}+\underset{\mathrm{SVD}}{(5\times\mathrm{hidden\_size}+r)\times r},\nonumber \\ &\underset{\mathrm{gradient\,of\,}\boldeq{W}_{out}}{\mathrm{hidden\_size}\times (4\times\mathrm{hidden\_size})}+\underset{\mathrm{gradient\,of\,}\boldeq{\Sigma}}{r\times r}\Big\}. \nonumber
\end{align}
Overall, the generated activations and the \gls{SVD} decomposition, as well as the required gradients for each Transformer module, are:
\begin{align}
\underset{\mathrm{one\_Transformer\_module}}{\mathrm{sum\_LASER\text{-}Grad}}\leftarrow& \mathrm{sum}\Big\{\underset{\mathrm{intermediate\,activations}}{12\times n\times\mathrm{hidden\_size}}+\underset{\mathrm{attention\,matrix\,with\,softmax}}{2\times n\times n\times\mathrm{attention\_head}}+\nonumber \\ &\underset{\boldeq{W}_{in}}{\mathrm{max\_LASER\text{-}Grad}}+\underset{\boldeq{W}_{out}}{\mathrm{max\_LASER\text{-}Grad}}\Big\},
\nonumber
\end{align}
Furthermore, the embedding layer generates $3\times$ the memory of the input sequence (because of input embedding, positional encoding, and LayerNorm for Roberta). We also need to consider the generated activations in the \gls{LLM} head, which is $2\times$ the memory of the input sequence, adding with the output over the vocabulary. 
\begin{align}
\mathrm{\underset{\mathrm{Roberta}}{\textbf{LASER-Grad}}}:&\,\underset{\mathrm{LLM}}{\mathrm{Parameters}}+\underset{\mathrm{embedding\,layer}}{3\times n\times\mathrm{hidden\_size}}+L\times\Big\{\underset{\mathrm{one\_Transformer\_module}}{\mathrm{sum\_LASER\text{-}Grad}}\Big\}\nonumber \\ &+\underset{\mathrm{LLM\,head}}{(2\times n\times\mathrm{hidden\_size}+n\times\mathrm{vocab\_size})}.
 \nonumber 
\end{align}

\subsection{GPT-J}
For GPT-J, the peak memory consumption is different from that of Roberta. The input sequence is first processed with an input embedding, and then passed into $L$ stacked Transformer modules, finally with an \gls{LLM} head. Each Transformer module has a multi-head self-attention (\glspl{RoPE} are applied to $\boldeq{Q}$ and $\boldeq{K}$) and a feed-forward network in parallel. We assume that the input sequence is $ \mathbb{\mathbb{R}}^{n\times \mathrm{hidden\_size}}$ and the batch size is $1$.

For $\mathrm{\underset{\mathrm{GPT\text{-}J}}{\textbf{Baseline}}}$ (direct inference), the \gls{LLM} is loaded with parameters (mainly weight memory). For dynamic memory, mainly the generated activations, we first only need to consider the maximum peak memory of one Transformer module. Then, we need to compare the maximum peak memory for Transformer modules and the peak memory in \gls{LLM} head. 

For the maximum peak memory of one Transformer module, we add the maximum peak memory in the multi-head self-attention and the maximum peak memory in the feed-forward network together. In the multi-head self-attention, we need to compare the attention matrix and $\boldeq{Q,K,V}$. In the feed-forward network, the layers with $\boldeq{W}_{in}$ or $\boldeq{W}_{out}$, as well as the activation function (GELU Activation for GPT-J), have the same maximum activations, i.e., in $\mathbb{\mathbb{R}}^{n\times (4\times\mathrm{hidden\_size})}$. Therefore, we have
\begin{align}
\underset{\mathrm{one\_Transformer\_module}}{\mathrm{max\_baseline}}\leftarrow& \mathrm{max}\Big\{\underset{\mathrm{attention\,matrix}}{n\times n\times\mathrm{attention\_head}},\underset{\boldeq{Q,K,V}}{3\times (n\times\mathrm{hidden\_size})}\Big\}\nonumber \\ &+\underset{\mathrm{feed\text{-}forward\,network}}{n\times (4\times\mathrm{hidden\_size})},
\nonumber \\
\mathrm{\underset{\mathrm{GPT\text{-}J}}{\textbf{Baseline}}}:&\,\underset{\mathrm{LLM}}{\mathrm{Parameters}}+\mathrm{max}\Big\{\underset{\mathrm{one\_Transformer\_module}}{\mathrm{max\_baseline}},\underset{\mathrm{LLM\,head}}{n\times\mathrm{vocab\_size}}\Big\}.
 \nonumber 
\end{align}

For $\mathrm{\underset{\mathrm{GPT\text{-}J}}{\textbf{LASER}}}$ \cite{sharma2024the}, in dynamic memory, we need to consider the \gls{SVD} decomposition $\boldeq{U}\boldeq{\Sigma} \boldeq{V}^T$ in feed-forward network. The introduced extra dynamic memory is in $\mathbb{\mathbb{R}}^{(5\times\mathrm{hidden\_size}+r)\times r}$ for both $ \boldeq{W}_{in} $ and $ \boldeq{W}_{out} $ in feed-forward network.
\begin{align}
\underset{\mathrm{one\_Transformer\_module}}{\mathrm{max\_LASER}}\leftarrow& \mathrm{max}\Big\{\underset{\mathrm{attention\,matrix}}{n\times n\times\mathrm{attention\_head}},\underset{\boldeq{Q,K,V}}{3\times (n\times\mathrm{hidden\_size})}\Big\} \nonumber \\
&+\underset{\mathrm{feed\text{-}forward\,network}}{n\times (4\times\mathrm{hidden\_size})}+\underset{\mathrm{SVD}}{(5\times\mathrm{hidden\_size}+r)\times r},
\nonumber \\
\mathrm{\underset{\mathrm{GPT\text{-}J}}{\textbf{LASER}}}:&\,\underset{\mathrm{LLM}}{\mathrm{Parameters}}+\mathrm{max}\Big\{\underset{\mathrm{one\_Transformer\_module}}{\mathrm{max\_LASER}},\underset{\mathrm{LLM\,head}}{n\times\mathrm{vocab\_size}}\Big\}.
 \nonumber 
\end{align}

For $\mathrm{\underset{\mathrm{GPT-J}}{\textbf{Ours}}}$, in dynamic memory, the $m$ \gls{SVD} components introduced extra memory in $\mathbb{\mathbb{R}}^{m\times n\times (4\times\mathrm{hidden\_size})}$. Therefore, we have
\begin{align}
\underset{\mathrm{one\_Transformer\_module}}{\mathrm{max\_ours}}\leftarrow& \mathrm{max}\Big\{\underset{\mathrm{attention\,matrix}}{n\times n\times\mathrm{attention\_head}},\underset{\boldeq{Q,K,V}}{3\times (n\times\mathrm{hidden\_size})}\Big\} \nonumber \\
&+\underset{\substack{\mathrm{feed\text{-}forward\,network\,with}\\m\,\mathrm{singular\,components}}}{(m+1)\times n\times (4\times\mathrm{hidden\_size})}+\underset{\mathrm{SVD}}{(5\times\mathrm{hidden\_size}+r)\times r},
\nonumber \\
\mathrm{\underset{\mathrm{GPT-J}}{\textbf{Ours}}}:&\,\underset{\mathrm{LLM}}{\mathrm{Parameters}}+\mathrm{max}\Big\{\underset{\mathrm{one\_Transformer\_module}}{\mathrm{max\_ours}},\underset{\mathrm{LLM\,head}}{n\times\mathrm{vocab\_size}}\Big\}.
 \nonumber 
\end{align}

For $\mathrm{\underset{\mathrm{GPT\text{-}J}}{\textbf{LASER-Grad}}}$ \cite{sreeram2025compress}, the generated activations, the SVD decomposition $\boldeq{U}\boldeq{\Sigma} \boldeq{V}^T$, and \gls{LLM} parameters are required to be stored in memory to get the gradient. Although parallelism of the multi-head self-attention and the feed-forward network is attainable together, we need to include both of them. In this case, we also need to include the intermediate generated activations. More specifically, when the input sequence is in $ \mathbb{\mathbb{R}}^{n\times \mathrm{hidden\_size}}$, for each Transformer module, there are $12\times$ (considering $\boldeq{Q},\boldeq{K},\boldeq{V}$, rotary position embeddings, LayerNorm, activation function, the input to each Transformer module, and the output of multi-head self-attention) the memory of the input sequence, adding with $2\times$ the memory of the attention weights (due to Softmax). For the feed-forward network with $\boldeq{W}_{in/out}$, based on the memory analysis in \cite{zhang2024revisiting}, we assume that the generated activations in the forward pass and the SVD decomposition could be released once the gradient of weights $\boldeq{W}$ and the corresponding singular values $\boldeq{\Sigma}$ in the backward pass are available. Therefore, we get the maximum memory for the layers with $\boldeq{W}_{in/out} $ in the feed-forward network, as denoted as follows, 
\begin{align}
\underset{\boldeq{W}_{in}}{\mathrm{max\_LASER\text{-}Grad}}\leftarrow & \mathrm{max}\Big\{ \underset{\mathrm{feed\text{-}forward\,network}}{n\times (4\times\mathrm{hidden\_size})}+\underset{\mathrm{SVD}}{(5\times\mathrm{hidden\_size}+r)\times r},\nonumber \\ &\underset{\mathrm{gradient\,of\,}\boldeq{W}_{in}}{(4\times\mathrm{hidden\_size})\times \mathrm{hidden\_size}}+\underset{\mathrm{gradient\,of\,}\boldeq{\Sigma}}{r\times r}\Big\}, \nonumber \\
\underset{\boldeq{W}_{out}}{\mathrm{max\_LASER\text{-}Grad}}\leftarrow & \mathrm{max}\Big\{ \underset{\mathrm{feed\text{-}forward\,network}}{n\times\mathrm{hidden\_size}}+\underset{\mathrm{SVD}}{(5\times\mathrm{hidden\_size}+r)\times r},\nonumber \\ &\underset{\mathrm{gradient\,of\,}\boldeq{W}_{out}}{\mathrm{hidden\_size}\times (4\times\mathrm{hidden\_size})}+\underset{\mathrm{gradient\,of\,}\boldeq{\Sigma}}{r\times r}\Big\}. \nonumber
\end{align}
Overall, the generated activations and the \gls{SVD} decomposition, as well as the required gradients for each Transformer module, are:
\begin{align}
\underset{\mathrm{one\_Transformer\_module}}{\mathrm{sum\_LASER\text{-}Grad}}\leftarrow& \mathrm{sum}\Big\{\underset{\mathrm{intermediate\,activations}}{12\times n\times\mathrm{hidden\_size}}+\underset{\mathrm{attention\,matrix\,with\,softmax}}{2\times n\times n\times\mathrm{attention\_head}}+\nonumber \\ &\underset{\boldeq{W}_{in}}{\mathrm{max\_LASER\text{-}Grad}}+\underset{\boldeq{W}_{out}}{\mathrm{max\_LASER\text{-}Grad}}\Big\},
\nonumber
\end{align}
Furthermore, the embedding layer generates $1\times$ the memory of the input sequence (because of only input embedding for GPT-J). We also need to consider the generated activations in the \gls{LLM} head, which is the output over the vocabulary. 
\begin{align}
\mathrm{\underset{\mathrm{GPT-J}}{\textbf{LASER-Grad}}}:&\,\underset{\mathrm{LLM}}{\mathrm{Parameters}}+\underset{\mathrm{embedding\,layer}}{1\times n\times\mathrm{hidden\_size}}+L\times\Big\{\underset{\mathrm{one\_Transformer\_module}}{\mathrm{sum\_LASER\text{-}Grad}}\Big\}\nonumber \\ &+\underset{\mathrm{LLM\,head}}{n\times\mathrm{vocab\_size}}.
 \nonumber 
\end{align}

\section{Considering More Weight Matrices in the Transformer Architecture}
\label{sec:appendix_more_weight}

\cite{sharma2024the} and \cite{sreeram2025compress} have presented that the weight matrices in the attention layers contribute to very little performance increase compared to the weight matrices in the feed-forward networks on GPT-J. To consider more weight matrices in the Transformer architecture, we also add the weight matrices of key projection $\boldeq{W}_{\mathrm{attention}\_{k}}$, query projection $\boldeq{W}_{\mathrm{attention}\_{q}}$, value projection $\boldeq{W}_{\mathrm{attention}\_{v}}$, and out projection $\boldeq{W}_{\mathrm{attention}\_{out}}$ in the attention layers to $\boldeq{W}_{in/out}$. 

We apply our proposed forward-only search across eight datasets on Roberta and only select the top-$5$ weight matrices for the subsequent rank-reduction strategy. Now, we have 6 types of weight matrices, and there are 12 layers of Transformer modules on Roberta. Therefore, there are a total of $6\times12-5=67$ weight matrices that are abandoned, which are not considered for the subsequent rank-reduction strategy. We take the Bios Profession dataset as an example, and the top-$5$ weight matrices are $ \left \{[\boldeq{W}_{out},11],[\boldeq{W}_{in},0],[\boldeq{W}_{in},1],[\boldeq{W}_{in},9],[\boldeq{W}_{in},3]\right \} $, which is the same as the weight matrices in Table \ref{tab:subset_roberta}. All the weight matrices for $\boldeq{W}_{\mathrm{attention}\_{k}}$, $\boldeq{W}_{\mathrm{attention}\_{q}}$, $\boldeq{W}_{\mathrm{attention}\_{v}}$, $\boldeq{W}_{\mathrm{attention}\_{out}}$ are in the abandoned 67 weight matrices. This example shows that when considering more weight matrices in the Transformer architecture, our forward-only approach can still get the same weight matrices as the case of only considering $\boldeq{W}_{in/out}$.

\begin{lstlisting}[mathescape=true]
#Top-5 $\color{blue}{\boldeq{W}}$ selected by our Algorithm 1
{[$\boldeq{W}_{out}$,11], [$\boldeq{W}_{in}$,0], [$\boldeq{W}_{in}$,1], [$\boldeq{W}_{in}$,9], [$\boldeq{W}_{in}$,3]}

#The abandoned 67 $\color{blue}{\boldeq{W}}$ that are not considered
{[$\boldeq{W}_{in}$,2], [$\boldeq{W}_{in}$,5], [$\boldeq{W}_{in}$,6], [$\boldeq{W}_{in}$,4], [$\boldeq{W}_{in}$,8], [$\boldeq{W}_{\mathrm{attention}\_{out}}$,0], [$\boldeq{W}_{in}$,7], [$\boldeq{W}_{\mathrm{attention}\_{out}}$,1], [$\boldeq{W}_{in}$,10], [$\boldeq{W}_{\mathrm{attention}\_{out}}$,11], [$\boldeq{W}_{\mathrm{attention}\_{out}}$,10], [$\boldeq{W}_{\mathrm{attention}\_{k}}$,9], [$\boldeq{W}_{in}$,11], [$\boldeq{W}_{\mathrm{attention}\_{out}}$,9], [$\boldeq{W}_{\mathrm{attention}\_{k}}$,3], [$\boldeq{W}_{\mathrm{attention}\_{out}}$,8], [$\boldeq{W}_{\mathrm{attention}\_{out}}$,2], [$\boldeq{W}_{\mathrm{attention}\_{out}}$,5], [$\boldeq{W}_{\mathrm{attention}\_{out}}$,7], [$\boldeq{W}_{\mathrm{attention}\_{k}}$,10], [$\boldeq{W}_{\mathrm{attention}\_{k}}$,5], [$\boldeq{W}_{\mathrm{attention}\_{k}}$,8], [$\boldeq{W}_{\mathrm{attention}\_{out}}$,6], [$\boldeq{W}_{\mathrm{attention}\_{q}}$,3], [$\boldeq{W}_{\mathrm{attention}\_{k}}$,7], [$\boldeq{W}_{\mathrm{attention}\_{k}}$,6], [$\boldeq{W}_{\mathrm{attention}\_{out}}$,3], [$\boldeq{W}_{\mathrm{attention}\_{q}}$,9], [$\boldeq{W}_{\mathrm{attention}\_{q}}$,11], [$\boldeq{W}_{out}$,5], [$\boldeq{W}_{\mathrm{attention}\_{out}}$,4], [$\boldeq{W}_{\mathrm{attention}\_{q}}$,4], [$\boldeq{W}_{\mathrm{attention}\_{k}}$,4], [$\boldeq{W}_{out}$,7], [$\boldeq{W}_{\mathrm{attention}\_{k}}$,11], [$\boldeq{W}_{\mathrm{attention}\_{q}}$,6], [$\boldeq{W}_{out}$,6], [$\boldeq{W}_{out}$,10], [$\boldeq{W}_{out}$,3], [$\boldeq{W}_{\mathrm{attention}\_{q}}$,8], [$\boldeq{W}_{\mathrm{attention}\_{q}}$,1], [$\boldeq{W}_{out}$,8], [$\boldeq{W}_{\mathrm{attention}\_{k}}$,0], [$\boldeq{W}_{out}$,1], [$\boldeq{W}_{out}$,2], [$\boldeq{W}_{\mathrm{attention}\_{q}}$,0], [$\boldeq{W}_{out}$,0], [$\boldeq{W}_{out}$,4], [$\boldeq{W}_{\mathrm{attention}\_{q}}$,2], [$\boldeq{W}_{\mathrm{attention}\_{q}}$,7], [$\boldeq{W}_{\mathrm{attention}\_{q}}$,5], [$\boldeq{W}_{\mathrm{attention}\_{k}}$,1], [$\boldeq{W}_{\mathrm{attention}\_{k}}$,2], [$\boldeq{W}_{\mathrm{attention}\_{v}}$,9], [$\boldeq{W}_{\mathrm{attention}\_{v}}$,10], [$\boldeq{W}_{out}$,9], [$\boldeq{W}_{\mathrm{attention}\_{q}}$,10], [$\boldeq{W}_{\mathrm{attention}\_{v}}$,8], [$\boldeq{W}_{\mathrm{attention}\_{v}}$,11], [$\boldeq{W}_{\mathrm{attention}\_{v}}$,3], [$\boldeq{W}_{\mathrm{attention}\_{v}}$,1], [$\boldeq{W}_{\mathrm{attention}\_{v}}$,6], [$\boldeq{W}_{\mathrm{attention}\_{v}}$,2], [$\boldeq{W}_{\mathrm{attention}\_{v}}$,4], [$\boldeq{W}_{\mathrm{attention}\_{v}}$,0], [$\boldeq{W}_{\mathrm{attention}\_{v}}$,7],[$\boldeq{W}_{\mathrm{attention}\_{v}}$,5]}
\end{lstlisting}

We further extend these experiments of considering more weight matrices for all eight datasets, as presented in Table \ref{tab:different_weights}. \ding{52} means the top-$j$ weight matrices are identical and preserve the same ordering as the weight matrices when only considering $\boldeq{W}_{in/out}$. The results show that our forward-only search can get the same top-$j$ weight matrices for all eight datasets. $\boldeq{W}_{\mathrm{attention}\_{k}}$, $\boldeq{W}_{\mathrm{attention}\_{q}}$, $\boldeq{W}_{\mathrm{attention}\_{v}}$, $\boldeq{W}_{\mathrm{attention}\_{out}}$ in the attention layers can be excluded from the search. Furthermore, the same top-$j$ weight matrices incur the same accuracy %as in Table \ref{tab:roberta_main} 
over eight datasets, which further justifies the effectiveness of our forward-only search. 

\begin{table}[!ht]
  \caption{Weight matrices of considering the weight matrices $\boldeq{W}_{\mathrm{attention}\_{k}}$, $\boldeq{W}_{\mathrm{attention}\_{q}}$, $\boldeq{W}_{\mathrm{attention}\_{v}}$, and $\boldeq{W}_{\mathrm{attention}\_{out}}$ in the attention layers, as well as $\boldeq{W}_{in/out}$. Our forward-only search can get the same top-$j$ weight matrices as the case of only considering $\boldeq{W}_{in/out}$.}
  \label{tab:different_weights}
  \centering
  \begin{tabular}{lcc}
    \toprule
    Considering $\boldeq{W}_{\mathrm{attention}\_{q/k/v/out}}$, and $\boldeq{W}_{in/out}$   & Same as only $\boldeq{W}_{in/out}$?  \\
    \midrule
    CounterFact &\ding{52} \\
    HotPotQA  &\ding{52}  \\
    FEVER   &\ding{52}  \\
    Bios Gender & \ding{52}  \\
    Bios Profession &\ding{52}  \\
    TruthfulQA  &\ding{52}  \\
    Epistemic Reasoning  & \ding{52}  \\
    WikidataQA  & \ding{52} \\
    \bottomrule
  \end{tabular}
\end{table}

\section{Our Forward-Only Approach Enhances Fine-Tuning}
\label{sec:appendix_finetuning}

To investigate the relationship between our forward-only approach and fine-tuning methods, we conducted various fine-tuning experiments with Roberta on FEVER and Bios Gender. These variants include fine-tuning the final candidate found by our forward-only approach, the worst candidate found by our forward-only approach, and all the $\boldeq{W}_{in}$ and all the $\boldeq{W}_{out}$ across twelve-layer Transformer modules. In the final candidate and the worst candidate found by our forward-only approach, we further test the cold start (fine-tuning the weight matrix) and warm start (fine-tuning the weight matrix with the corresponding rank reduction). We exploit $\mathcal{D}_{\mathrm{std}}$ (20\% of $\mathcal{D}$) as training data, while the rest 80\% of $\mathcal{D}$ as test data. AdamW optimizer and learning rate scheduler are exploited. Hyperparameters are listed in Table \ref{tab:hyperpara}. 

As shown in Table \ref{tab:finetuning_fullepoch}, for FEVER, fine-tuning the final candidate found by our forward-only approach (only 1.89\% of the total parameters are trained) achieves the accuracy of 75.2\text{\scriptsize$\pm$0.14}\%, while fine-tuning all the $\boldeq{W}_{in}$ (22.7\% of the total parameters are trained) achieves the accuracy of 76.5\text{\scriptsize$\pm$0.16}\%. Similarly, fine-tuning all the $\boldeq{W}_{out}$ achieves the accuracy of 76.5\text{\scriptsize$\pm$0.16}\%. For Bios Gender, fine-tuning the final candidate achieves the accuracy of 95.1\text{\scriptsize$\pm$0.01}\%, while fine-tuning all the $\boldeq{W}_{in}$ achieves the accuracy of 95.6\text{\scriptsize$\pm$0.02}\%, and fine-tuning all the $\boldeq{W}_{out}$ achieves the accuracy of 95.6\text{\scriptsize$\pm$0.01}\%. These results show that the final candidate found by our approach is effective for fine-tuning and can achieve comparable accuracy with $12\times$ fewer trainable parameters than fine-tuning all $\boldeq{W}_{in}$ or all $\boldeq{W}_{out}$. 

\begin{table}[!ht]
  \caption{Hyperparameters for fine-tuning experiments in Table \ref{tab:finetuning_fullepoch} and Table \ref{tab:finetuning}.}
  \label{tab:hyperpara}
  \centering
  \resizebox{\columnwidth}{!}{
  \setlength{\tabcolsep}{3pt}
\begin{threeparttable}
  \begin{tabular}{lcccccc}
    \toprule
    Dataset   & Setting & Batch Size & Fine-Tuning Epoch & Learning Rate & Weight Decay   \\
    \midrule
    \multirow{2}{*}{FEVER}& Cold Start/Warm Start (Table \ref{tab:finetuning}) & 256 & 1 & 6e-5 & 0.01   \\
    & Fine-Tuning\tnote{${\ddagger}\,$} (Table \ref{tab:finetuning_fullepoch}) & 256&50&6e-5&0.01 \\
    \midrule
     \multirow{2}{*}{Bios Gender} & Cold Start/Warm Start (Table \ref{tab:finetuning}) & 256 & 1 & 2e-6 & 0.01   \\
    & Fine-Tuning\tnote{${\ddagger}\,$} (Table \ref{tab:finetuning_fullepoch}) & 256&10&2e-6&0.01 \\
    \bottomrule
  \end{tabular}
  \begin{tablenotes}
\footnotesize
\item[]Note that in Fine-Tuning${\ddagger}$, fine-tuning the final candidate has $12\times$ less trainable parameters compared to all $\boldeq{W}_{in}$ or all $\boldeq{W}_{out}$. More fine-tuning epochs tend to improve the accuracy of fine-tuning with relatively fewer parameters, e.g., 20 epochs for Bios Gender with fine-tuning the final candidate achieving the accuracy of 95.1\text{\scriptsize$\pm$0.01}\%.
\end{tablenotes}
  \end{threeparttable}}
\end{table}

\begin{table}[!ht]
  \caption{Fine-Tuning Roberta with multiple epochs across multiple random seeds.}
  \label{tab:finetuning_fullepoch}
  \centering
\resizebox{\columnwidth}{!}{
\setlength{\tabcolsep}{2pt}
  \begin{tabular}{lcccccccc}
    \toprule
    \multirow{3}{*}{Dataset} &\multicolumn{4}{c}{FEVER}  & \multicolumn{4}{c}{Bios Gender} \\
    \cmidrule(lr){2-5}
    \cmidrule(lr){6-9}
    &\multicolumn{3}{c}{Fine-Tuning }&&\multicolumn{3}{c}{Fine-Tuning }&  \\
    \cmidrule(lr){2-4}
    \cmidrule(lr){6-8}
    &Final Candidate (50 epochs)& \multirow{2}{*}{All $\boldeq{W}_{in}$}&\multirow{2}{*}{All $\boldeq{W}_{out}$}& \multirow{2}{*}{Ours (std)}&Final Candidate (10 epochs)& \multirow{2}{*}{All $\boldeq{W}_{in}$}&\multirow{2}{*}{All $\boldeq{W}_{out}$}& \multirow{2}{*}{Ours (std)}\\
    &Final Candidate (100 epochs)&&&&Final Candidate (20 epochs)&&&\\
    \midrule
    \rowcolor{mygreen!60}  Trainable Params$\downarrow$ &1.89\%&\multicolumn{2}{c}{22.7\%} &- & 1.89\%&\multicolumn{2}{c}{22.7\%} &-\\
    \multirow{2}{*}{Acc}  & 74.6\text{\scriptsize$\pm$0.24}&\multirow{2}{*}{76.5\text{\scriptsize$\pm$0.16}}&\multirow{2}{*}{76.5\text{\scriptsize$\pm$0.16}}&\multirow{2}{*}{52.5}&94.2\text{\scriptsize$\pm$0.02}&\multirow{2}{*}{95.6\text{\scriptsize$\pm$0.02}}&\multirow{2}{*}{95.6\text{\scriptsize$\pm$0.01}}&\multirow{2}{*}{93.4}\\
    &(75.2\text{\scriptsize$\pm$0.14})&&&&(95.1\text{\scriptsize$\pm$0.01})&&&\\
    \bottomrule
  \end{tabular}}
\end{table}
As presented in Table \ref{tab:finetuning}, the warm start with the final candidate performs better than the cold start in one-epoch fine tuning (one forward pass and one backward pass), while our forward-only \camera{search} (two forward passes) performs comparably. For Roberta on Bios Gender, the final candidate is $[\boldeq{W}_{in},9,90\%]$ and the worst candidate is $[\boldeq{W}_{in},1,1\%]$. 

\begin{figure}[!ht]
 \centering
    \includegraphics[width=0.5\linewidth]{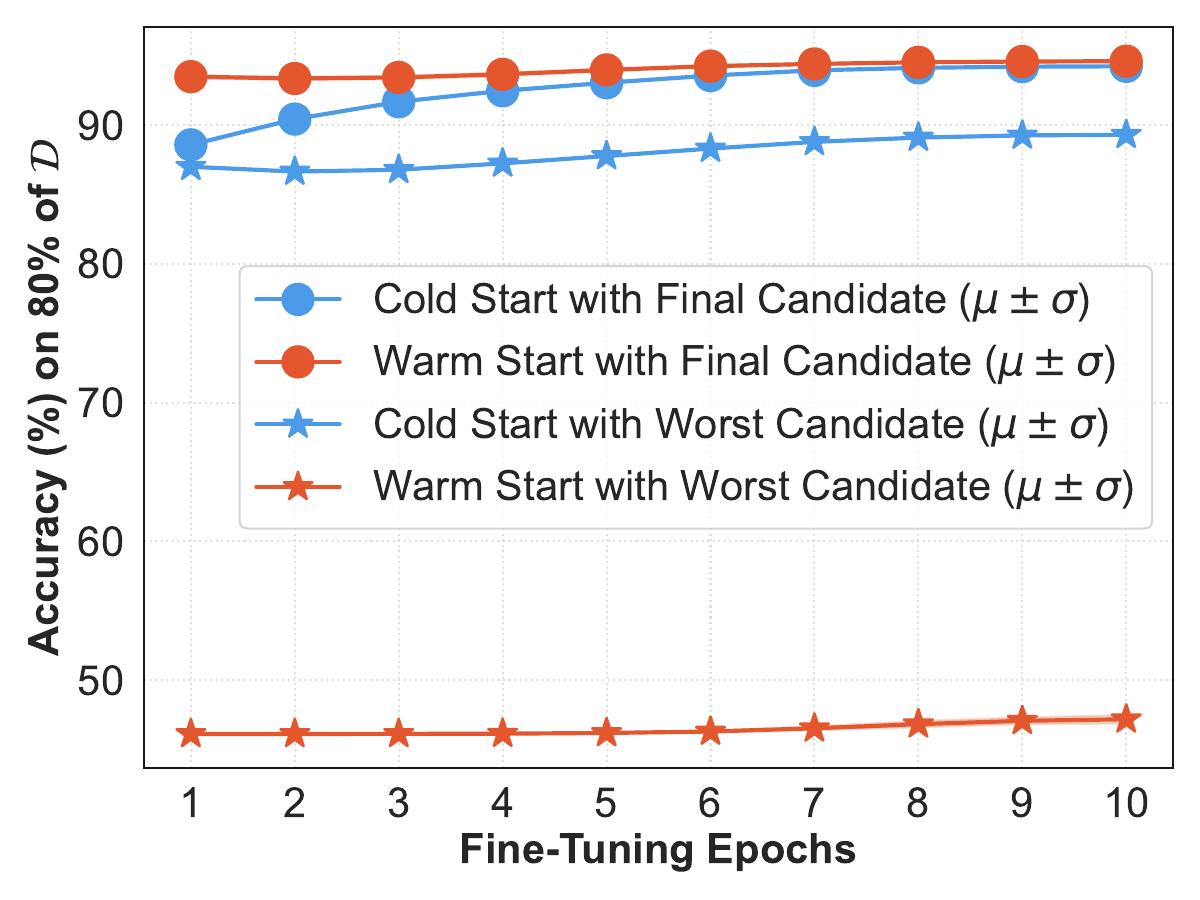}
  \caption{Fine-tuning the final candidate with the warm start performs the best.}
  \label{fig:candidate_fine-tuning}
\end{figure}

Besides, as shown in Fig. \ref{fig:candidate_fine-tuning}, fine-tuning the final candidate found by our forward-only approach with the warm start performs better than its cold start, especially at the early fine-tuning epochs. While for the worst candidate found by our forward-only approach, fine-tuning the worst candidate with the cold start only gets the accuracy of 89.3\text{\scriptsize$\pm$0.03}\%, which is significantly worse than the final candidate with the cold start. At the same time, fine-tuning the worst candidate with the warm start even performs significantly worse than its cold start, which demonstrates the effectiveness of our forward-only approach. Overall, these results consistently present that our forward-only approach can enhance the fine-tuning process.

\begin{table}[!ht]
  \caption{Fine-Tuning Roberta with one epoch across multiple random seeds.}
  \label{tab:finetuning}
  \centering
\resizebox{\columnwidth}{!}{
  \begin{tabular}{lcccccc}
    \toprule
    \multirow{3}{*}{Dataset} &\multicolumn{3}{c}{FEVER}  & \multicolumn{3}{c}{Bios Gender} \\
    \cmidrule(lr){2-4}
    \cmidrule(lr){5-7}
    & \multicolumn{2}{c}{Fine-Tuning Final Candidate} &  \multirow{2}{*}{Ours (std)} & \multicolumn{2}{c}{Fine-Tuning  Final Candidate} & \multirow{2}{*}{Ours (std)} \\
    \cmidrule(lr){2-3}
    \cmidrule(lr){5-6}
    & Cold Start & Warm Start & &Cold Start&Warm Start \\
    \midrule
    Acc  &50.00\text{\scriptsize$\pm$0.00} & \textbf{53.35\text{\scriptsize$\pm$0.17}} & 52.5 & 88.58\text{\scriptsize$\pm$0.03} & \textbf{93.50\text{\scriptsize$\pm$0.11}} & 93.4  \\
    \bottomrule
  \end{tabular}}
\end{table}

%%%%%%%%%%%%%%%%%%%%%%%%%%%%%%%%%%%%%%%%%%%%%%%%%%%%%%%%%%%%

\section{Extended Experiments on Various LLMs}
\label{sec:appendix_other_llm}

We extend experiments to various \glspl{LLM} including LLaMA2 \cite{touvron2023llama} with gated feed-forward networks (Llama2-7B), Mistral \citep{jiang2023mistral7b} with \gls{GQA} and sliding-window attention (Mistral-7B), Qwen2 \citep{yang2024qwen2technicalreport} with instruction-tuning and multilingual capabilities (Qwen2-7B-Instruct), Qwen2 with \gls{MoE} (Qwen2-57B-A14B-Instruct), Gemma \citep{team2024gemma} (Gemma-7B), and DeepSeek \citep{bi2024deepseek} (DeepSeekV1-7B-Chat) with data engineering, as well as the latest Gemma 4 \citep{google_gemma4_2026} (Gemma4-E2B-it) with the hybrid attention and per-layer embeddings (PLE). %built-in reasoning mode. 
For Qwen2-57B-A14B-Instruct\footnote{For Qwen2-57B-A14B-Instruct, we use 2 NVIDIA Tesla A40 \gls{GPU} with 48GB \gls{RAM} for each \gls{GPU}.} with NormalFloat4 quantization \citep{dettmers2023qlora} and BFloat16 computation \citep{wang2019bfloat16}, we consider $\boldeq{W}_{in/out} $ in the shared experts of \gls{MoE}. The number of stacked Transformer modules is listed in Table \ref{tab:setting_other_llms}. The final candidates for LASER \citep{sharma2024the} and our forward-only approach on Epistemic Reasoning are presented in Table \ref{tab:para_otherllm}.
\begin{table}[!ht]
  \caption{The number of stacked Transformer modules for various \glspl{LLM}.}
  \label{tab:setting_other_llms}
  \centering
    \resizebox{\columnwidth}{!}{
  \begin{tabular}{ccccc}
    \toprule
    \multirow{2}{*}{\glspl{LLM}}     & LLaMA2-7B & Qwen2-7B-Instruct/Gemma-7B & \multirow{2}{*}{DeepSeekV1-7B-Chat} & \multirow{2}{*}{Gemma4-E2B-it}\\
    &Mistral-7B & Qwen2-57B-A14B-Instruct &  & \\ 
    \midrule
    $L$  &32 & 28 & 30 & 35\\
    \bottomrule
  \end{tabular}}
\end{table}

\begin{table}[!ht]
  \caption{Final candidates for LASER and our forward-only approach with various \glspl{LLM} on Epistemic Reasoning in Table \ref{tab:otherllms}, including $\boldeq{W}_{in/out}$ in the feed-forward network of Transformer architecture, the layer index $l$ of stacked Transformer modules, and the retained top-$k\%$ ranks ($0.5\% \le k\% \le 90\%$), as well as considering top-$j$ Weight Matrices and last-$m$-rank \gls{SVD} component.}
  \label{tab:para_otherllm}
  \centering
  \begin{tabular}{lcccc}
    \toprule
    LLMs   & LASER (std) & Ours (std) & top-$j$ & last-$m$ \\
    \midrule
    LLaMA2-7B  & $[\boldeq{W}_{out},28,1\%]$ & $[\boldeq{W}_{in},31,40\%]$ & top-$10$ & last-$50$ \\
    %DeepseekV1-7B-Base  & $[\boldeq{W}_{out},1,40\%]$ &  $[\boldeq{W}_{in},0,20\%]$ \\
    DeepseekV1-7B-Chat  & $[\boldeq{W}_{out},11,0.5\%]$ &  $[\boldeq{W}_{in},7,0.5\%]$ &top-$2$ & last-$20$ \\
    Mistral-7B & $[\boldeq{W}_{in},20,0.5\%]$  &  $[\boldeq{W}_{in},20,0.5\%]$ &top-$7$ & last-$20$ \\
    Qwen2-7B-Instruct & $[\boldeq{W}_{in},2,1\%]$ & $[\boldeq{W}_{in},2,1\%]$  &top-$\camera{2}$ & last-$20$\\
    Gemma-7B & $[\boldeq{W}_{out},3,\camera{5}\%]$ & $[\boldeq{W}_{in},\camera{7},20\%]$ &top-$\camera{5}$ & last-$\camera{2}0$\\
    Gemma4-E2B-it & $[\boldeq{W}_{in},2,1\%]$ & $[\boldeq{W}_{out},25,0.5\%]$&top-$6$ & last-$50$\\
    Qwen2-57B-A14B-Instruct &$[\boldeq{W}_{in},9,5\%]$  & $[\boldeq{W}_{in},9,5\%]$ &top-$4$ & last-$20$\\
    \bottomrule
  \end{tabular}
\end{table}

Furthermore, we also extend the experiments across various datasets. The final candidates for LASER \citep{sharma2024the} and our forward-only approach on LLaMA2 are presented in Tables \ref{tab:para_otherllm} and \ref{tab:para_llama}. The corresponding results in Acc and Speedup are presented in Table~\ref{tab:llama_main}. Our forward-only approach achieves a greater speedup, while ours (std) and ours (hierarchical) achieve greater average improvement in Acc compared to LASER. Overall, ours (hierarchical) achieves a balanced trade-off between average Speedup and average improvement in Acc across ours (std), ours (hierarchical), and ours (subset).

\begin{table}[!ht]
  \caption{Final candidates for our forward-only approach on LLaMA2 in Table \ref{tab:llama_main}, including $\boldeq{W}_{in/out}$ in the feed-forward network of Transformer architecture, the layer index $l$ ($0 \le  l \le 31$) of stacked Transformer modules, and the retained top-$k\%$ ranks ($0.5\% \le k\% \le 90\%$).}
  \label{tab:para_llama}
  \centering
    \resizebox{\columnwidth}{!}{
  \begin{tabular}{lcccc}
    \toprule
    Dataset   & Ours (subset) & Ours (std)  &  Ours (hierarchical)    & Hit?   \\
    \midrule
    CounterFact & $[\boldeq{W}_{in},9,10\%]$ & $[\boldeq{W}_{in},31,60\%]$ &  $[\boldeq{W}_{in},31,60\%]$& Yes \\
    HotPotQA & $[\boldeq{W}_{in},6,5\%]$ &  $[\boldeq{W}_{in},10,0.5\%]$ &  $[\boldeq{W}_{in},9,5\%]$&  No \\
    FEVER  & $[\boldeq{W}_{in},9,90\%]$ & $[\boldeq{W}_{in},30,20\%]$ &  $[\boldeq{W}_{in},9,0.5\%]$& No \\
    Bios Gender& $[\boldeq{W}_{in},7,0.5\%]$ & $[\boldeq{W}_{in},7,0.5\%]$ &  $[\boldeq{W}_{in},7,0.5\%]$& Yes\\
    Bios Profession& $[\boldeq{W}_{in},16,5\%]$ & $[\boldeq{W}_{in},16,0.5\%]$ &  $[\boldeq{W}_{in},16,0.5\%]$& Yes \\
    TruthfulQA & $[\boldeq{W}_{out},0,5\%]$ & $[\boldeq{W}_{in},30,10\%]$ &  $[\boldeq{W}_{in},30,1\%]$& No \\
    Epistemic Reasoning & $[\boldeq{W}_{in},16,1\%]$ & $[\boldeq{W}_{in},31,40\%]$ &  $[\boldeq{W}_{in},31,40\%]$&Yes  \\
    WikidataQA & $[\boldeq{W}_{out},0,80\%]$ & $[\boldeq{W}_{in},30,5\%]$ &  $[\boldeq{W}_{in},30,5\%]$& Yes \\
    \bottomrule
  \end{tabular}}
\end{table}

\begin{table}[!ht]
  \caption{LLaMA2 evaluation with reported accuracy (Acc, \%) and Speedup. Ours (hierarchical) achieves a balanced trade-off between average Speedup and average improvement in Acc across ours (std), ours (hierarchical), and ours (subset) settings.}
  \label{tab:llama_main}
  \centering

  {\setlength{\tabcolsep}{1pt}
  \begin{tabular}{lccc>{\columncolor{myblue!60}}cc>{\columncolor{myblue!60}}cc>{\columncolor{myblue!60}}c}
    \toprule
    & \multirow{3}{*}{Baseline} & \multirow{3}{*}{LASER}  &  \multicolumn{6}{c}{Ours} \\
    \cmidrule(lr){4-9}
    Dataset & &&\multicolumn{2}{c}{(std)}&\multicolumn{2}{c}{(hierarchical)}&\multicolumn{2}{c}{(subset)} \\
    \cmidrule(lr){2-2}
    \cmidrule(lr){3-3}
    \cmidrule(lr){4-5}
    \cmidrule(lr){6-7}
    \cmidrule(lr){8-9}
    &Acc&Acc&Acc&Speedup$\uparrow$&Acc&Speedup$\uparrow$&Acc&Speedup$\uparrow$\\
    \midrule
    CounterFact & 35.6& \textbf{37.6} &35.9&6.4$\times$&35.9&53.9$\times$&34.9&832$\times$\\
    HotPotQA  &16.5&17.2 &19.7&6.4$\times$&\textbf{20.5}&43.9$\times$&17.5&185$\times$\\
    FEVER   &59.3&\textbf{64.5} & \textbf{64.5} &6.4$\times$ &62.9 & 42.7$\times$& 59.9 &165$\times$\\
    Bios Gender  &75.5&88.4 &\textbf{91.5}&6.4$\times$&\textbf{91.5}&51.7$\times$&\textbf{91.5}&502$\times$\\
    Bios Profession &85.0&86.7 &\textbf{87.0}&6.4$\times$&\textbf{87.0}&46.6$\times$&85.7&243$\times$\\
    TruthfulQA  &50.5&\textbf{56.2} &\textbf{56.2}&6.4$\times$&\textbf{56.2}&32.5$\times$&53.7&74.5$\times$\\
    Epistemic Reasoning &44.8&63.4 &\textbf{63.5}&6.4$\times$&\textbf{63.5}&17.6$\times$&62.1&25.3$\times$\\
    WikidataQA  &59.5&\textbf{62.0} &60.9&6.4$\times$&60.9&47.1$\times$&59.8&257$\times$\\
    \midrule
    Average $\Delta$Acc$\uparrow$  & 0.00 & 6.16 & \multicolumn{2}{c}{\textbf{6.56}}  & \multicolumn{2}{c}{6.46}  & \multicolumn{2}{c}{4.80}   \\
    \rowcolor{myblue!60} Average Speedup$\uparrow$ & - & 1$\times$ & \multicolumn{2}{c}{6.4$\times$}& \multicolumn{2}{c}{42.0$\times$}&\multicolumn{2}{c}{\textbf{285$\times$}}\\
    \bottomrule
  \end{tabular}}
\end{table}

\clearpage
\section{Intuitive Explanation}
\label{sec:appendix_theory}
In Equation (\ref{eq:rank_forward}), with $\boldeq{W}\in \mathbb{\mathbb{R}}^{h\times d}= \boldeq{U}\boldeq{\Sigma}\boldeq{V}^T=\sum_{i=1}^r\sigma_{i}\boldeq{u}_i\otimes\boldeq{v}_i^T$ in full \gls{SVD}, we have
\begin{align}
\boldeq{X}\boldeq{W}^T = \sum_{i=1}^r\sigma_i\cdot(\boldeq{X}\boldeq{v}_i)\otimes\boldeq{u}_i^T=\sum_{i=1}^r\boldeq{Y}^{(i)}=\boldeq{Y}, \mathrm{with}\quad r=\min(h,d). \nonumber
\end{align}

The gradient of the loss function $\mathcal{L}$ with respect to $\boldeq{W}$ ($\nabla_{\boldeq{W}}\mathcal{L}$, simplified as $\nabla\boldeq{W}$), is backpropagated from the gradient of the loss function $\mathcal{L}$ with respect to $\boldeq{Y}$ ($\nabla_{\boldeq{Y}}\mathcal{L}$, simplified as $\nabla\boldeq{Y}$).
\begin{align}
\nabla\boldeq{W} &= (\nabla\boldeq{Y})^T\boldeq{X} = \sum_{p=1}^n(\nabla\boldeq{Y}_{p,:})^T\boldeq{X}_{p,:},
\nonumber 
\end{align}
where $n$ is the sequence length and $p$ is the $p$-th token in the sequence. Next, we have the gradient of $i$-th singular value $\nabla\sigma_i$ as follows,
\begin{align}
\nabla\sigma_i &= \boldeq{u}_i^T \nabla\boldeq{W} \boldeq{v}_i = \boldeq{u}_i^T (\nabla\boldeq{Y})^T\boldeq{X} \boldeq{v}_i = \sum_{p=1}^n  \boldeq{u}_i^T (\nabla\boldeq{Y}_{p,:})^T\boldeq{X}_{p,:} \boldeq{v}_i,\nonumber
\end{align}
where $\nabla\boldeq{Y}^{(i)}= \nabla\boldeq{Y}$. Moreover, we exploit the Frobenius norm of $\nabla\sigma_i$ and the linearity of the Frobenius inner product, and therefore we have:
\begin{align}
\left \| \nabla\sigma_i \right \|^2_F  &=\left \langle\sum_{p=1}^n  \boldeq{u}_i^T (\nabla\boldeq{Y}_{p,:})^T\boldeq{X}_{p,:} \boldeq{v}_i, \sum_{q=1}^n  \boldeq{u}_i^T (\nabla\boldeq{Y}_{q,:})^T\boldeq{X}_{q,:} \boldeq{v}_i  \right \rangle_F \nonumber\\
&= \sum_{p=1}^n\sum_{q=1}^n \left \langle \boldeq{u}_i^T (\nabla\boldeq{Y}_{p,:})^T\boldeq{X}_{p,:}\boldeq{v}_i, \boldeq{u}_i^T (\nabla\boldeq{Y}_{q,:})^T\boldeq{X}_{q,:}\boldeq{v}_i  \right \rangle_F, \label{eq:gradient_of_sigma}
\end{align}
where $p$ and $q$ denote the different tokens in the sequence. Moreover, utilizing the compatibility between the Frobenius inner product and the matrix outer product, for any $\boldeq{a}$ and $\boldeq{c}$ ($\boldeq{b}$ and $\boldeq{d}$) having the same dimension, we have the formula $
\left \langle \boldeq{a}\boldeq{b}^T, \boldeq{c}\boldeq{d}^T  \right \rangle_F = \mathrm{tr}\left((\boldeq{a}\boldeq{b}^T)^T(\boldeq{c}\boldeq{d}^T)\right)=(\boldeq{a}^T\boldeq{c})(\boldeq{b}^T\boldeq{d}),
$
where $\mathrm{tr}$ is the matrix trace operator. Then, Equation (\ref{eq:gradient_of_sigma}) can be simplified as follows:
\begin{align}
\left \| \nabla\sigma_i \right \|^2_F  &= \sum_{p=1}^n\sum_{q=1}^n\left( (\boldeq{X}_{p,:}\boldeq{v}_i)(\boldeq{X}_{q,:}\boldeq{v}_i)^T\right)\cdot \left( 
(\nabla\boldeq{Y}_{p,:}\boldeq{u}_i )(\nabla\boldeq{Y}_{q,:}\boldeq{u}_i )^T
\right). \nonumber
\end{align}
When considering $\boldeq{V}\in \mathbb{\mathbb{R}}^{d\times d}$ in full \gls{SVD} as the orthogonal matrix (i.e., $\boldeq{V}\boldeq{V}^T=\boldeq{I}$), for full ranks, the angle between $\boldeq{X}_{p,:}$ and $\boldeq{X}_{q,:}$ contributes as follows:
\begin{align}
\sum_{p=1}^n\sum_{q=1}^n\sum_{i=1}^d(\boldeq{X}_{p,:}\boldeq{v}_i)(\boldeq{X}_{q,:}\boldeq{v}_i)^T & = \sum_{p=1}^n\sum_{q=1}^n\boldeq{X}_{p,:}\left(\sum_{i=1}^d\boldeq{v}_i\boldeq{v}_i^T\right)(\boldeq{X}_{q,:})^T
\nonumber
\\ &= \sum_{p=1}^n\sum_{q=1}^n \boldeq{X}_{p,:}(\boldeq{X}_{q,:})^T \nonumber \\
&=
\sum_{p=1}^n\sum_{q=1}^n\| \boldeq{X}_{p,:} \ \|  \| \boldeq{X}_{q,:} \| \cos{\theta_{p,q}}, \label{eq:cos_similarity}
\end{align}
where $\cos{\theta_{p,q}}$ denotes the cosine similarity of the angle between $\boldeq{X}_{p,:}$ and $\boldeq{X}_{q,:}$. Therefore, given the \gls{LLM} with the deterministic \gls{SVD}, $\cos{\theta_{p,q}}$ will exert an influence on the gradient of each singular value among $(\nabla\sigma_1,\nabla\sigma_2,...,\nabla\sigma_r)$ by decomposing the angle between $\boldeq{X}_{p,:}$ and $\boldeq{X}_{q,:}$ into individual singular components with $(\boldeq{v}_1,\boldeq{v}_2,...,\boldeq{v}_d)$. This decomposition also applies for the backpropagated gradients with the $(\boldeq{u}_1,\boldeq{u}_2,...,\boldeq{u}_h)$ and the orthogonal matrix $\boldeq{U}\in \mathbb{\mathbb{R}}^{h\times h}$ (i.e., $\boldeq{U}\boldeq{U}^T=\boldeq{I}$) \footnote{\cite{wang2025angles}:``\textit{The number of gradient components received by a neuron is proportional to the frequency of its activation}'' and ``\textit{Tokens with higher angle concentration activate similar neurons due to shared value patterns}''. Moreover, ``\textit{the more concentrated the angles between tokens, the larger the gradient norm}''.}. 
\begin{align}
\sum_{p=1}^n\sum_{q=1}^n\sum_{i=1}^h (\nabla\boldeq{Y}_{p,:}\boldeq{u}_i) (\nabla\boldeq{Y}_{q,:}\boldeq{u}_i)^T = \sum_{p=1}^n\sum_{q=1}^n\nabla\boldeq{Y}_{p,:} (\nabla\boldeq{Y}_{q,:})^T.  \label{eq:gradient_y}
\end{align}

\begin{figure}[!ht]
  \centering
\begin{subfigure}{0.48\textwidth}
  \begin{subfigure}{0.48\textwidth}
    \includegraphics[width=\linewidth]{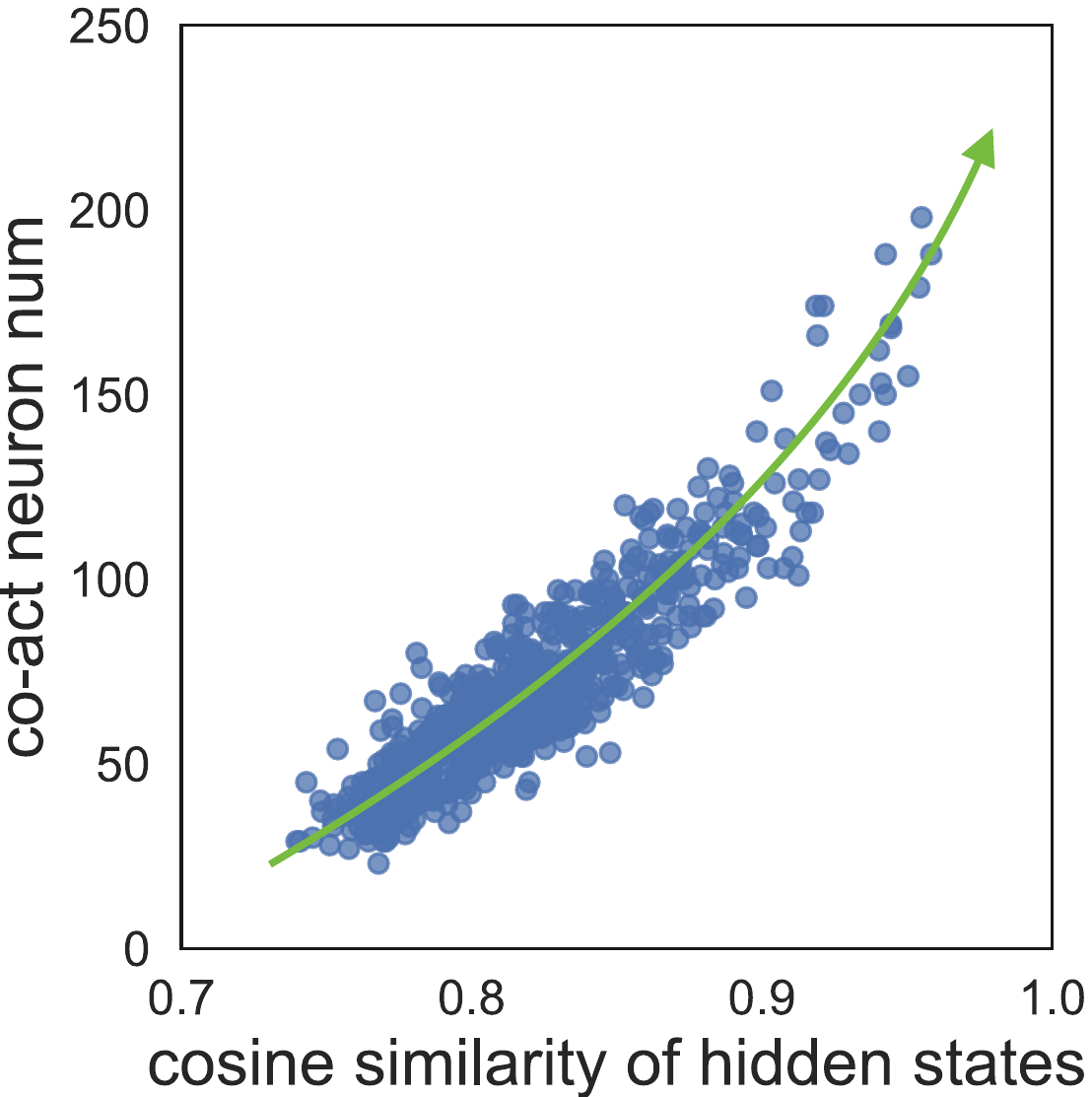}
  \end{subfigure}
  \begin{subfigure}{0.48\textwidth}
    \includegraphics[width=\linewidth]{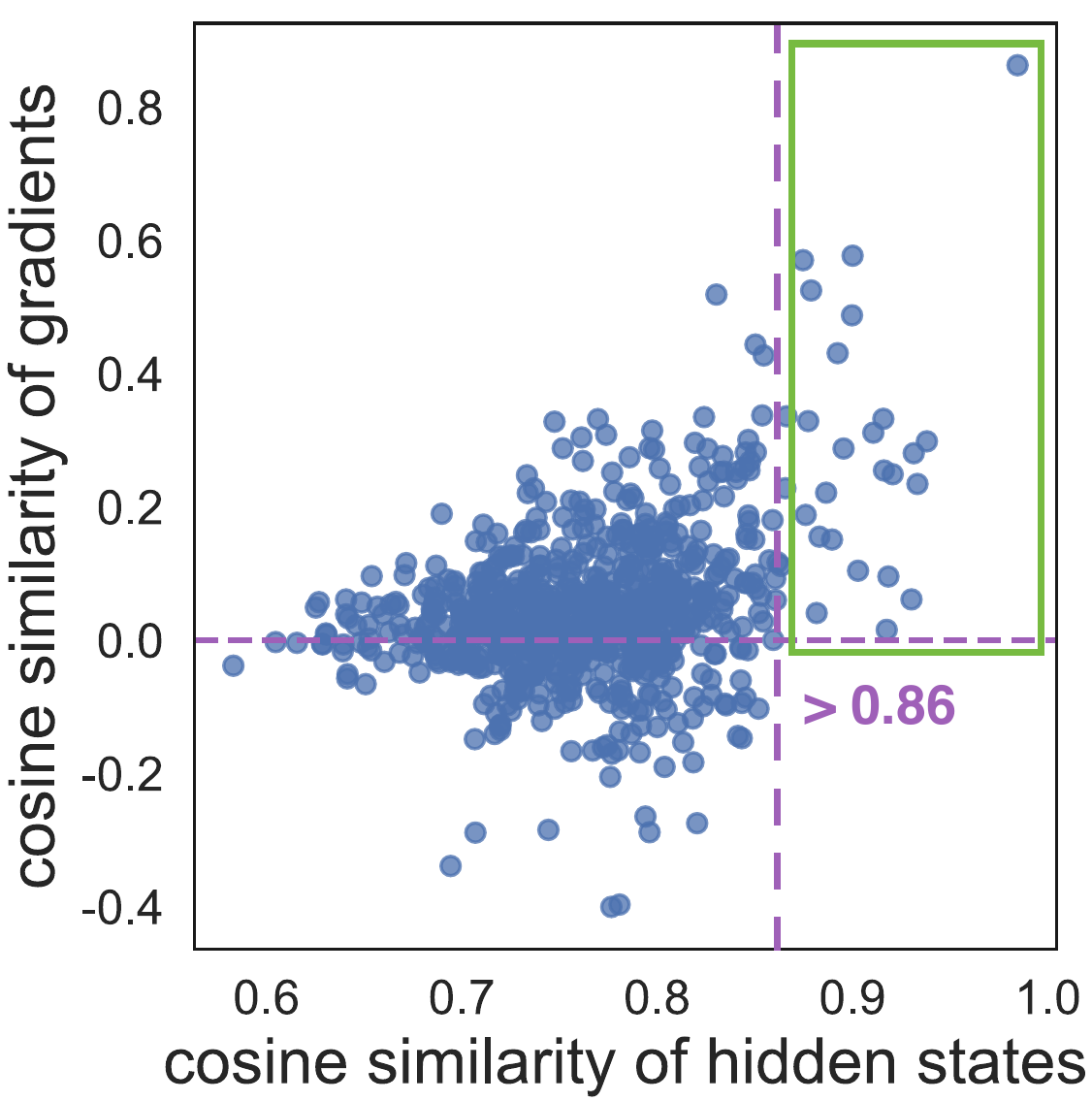}
  \end{subfigure}
  \caption{$\boldeq{W}_{in}$}
\end{subfigure}
  \hfill
\begin{subfigure}{0.48\textwidth}
  \begin{subfigure}{0.48\textwidth}
    \includegraphics[width=\linewidth]{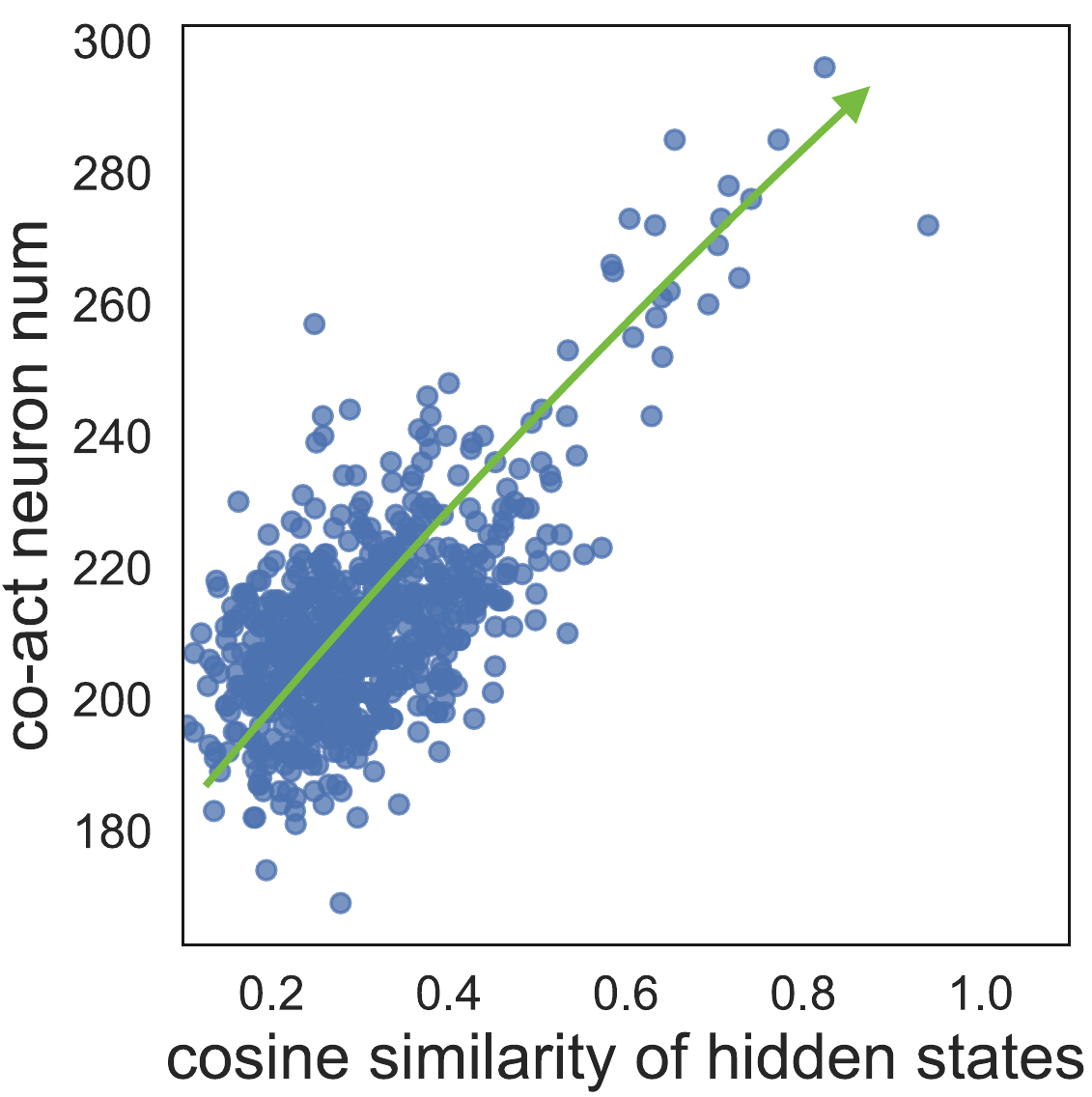}
  \end{subfigure}
  \begin{subfigure}{0.48\textwidth}
    \includegraphics[width=\linewidth]{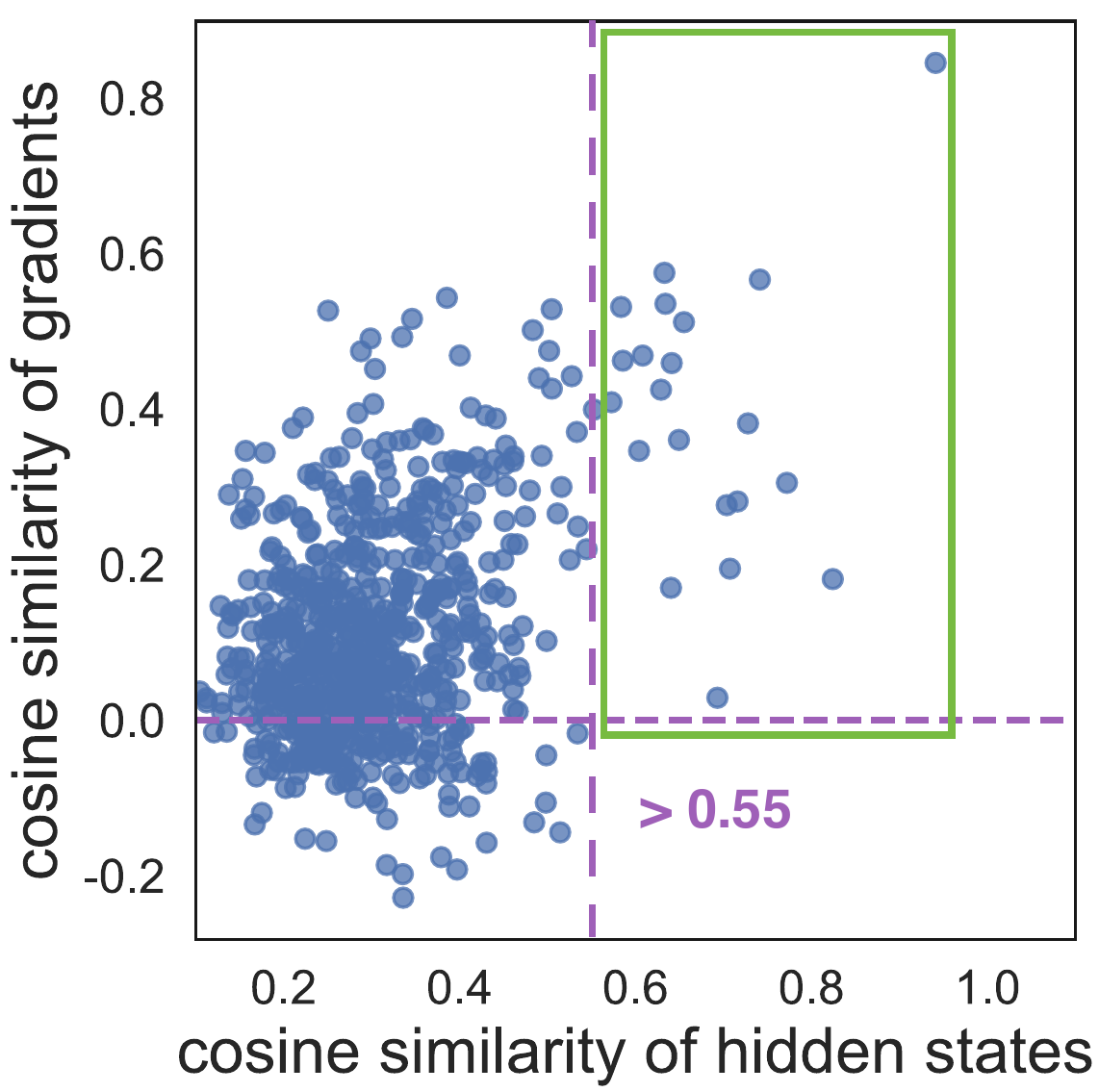}
  \end{subfigure}
  \caption{$\boldeq{W}_{out}$}
\end{subfigure}
\caption{Visualization of the relationship between cosine similarity of hidden states ($\cos{\theta_{p,q}}$) and the number of co-activated neurons, as well as $\cos{\theta_{p,q}}$ and cosine similarity of gradients ($\cos{\gamma_{p,q}}$) for (a) $\boldeq{W}_{in}$ (Layer 6) and (b) $\boldeq{W}_{out}$ (Layer 4) with Roberta on the FEVER dataset. More layers for $\boldeq{W}_{in}$ and $\boldeq{W}_{out}$ are detailed in Fig. \ref{fig:in_gradient_all} (a) and Fig. \ref{fig:out_gradient_all} (a).}
  \label{fig:cos_influence}
\end{figure}

To analyze Equation (\ref{eq:cos_similarity}) and Equation (\ref{eq:gradient_y}), we evaluate how the angle between $\boldeq{X}_{p,:}$ and $\boldeq{X}_{q,:}$ (which is available in the forward pass) influences the $\nabla\boldeq{Y}_{p,:} (\nabla\boldeq{Y}_{q,:})^T$ (which is computed in the backward pass). We exploit the $\nabla\boldeq{Y}_{p,:} (\nabla\boldeq{Y}_{q,:})^T = \| \nabla\boldeq{Y}_{p,:} \ \|  \| \nabla\boldeq{Y}_{q,:} \| \cos{\gamma_{p,q}}$, where $\gamma$ denotes the angle between $\nabla\boldeq{Y}_{p,:}$ and $\nabla\boldeq{Y}_{q,:}$. Taking the evaluation with Roberta on the FEVER dataset, we present the relationship between cosine similarity of hidden states ($\cos{\theta_{p,q}}$) and the number of co-activated neurons, as well as $\cos{\theta_{p,q}}$ and cosine similarity of gradients ($\cos{\gamma_{p,q}}$) in Fig. \ref{fig:cos_influence}. The results show that a higher $\cos{\theta_{p,q}}$ tends to activate the similar patterns among neurons, i.e., implying an increased number of co-activated neurons\footnote{This is also observed in \cite{wang2025angles}.}. At the same time, when $\cos{\theta_{p,q}}$ exceeds a certain threshold, $\cos{\gamma_{p,q}}$ tends to become positive. Therefore, a sufficiently large $\cos{\theta_{p,q}}$ tends to increase the $\left \| \nabla\boldeq{W} \right \|^2_F$\footnote{This is also observed in \cite{wang2025angles}.} because $\left \| \nabla\boldeq{W} \right \|^2_F = \sum_{p=1}^n\sum_{q=1}^n\| \boldeq{X}_{p,:} \ \|  \| \boldeq{X}_{q,:} \| \cos{\theta_{p,q}} \| \nabla\boldeq{Y}_{p,:} \ \|  \| \nabla\boldeq{Y}_{q,:} \| \cos{\gamma_{p,q}}$.

Therefore, this change in $\nabla\boldeq{W}$ induced by $\cos{\theta_{p,q}}$ tends to influence the $\nabla\sigma_i$ and therefore tends to re-allocate the updating power within different $\left \| \nabla\sigma_i \right \|^2_F$, as shown below,
\begin{align}
\left \| \nabla\boldeq{W} \right \|^2_F = \left \| \boldeq{U}^T\nabla\boldeq{W}\boldeq{V} \right \|^2_F\ge \sum_{i=1}^r\left \| \nabla\sigma_i \right \|^2_F \quad \mathrm{with} \quad  \nabla\sigma_i =\boldeq{u}_i^T \nabla\boldeq{W} \boldeq{v}_i = \langle \nabla\boldeq{W}, \boldeq{u}_i \boldeq{v}_i^T \rangle_F, \nonumber
\end{align}
where $\langle \boldeq{u}_i \boldeq{v}_i^T, \boldeq{u}_j \boldeq{v}_j^T \rangle_F = 0$ for $i\ne j$ and $i,j \in \{1, 2, \dots, r\}$ because $\langle \boldeq{u}_i \boldeq{v}_i^T, \boldeq{u}_j \boldeq{v}_j^T \rangle_F = (\boldeq{u}_i^T\boldeq{u}_j)(\boldeq{v}_i^T\boldeq{v}_j)$ and the orthogonality of $\boldeq{U}$ and $\boldeq{V}^T$ matrices in full \gls{SVD}. Besides, this study \citep{kaushik2026do} shows that deep neural networks converge to shared subspaces \camera{and this study \citep{sun2026unleashing} shows that fine-tuning the top singular values is effective, as well as this study \citep{li2024basis} that fine-tunes the singular values and keeps the most important components/bases. Potentially, given this heuristic explanation,} %$\left \| \nabla\boldeq{W} \right \|^2_F \approx  \sum_{i=1}^r\left \| \nabla\sigma_i \right \|^2_F$, 
the increased $\left \| \nabla\boldeq{W} \right \|^2_F$ tends to amplify/attenuate the updating power of different $\left \| \nabla\sigma_i \right \|^2_F$. In this case, a higher $\cos{\theta_{p,q}}$ tends to have more effective updating power, improving adaptation to downstream tasks. 

Overall, we provide some intuition for the relationship between angle concentration and the gradients of singular values. Angle concentration is available during the forward pass, unlike gradients computed in the backward pass, making it well suited for \gls{LLM} adaptation in \gls{LLM} serving.

\begin{figure}[H]
 \centering
   \begin{subfigure}{\textwidth}
    \centering
\includegraphics[width=0.8\linewidth]{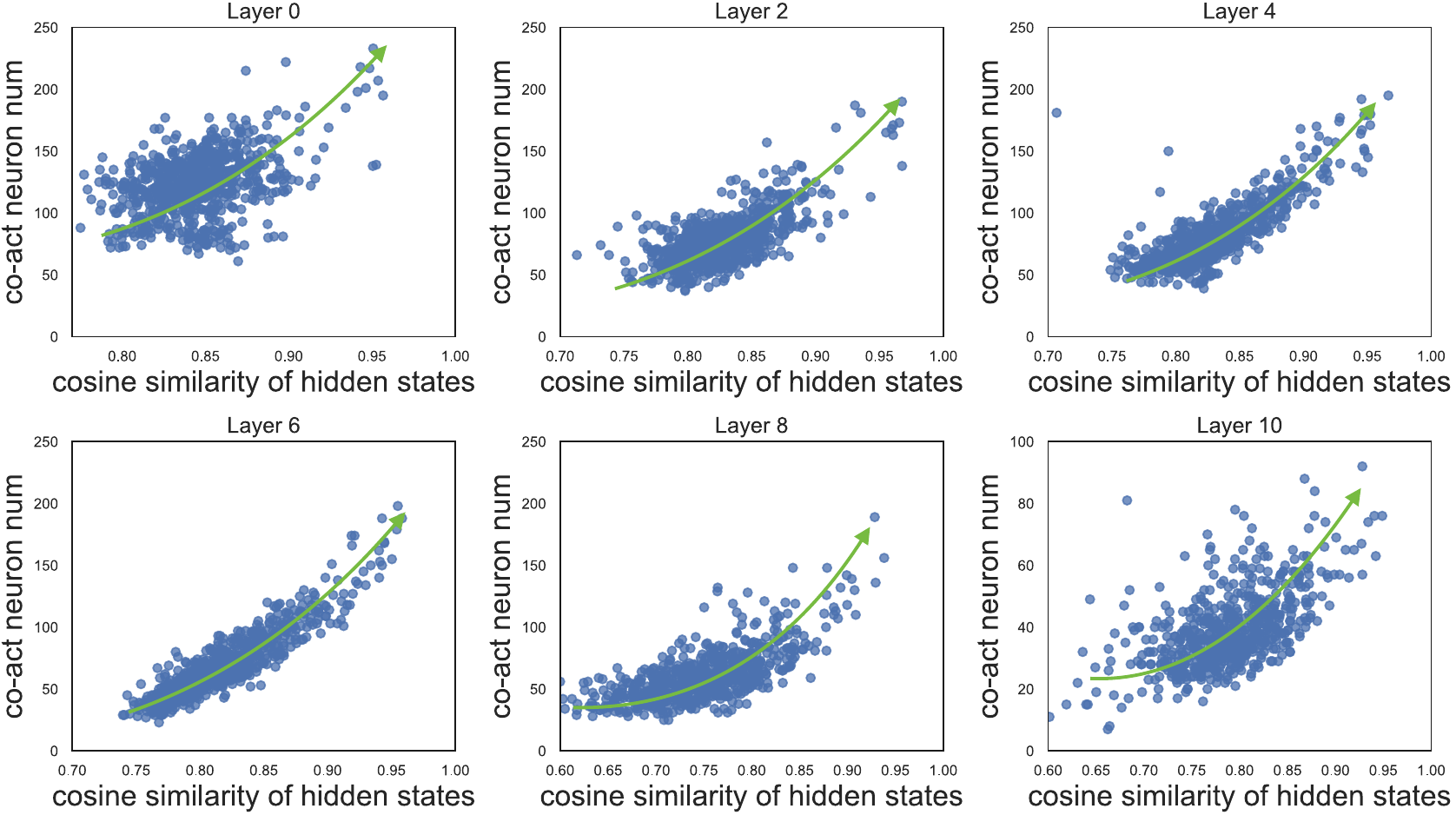}
  \end{subfigure}
   \begin{subfigure}{\textwidth}
    \centering
    \includegraphics[width=0.8\linewidth]{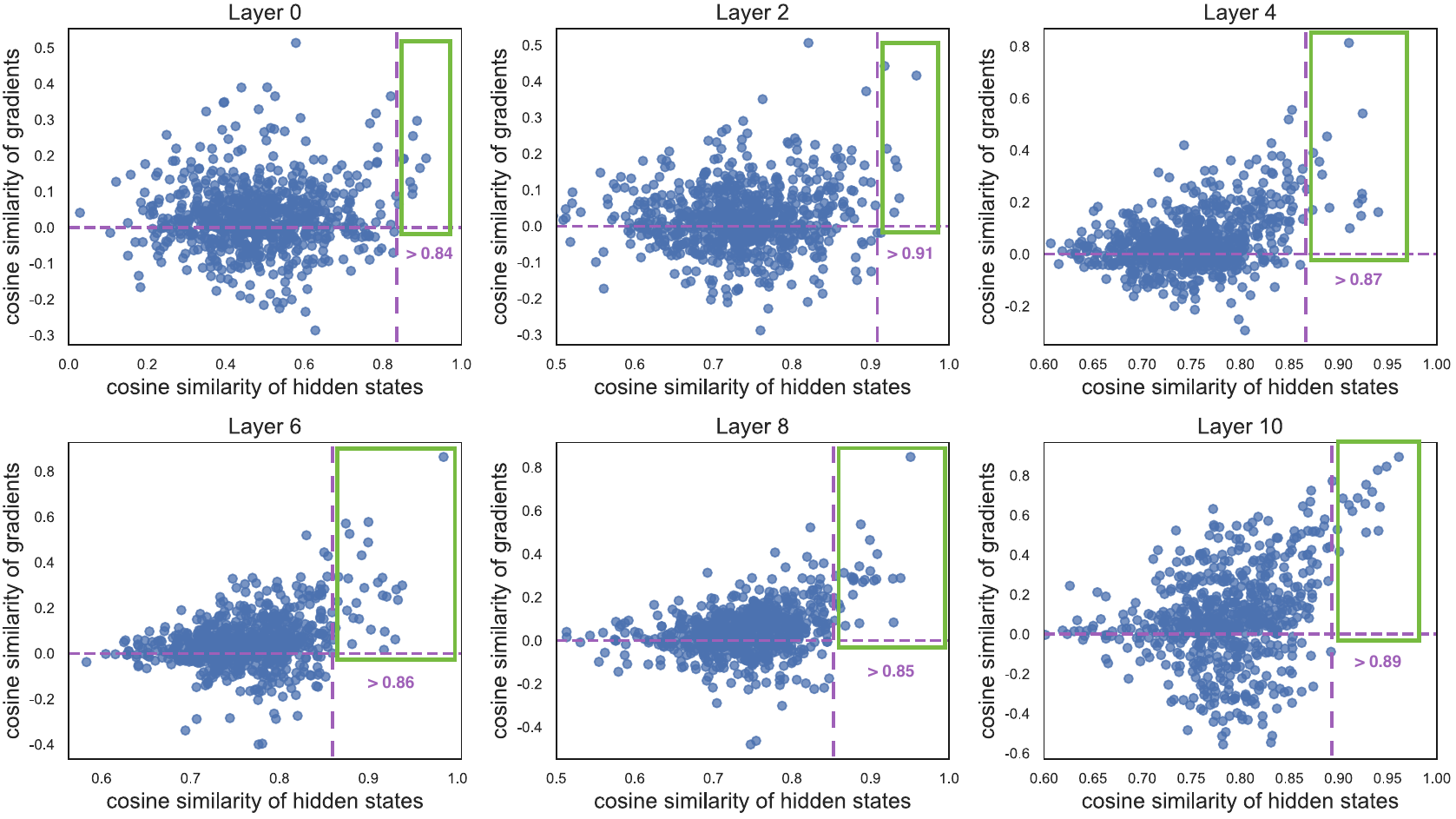}
  \caption{Visualization of the relationship between cosine similarity of hidden states ($\cos{\theta_{p,q}}$) and the co-activated neurons, as well as the relationship between cosine similarity of hidden states ($\cos{\theta_{p,q}}$) and cosine similarity of gradients ($\cos{\gamma_{p,q}}$) for $\boldeq{W}_{in}$ with Roberta on the FEVER dataset.}
  \end{subfigure}
   \begin{subfigure}{\textwidth}
    \centering
\includegraphics[width=0.8\linewidth]{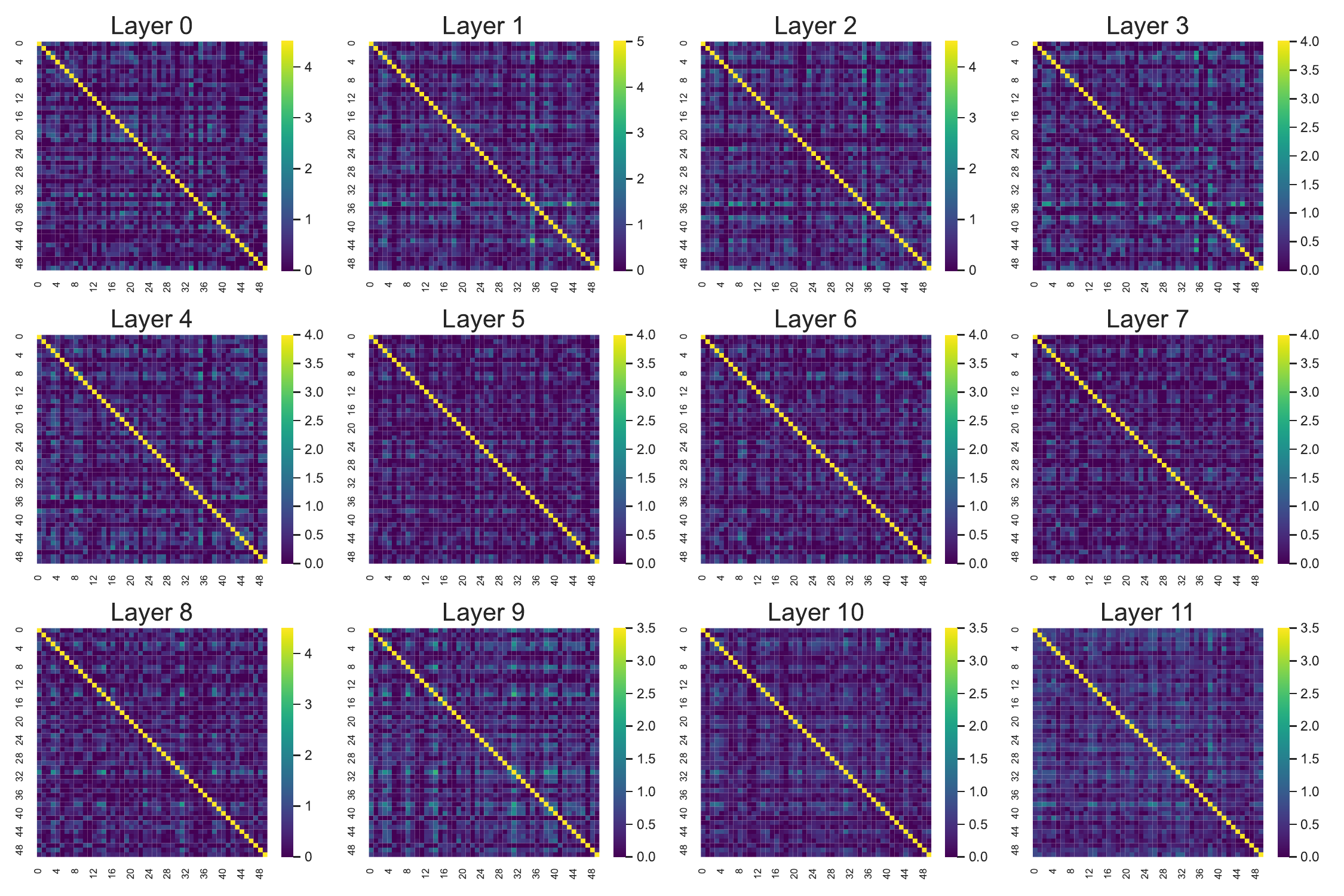}
  \caption{Visualization of \camera{submatrix of}  $\boldeq{W}_{in}^T\boldeq{W}_{in}$ at different layers in Roberta. The approximately orthogonal matrix of $\boldeq{W}_{in}$ holds $\boldeq{W}_{in}^T\boldeq{W}_{in} \approx \lambda\boldeq{I}$, where $\lambda$ is a constant.}
  \end{subfigure}
  \caption{Visualization for $\boldeq{W}_{in}$.}
  \label{fig:in_gradient_all}
\end{figure}

\begin{figure}[H]
 \centering
   \begin{subfigure}{\textwidth}
    \centering
\includegraphics[width=0.8\linewidth]{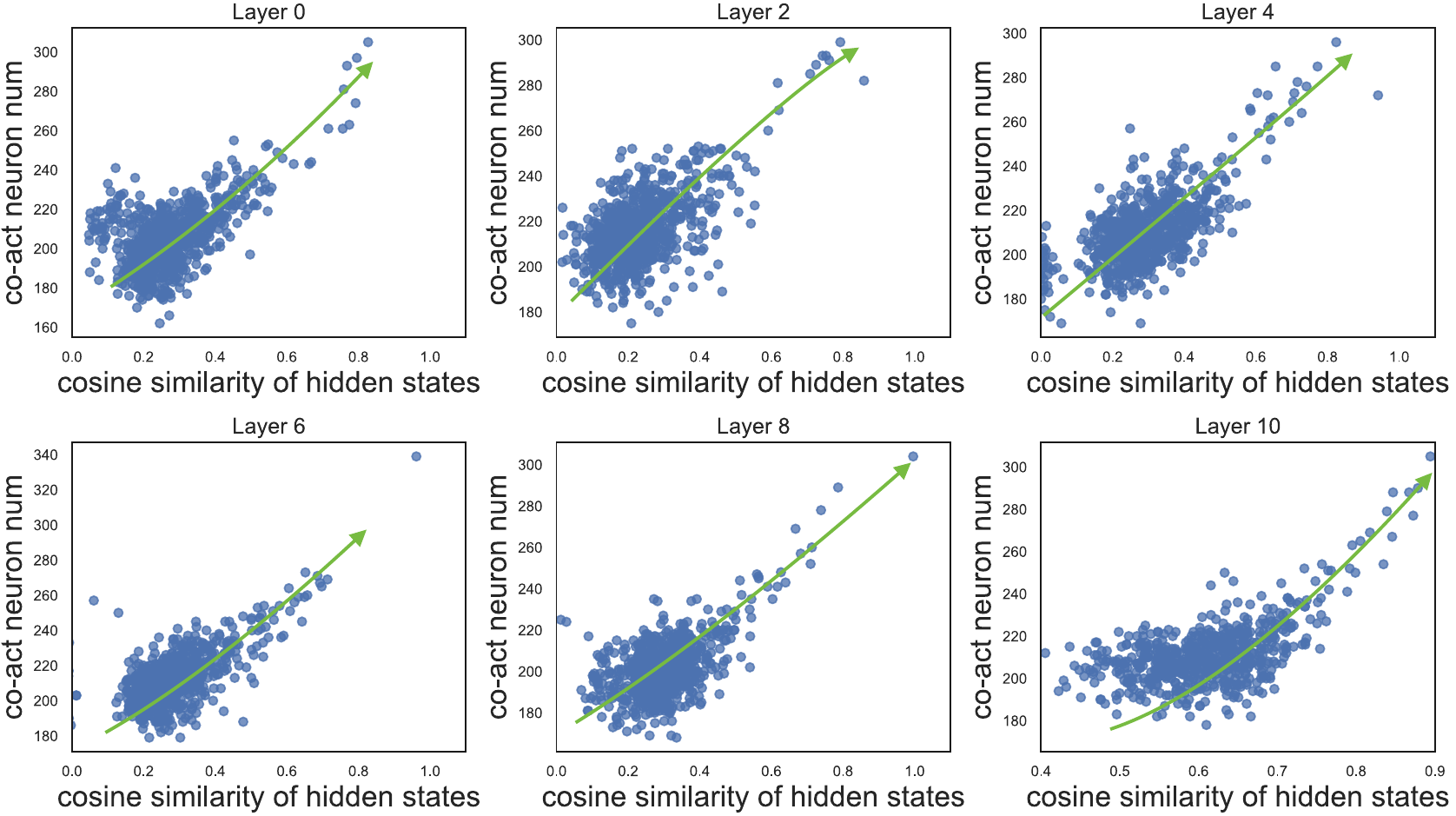}
  \end{subfigure}
   \begin{subfigure}{\textwidth}
    \centering
    \includegraphics[width=0.8\linewidth]{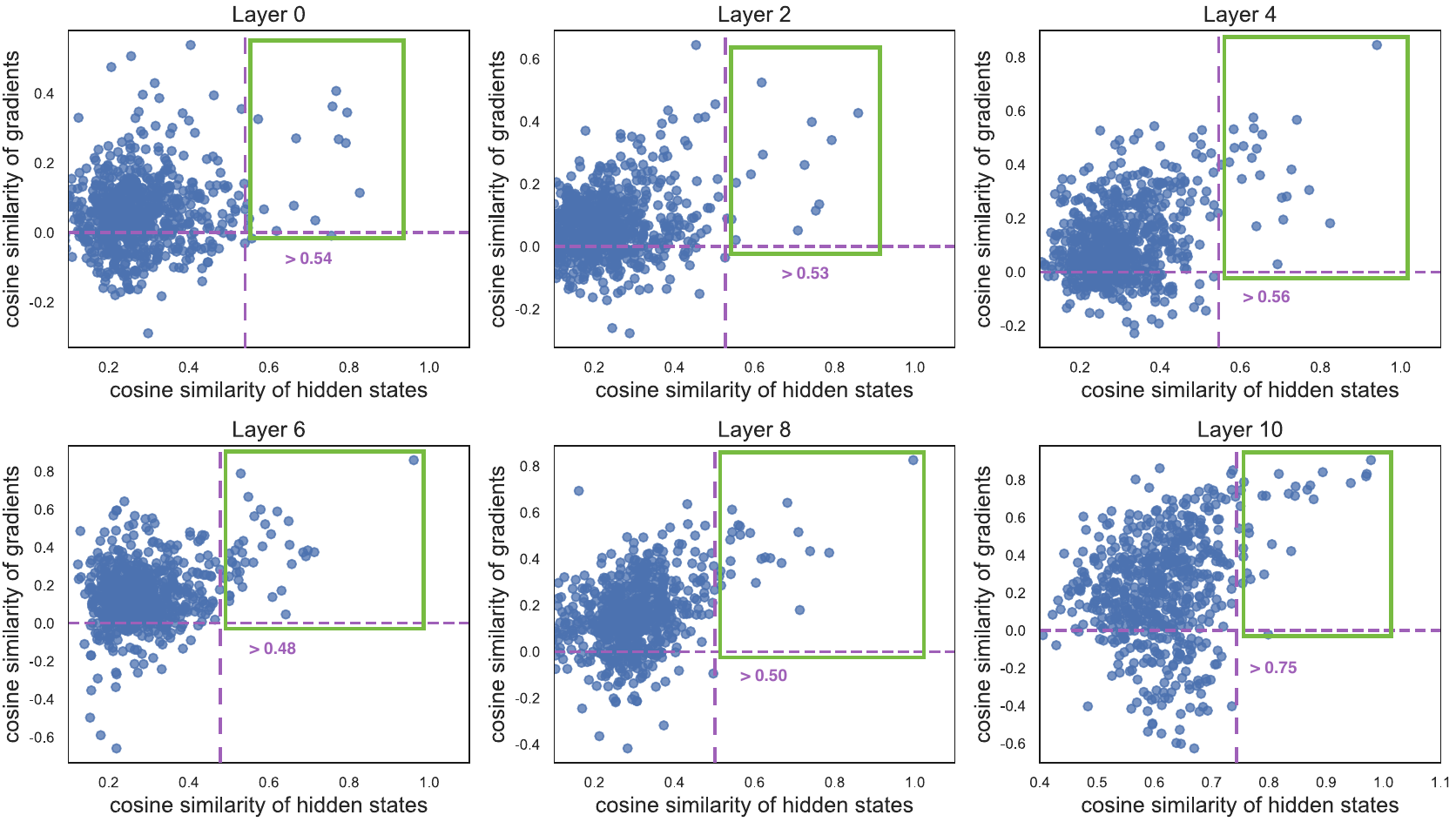}
  \caption{Visualization of the relationship between cosine similarity of hidden states ($\cos{\theta_{p,q}}$) and the co-activated neurons, as well as the relationship between cosine similarity of hidden states ($\cos{\theta_{p,q}}$) and cosine similarity of gradients ($\cos{\gamma_{p,q}}$) for $\boldeq{W}_{out}$ with Roberta on the FEVER dataset.}
  \end{subfigure}
   \begin{subfigure}{\textwidth}
    \centering
    \includegraphics[width=0.8\linewidth]{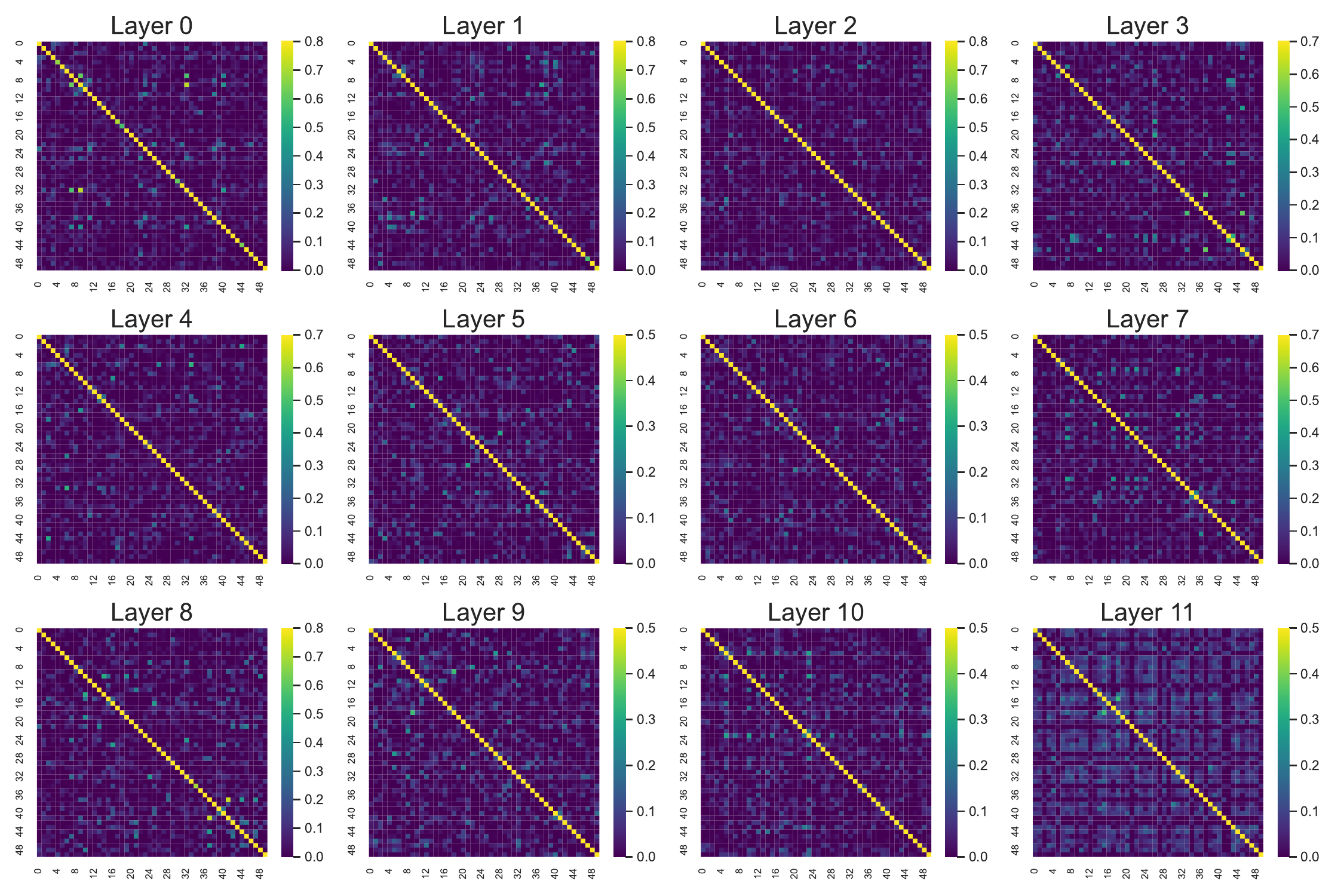}
  \caption{Visualization of \camera{submatrix of} $\boldeq{W}_{out}^T\boldeq{W}_{out}$ at different layers in Roberta. The approximately orthogonal matrix of $\boldeq{W}_{out}$ holds $\boldeq{W}_{out}^T\boldeq{W}_{out} \approx \lambda\boldeq{I}$, where $\lambda$ is a constant.}
  \end{subfigure}
  \caption{Visualization for $\boldeq{W}_{out}$.}
  \label{fig:out_gradient_all}
\end{figure}

Additionally, due to the approximate orthogonality of $\boldeq{W}_{in/out}$ as shown in Fig. \ref{fig:in_gradient_all} (b) and Fig. \ref{fig:out_gradient_all} (b), the angle between $\boldeq{X}_{p,:}$ and $\boldeq{X}_{q,:}$ is similar to that between $\boldeq{Y}_{p,:}$ and $\boldeq{Y}_{q,:}$. Therefore, we exploit the angle concentration of hidden states $\boldeq{Y}$ in our paper. Accordingly, we exploit the angle concentration of activations induced by each \gls{SVD} component, \camera{as a heuristic indicator}, to measure its contribution---\textit{dispersed} or \textit{concentrated}---to the angle concentration of the hidden states $\boldeq{Y}$. Consequently, this contribution (\textit{dispersed} or \textit{concentrated}) implies which singular components dominate the informative versus noisy contributions to the update and subsequent effective adaptation.

\newpage

\section{Transmuting Weight Updates to Steering Prompts}
\label{sec:appendix_weights_to_prompt}

Recent studies \citep{dherin2025learning,mazzawi2025transmuting,goldwaser2025equivalence} reveal the implicit weight updates within the context, namely, transmuting added/modified prompts into weight updates. In our paper, we transmute weight updates into steering prompts so that the parameters of \glspl{LLM} can stay unmodified. In Equation (\ref{eq:rank_forward}), we have the standard forward pass, i.e., $\boldeq{X}\boldeq{W}^T =\boldeq{Y}$. Given the final candidate with deterministic rank reduction from our forward-only approach, we can get the weight update $\Delta\boldeq{W}$. Considering the weights without updates and with updates, we have 
\begin{align}
\begin{cases}
\boldeq{X}\boldeq{W}^T=\boldeq{Y}
 \\
\boldeq{X}(\boldeq{W}+\Delta\boldeq{W})^T=\boldeq{Y}^\prime
\end{cases}
\Longrightarrow 
\boldeq{Y}^\prime - \boldeq{Y} = \boldeq{X}\Delta\boldeq{W}^T, \nonumber
\end{align}
where $\boldeq{Y}^\prime$ are the hidden states of the output with the weight-updated \glspl{LLM}. Therefore, for a specific weight update $\Delta\boldeq{W}$ and a specific input $\boldeq{X}$, we transmute this weight update into a change in output. 

To extend this input-wise transmutation to the unified transmutation for downstream tasks, we derive the steering prompts by averaging over validation datasets $\mathcal{D}_{\mathrm{std}}$. The steering prompts $\Delta\boldeq{Y}$, \camera{which broadcasts across token positions}, are denoted:
\begin{align}
\Delta\boldeq{Y} = \mathbb{E}_{\boldeq{X} \sim \mathcal{D}_{\mathrm{std}}}\left[\frac{1}{n}  \sum_{i=1}^{n}  \boldeq{X}_{i,:}\Delta\boldeq{W}^T \right], \nonumber
\end{align}
where $\boldeq{X}_{i, :}$ is the $i$-th token with the sequence length $n$ of $\boldeq{X}$.

Taking the evaluation with Roberta on the Bios Profession dataset as the example, the final candidate with rank reduction from our forward-only approach is $[\boldeq{W}_{in},3,90\%]$ from Table \ref{tab:para_roberta}. We transmute weight updates to steering prompts, with different sizes of validation datasets, as presented in Table \ref{tab:acc_with_deltaY}. The results show that the steering prompts achieve nearly identical accuracy to the weight updates, which justifies that we can transmute weight updates to steering prompts with unmodified parameters.
\begin{table}[H]
  \caption{The reported accuracy (Acc, \%) for Roberta on Bios Profession. We can transmute weight updates to steering prompts with unmodified parameters in \glspl{LLM}.}
  \label{tab:acc_with_deltaY}
  \centering
  \begin{tabular}{ccccccc}
    \toprule
     \multirow{2}{*}{Baseline} & \multirow{2}{*}{Ours (with $\Delta\boldeq{W}$)} & \multicolumn{5}{c}{Steering Prompts (transmute $\Delta\boldeq{W}$ to $\Delta\boldeq{Y}$)}\\
     \cmidrule(lr){3-7}
& &20\% of $\mathcal{D}_{\mathrm{std}}$&40\% of $\mathcal{D}_{\mathrm{std}}$&60\% of $\mathcal{D}_{\mathrm{std}}$&80\% of $\mathcal{D}_{\mathrm{std}}$&$\mathcal{D}_{\mathrm{std}}$\\ 
    \midrule
    64.5  & 71.9 & 72.0 & 71.9 & 71.8 & 72.0 & 72.0\\
    \bottomrule
  \end{tabular}
\end{table}

\begin{figure}[H]
\vspace{-10pt}
 \centering
    \includegraphics[width=0.7\linewidth]{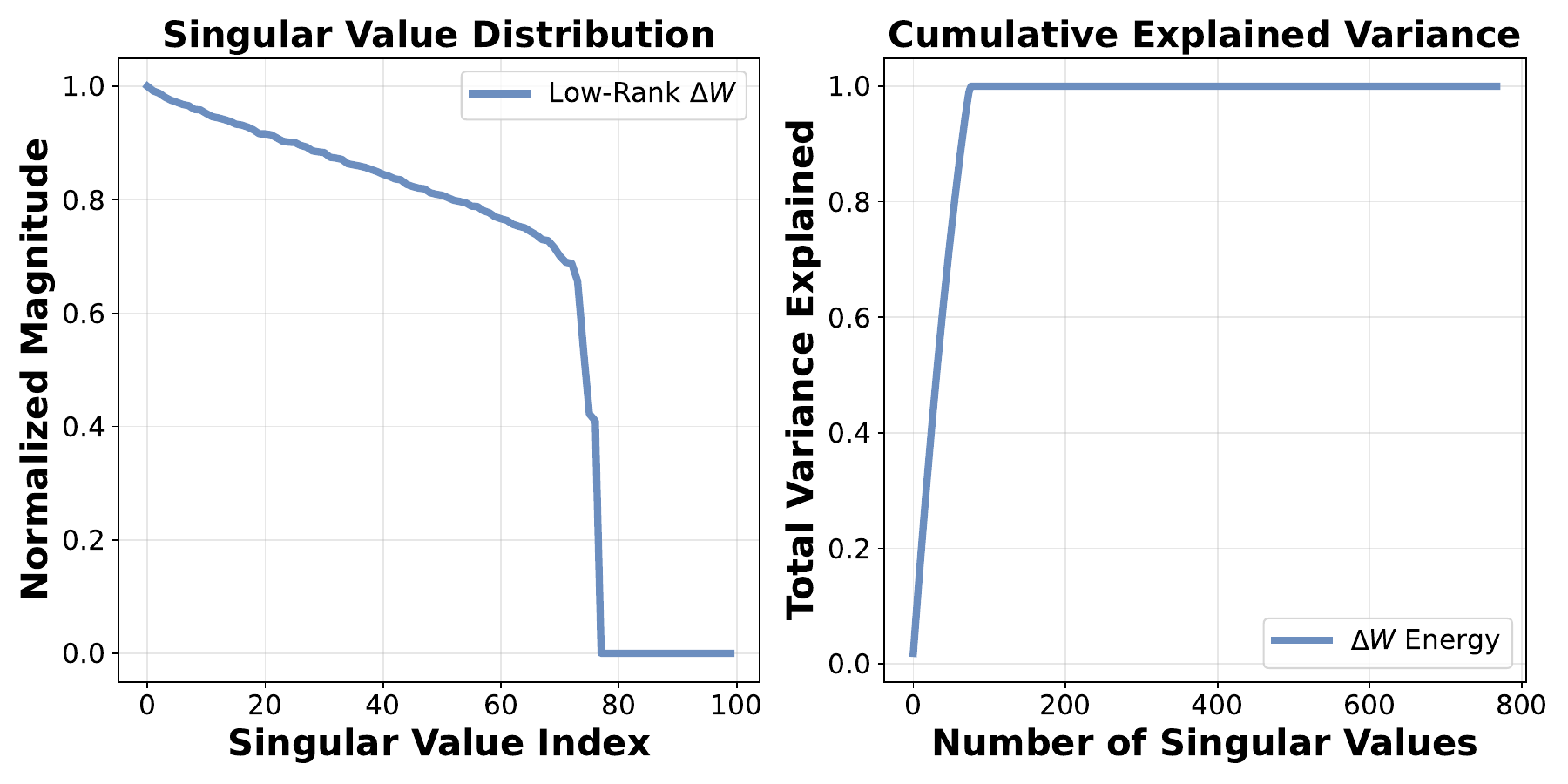}
  \caption{The \gls{SVD} of the weight updates $\Delta\boldeq{W}$.}
  \label{fig:deltaW_svd}
  \vspace{-10pt}
\end{figure}

\vspace{-15pt}
Besides, we analyze the \gls{SVD} of $\Delta\boldeq{W}$, as demonstrated in Fig. \ref{fig:deltaW_svd}. The effective ranks of $\Delta\boldeq{W}$ are equal to the removed ranks (removing $1-k\%$ of ranks with $k\%=90\%$), i.e., $10\%\times\underset{\mathrm{hidden\_size}}{768}$. 
Therefore, we can further decompose the steering prompts $\Delta\boldeq{Y}$ into $\Delta\boldeq{Y}^{(i)}$ in different \gls{SVD} components, $\Delta\boldeq{Y} = \sum_{i=1}^e\Delta\boldeq{Y}^{(i)}$, where $e$ is the effective ranks and $e<r$. Moreover, given this study \citep{kaushik2026do} showing that deep neural networks converge to shared subspaces and these studies \citep{meng2022locating,li2024basis,meng2024pissa,sun2026unleashing} showing the capacity of low-rank update in \gls{SVD} components (even the singular value itself), future work could investigate how \camera{shared and generated expert weight matrices \citep{du2026didi,du2026nestedvla}} and steering prompts $\Delta\boldeq{Y}^{(i)}$ in different ranks are related to adaptation across downstream tasks and \glspl{LLM} \camera{\citep{touvron2023llama,jiang2023mistral7b,yang2024qwen2technicalreport,huangnas,bi2024deepseek,ren2025optin,google_gemma4_2026}}.

\addtocontents{toc}{\protect\setcounter{tocdepth}{1}}
\section{Additional Experiments and Analyses}
\label{sec:update_reviewer}

\subsection{Empirical Results to Show the Efficiency Compared to LASER-Grad}
\label{sec:emirical_laser_grad}
We did an initial empirical measurement of realistic memory consumption, computational complexity, and execution time. We mainly measure the memory change during the core running, i.e., calculating angle concentration for our approach (only forward passes) and calculating gradients for LASER-Grad \cite{sreeram2025compress} (forward pass and backward pass). The empirical results are shown in Table \ref{tab:empirical_laser_grad}. For Roberta on the FEVER dataset, LASER-Grad \cite{sreeram2025compress} has around 2.6x more memory change than our approach; for Roberta on the Bios Profession dataset with a longer average sequence length, LASER-Grad \cite{sreeram2025compress} has around 4.5x more memory change than our approach. The results are expected because our forward-only approach avoids the backward pass. Moreover, our forward-only search achieves lower computational complexity and shorter execution time when compared to LASER-Grad \cite{sreeram2025compress}. 

\begin{table}[!ht]
  \caption{Measurement of realistic memory consumption, computational complexity, and execution time. Note that the average complexity and average execution time are averages over all samples in $\mathcal{D}_{\mathrm{subset}}$. When calculating \gls{FLOPs} for a specific sample, the backward pass has approximately twice the \gls{FLOPs} of the forward pass \cite{kaplan2020scaling,yin2025compute}.}
  \label{tab:empirical_laser_grad}
  \centering
   % \resizebox{\columnwidth}{!}{
  \begin{tabular}{ccccc}
    \toprule
    Roberta & Our Approach & Laser-Grad  & Our Approach & Laser-Grad \\
    \midrule
    Dataset  & \multicolumn{2}{c}{FEVER}  & \multicolumn{2}{c}{Bios Profession} \\
    Average Sequence Length & \multicolumn{2}{c}{27}  & \multicolumn{2}{c}{100} \\
    \cmidrule(lr){2-3}
    \cmidrule(lr){4-5}
    Memory Change$\downarrow$ & 130 \gls{MB} & 343 \gls{MB} & 235 \gls{MB} & 1062 \gls{MB} \\
    Average Complexity$\downarrow$ & 3G \gls{FLOPs} & 10G \gls{FLOPs} &13G \gls{FLOPs} & 38G \gls{FLOPs} \\
    Average Execution Time$\downarrow$ &1.00 second & 3.34 seconds & 1.31 seconds & 5.55 seconds \\
    \bottomrule
  \end{tabular}%}
\end{table}

\subsection{Comparison to In-context Learning}
\label{sec:compare_ICL}

We did extra experiments with LLaMA2 \citep{touvron2023llama} on the FEVER dataset using \gls{ICL}. In these experiments, few-shot prompts improve performance compared to zero-shot prompts (i.e., Baseline), as shown in Table \ref{tab:compare_ICL}. For example, the accuracy of 4-shot prompts is 76.2\%, and the accuracy of 16-shot prompts is 77.2\%, while the accuracy of zero-shot prompts is 59.3\%. Therefore, \gls{ICL} as a tuning-free method, is also expected to achieve better accuracy but at a higher cost (for example, adding more examples from $\mathcal{D}_{\mathrm{subset}}$ with longer tokens, incurring more inference cost).

\begin{table}[!ht]
  \caption{The higher the cost in \gls{ICL} computation, the better the accuracy overall.}
  \label{tab:compare_ICL}
  \centering
   \resizebox{\columnwidth}{!}{
  \begin{tabular}{cccccccc}
    \toprule
    LLaMA2 on FEVER & Baseline & Ours (std)  & \gls{ICL} (1-shot) & \gls{ICL} (2-shot) & \gls{ICL} (4-shot) & \gls{ICL} (8-shot) & \gls{ICL} (16-shot) \\
    \midrule
    Accuracy$\uparrow$ & 59.3&64.5&74.6&75.4&76.2&74.3&77.2 \\
    Additional Tokens$\downarrow$ &0&0&33&61&114&237&461 \\
    \bottomrule
  \end{tabular}}
\end{table}

\subsection{Comparison to Simpler Forward-Only Baselines}
\label{sec:compare_simple_FO}
To have the new forward-only baseline, we directly replace the module of angle concentration (cosine similarity) by activation norms and activation variance, cosine dispersion without \gls{SVD} decomposition, as well as the random top-layer selection, while the harness is the same as our approach. Note that cosine dispersion without \gls{SVD} decomposition directly measures the dispersion of different weights. Overall, the results are shown in Table \ref{tab:compare_simple_FO}.

\begin{table}[!ht]
  \caption{Our forward-only approach is more effective and competitive than other simpler forward-only baselines. $\star$Note that for activation norms (Roberta on Bios Profession), the top-5 weight matrices are $[\boldeq{W}_{out},8],[\boldeq{W}_{out},9],[\boldeq{W}_{in},1],[\boldeq{W}_{in},2],[\boldeq{W}_{in},3]$. Therefore, by the harness of our efficient hierarchical search, the final candidate is $[\boldeq{W}_{in},3]$.}
  \label{tab:compare_simple_FO}
  \centering
   \resizebox{\columnwidth}{!}{
  \begin{tabular}{ccccc}
    \toprule
    \multirow{2}{*}{Roberta} & \multicolumn{2}{c}{Bios Gender}   & \multicolumn{2}{c}{Bios Profession}  \\
    \cmidrule(lr){2-3}
    \cmidrule(lr){4-5}
      & Final Candidate   & Accuracy$\uparrow$(\%)   & Final Candidate   & Accuracy$\uparrow$(\%) \\
    \midrule
    Direct Inference & - & 87.5 & - & 64.5 \\
    Activation Norms & $[\boldeq{W}_{in},2,80\%]$ & 90.0 & $[\boldeq{W}_{in},3,90\%]\star$ & 71.9 \\
    Activation Variance & $[\boldeq{W}_{out},11,5\%]$ & 91.8 & $[\boldeq{W}_{out},7,1\%]$ & 67.2 \\
    Cosine Dispersion Without SVD & $[\boldeq{W}_{out},0,40\%]$ & 89.2 & $[\boldeq{W}_{out},1,1\%]$ & 70.9 \\
    {Random Top-Layer Selection} & $[\boldeq{W}_{in},11,1\%]$ & 91.8 & $[\boldeq{W}_{out},11,40\%]$ & 65.1 \\
    {Random Top-Layer Selection} & $[\boldeq{W}_{out},10,60\%]$ & 89.9 & $[\boldeq{W}_{in},11,80\%]$ & 64.9 \\
    Ours (hierarchical) & $[\boldeq{W}_{in},9,90\%]$ & 93.4 & $[\boldeq{W}_{in},3,90\%]$ & 71.9 \\
    \bottomrule
  \end{tabular}}
\end{table}

The results first show that random top-layer selection underperforms our approach in the evaluated settings. For Roberta on Bios Profession, the activation variance cannot compete with our approach. While the activation norms search the same final candidate as ours, resulting in 71.9\% accuracy, activation norms cannot be generalized to other cases, such as Roberta on Bios Gender. For Roberta on Bios Gender, the activation norms and activation variance are both worse than our approach. Moreover, the cosine dispersion without SVD decomposition also cannot compete with our approach over Bios Gender and Bios Profession. Overall, our approach, as the forward-only approach, is more effective and competitive than other simpler forward-only baselines.

\end{document}